%% file: errata.tex
\documentclass{article}

\PassOptionsToPackage{numbers, compress}{natbib}
\usepackage[preprint]{neurips_2025}

\usepackage[utf8]{inputenc} % allow utf-8 input
\usepackage[T1]{fontenc}    % use 8-bit T1 fonts
\usepackage{hyperref}       % hyperlinks
\usepackage{url}            % simple URL typesetting
\usepackage{booktabs}       % professional-quality tables
\usepackage{amsfonts}       % blackboard math symbols
\usepackage{nicefrac}       % compact symbols for 1/2, etc.
\usepackage{microtype}      % microtypography
\usepackage{amssymb}
\usepackage[table]{xcolor}   % for colored squares
\usepackage{caption}  
\usepackage{amsmath}
\usepackage{graphicx}
\usepackage{wrapfig}
\usepackage{listings}
\usepackage{pifont}
\usepackage[utf8]{inputenc}
\usepackage{tcolorbox}
\usepackage{multirow}
\usepackage{enumitem}
\usepackage{longtable,array}
\usepackage{tabularray}
\UseTblrLibrary{varwidth}
\newcolumntype{P}[1]{>{\raggedright\arraybackslash}p{#1}}
\definecolor{prop}{rgb}{0.973,0.463,0.427} 
\definecolor{think}{rgb}{0.000,0.749,0.769}
\definecolor{text}{rgb}{0.486,0.682,0.000}

\newcommand{\cmark}{\textcolor{green!80!black}{\ding{51}}}
\newcommand{\xmark}{\textcolor{red}{\ding{55}}}

\newcommand{\benchmark}{\textsc{Spot}}

\newcommand{\eg}{\emph{e.g.,} }

\title{When AI Co‑Scientists Fail: \benchmark—a Benchmark for Automated Verification of Scientific Research}

\author{Guijin Son$^{1,2}$ \quad Jiwoo Hong$^{3}$ \quad Honglu Fan$^{2}$ \quad Heejeong Nam$^{4}$ \\  \textbf{Hyunwoo Ko}$^{1}$ \quad \textbf{Seungwon Lim}$^{5}$ \quad \textbf{Jinyeop Song}$^{6}$  \quad \textbf{Jinha Choi}$^{5}$ \\ \textbf{Gonçalo Paulo}$^{2}$ \quad \textbf{Youngjae Yu}$^{5}$ \quad \textbf{Stella Biderman}$^{2}$ \\ \\ OneLineAI$^{1}$ \qquad EleutherAI$^{2}$ \qquad KAIST AI$^{3}$ \qquad Boeing Korea$^{4}$ \\  Yonsei University$^{5}$ \qquad MIT$^{6}$\\ \\
\texttt{spthsrbwls123@yonsei.ac.kr} }

\begin{document}

\maketitle

\begin{abstract}
Recent advances in large language models (LLMs) have fueled the vision of automated scientific discovery, often called AI Co-Scientists. To date, prior work casts these systems as generative co-authors responsible for crafting hypotheses, synthesizing code, or drafting manuscripts. In this work, we explore a complementary application: using LLMs as verifiers to automate the \textbf{academic verification of scientific manuscripts}. To that end, we introduce \benchmark{}, a dataset of 83 published papers paired with 91 errors significant enough to prompt errata or retraction, cross-validated with actual authors and human annotators. Evaluating state-of-the-art LLMs on \benchmark{}, we find that none surpasses 21.1\% recall or 6.1\% precision (o3 achieves the best scores, with all others near zero). Furthermore, confidence estimates are uniformly low, and across eight independent runs, models rarely rediscover the same errors, undermining their reliability. Finally, qualitative analysis with domain experts reveals that even the strongest models make mistakes resembling student-level misconceptions derived from misunderstandings. These findings highlight the substantial gap between current LLM capabilities and the requirements for dependable AI-assisted academic verification.
% \footnote{: \url{https://huggingface.co/datasets/amphora/SPOT-MetaData}}
\begin{tblr}{
  colsep    = .1em,
  rowsep    = .1pt,
  stretch   = 0,
  cells     = {valign=m}
}
  \centering
  \raisebox{-1.6ex}{\includegraphics[width=2.0em,keepaspectratio]{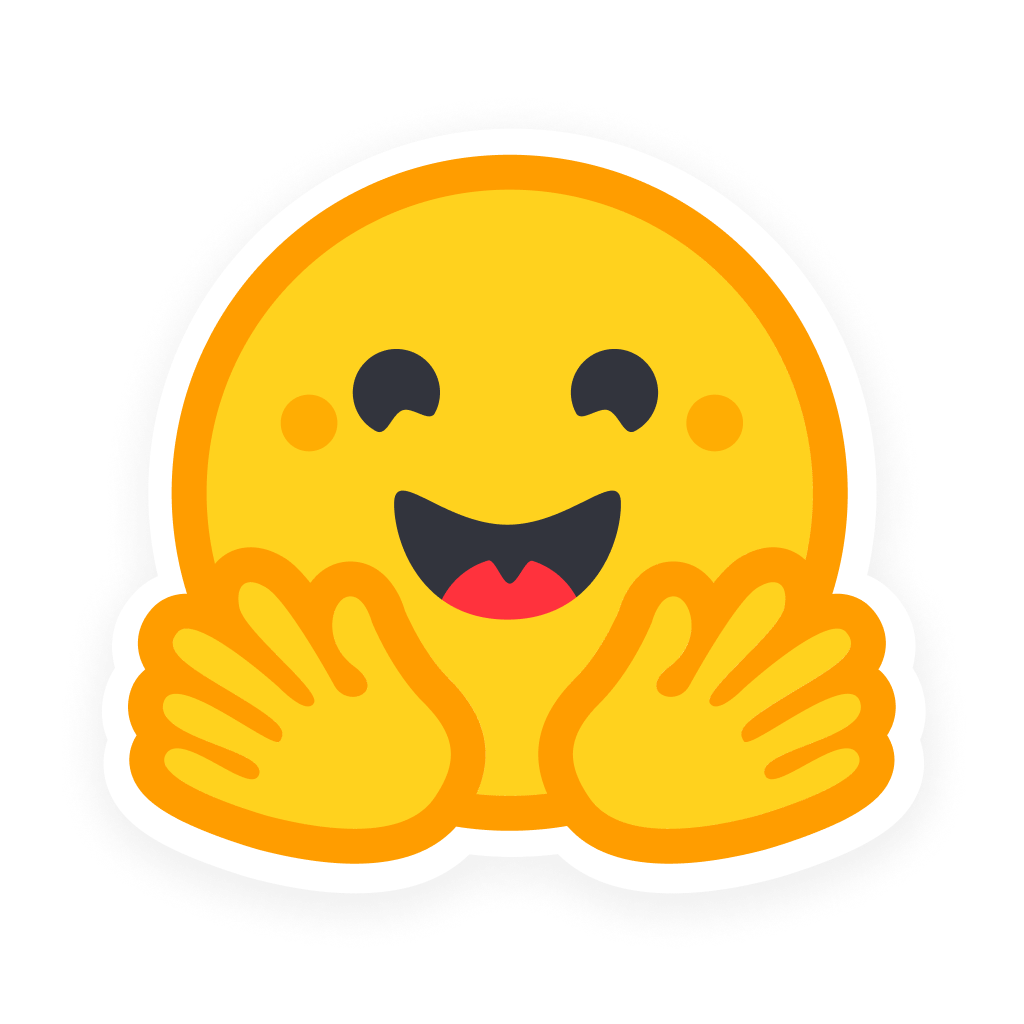}}
  & {\footnotesize\url{https://huggingface.co/datasets/amphora/SPOT-MetaData}}
\end{tblr}

\end{abstract}

\section{Introduction}

\input{section/introduction}

\section{\benchmark: Automating Error Detection in Scientific Research}
\label{sec:benchmark}
\input{section/benchmark}

\section{Main Results and Analysis}\label{sec:results}

\input{section/results}

% \section{Additional Analysis}
% \label{sec:add_results}

% \begin{itemize}
%   \item \textbf{Test‐time scaling}
%     \begin{itemize}
%       \item S1-style budget forcing
%       \item Best-of-N Majority Voting
%     \end{itemize}

% \end{itemize}

\section{Case Study}
\label{sec:case_study}

\input{section/case_study}

%qualitative analysis on llm outputs

\section{Related Works}
\label{sec:related_works}

\input{section/related_works}

\section{Conclusion}
\label{sec:conclusion}
\input{section/conclusion}

% \begin{ack}
% meh
% \end{ack}

\newpage
\bibliographystyle{unsrtnat} 
\bibliography{custom}

%%%%%%%%%%%%%%%%%%%%%%%%%%%%%%%%%%%%%%%%%%%%%%%%%%%%%%%%%%%%

\appendix

%%%%%%%%%%%%%%%%%%%%%%%%%%%%%%%%%%%%%%%%%%%%%%%%%%%%%%%%%%%%

\newpage
\section*{NeurIPS Paper Checklist}

\begin{enumerate}

\item {\bf Claims}
    \item[] Question: Do the main claims made in the abstract and introduction accurately reflect the paper's contributions and scope?
    \item[] Answer: \answerYes{} % Replace by \answerYes{}, \answerNo{}, or \answerNA{}.
    \item[] Justification: The abstract and introduction is written to summarize our work.

\item {\bf Limitations}
    \item[] Question: Does the paper discuss the limitations of the work performed by the authors?
    \item[] Answer: \answerYes{} % Replace by \answerYes{}, \answerNo{}, or \answerNA{}.
    \item[] Justification: See Appendix~\ref{app:limitations}.

\item {\bf Theory assumptions and proofs}
    \item[] Question: For each theoretical result, does the paper provide the full set of assumptions and a complete (and correct) proof?
    \item[] Answer: \answerYes{} % Replace by \answerYes{}, \answerNo{}, or \answerNA{}.
    \item[] Justification: This paper discusses a novel benchmark without theoretical results. We provide how our confidence metric is derived in Appendix~\ref{app:confidence_for_passk}.

    \item {\bf Experimental result reproducibility}
    \item[] Question: Does the paper fully disclose all the information needed to reproduce the main experimental results of the paper to the extent that it affects the main claims and/or conclusions of the paper (regardless of whether the code and data are provided or not)?
    \item[] Answer: \answerYes{} % Replace by \answerYes{}, \answerNo{}, or \answerNA{}.
    \item[] Justification: We provide details on the parsing process in Section~\ref{sec:benchmark}, and further details on the prompts and generation configurations in Appendix~\ref{app:details_on_eval}. The benchmark and codes are also provided through supplementary materials.

\item {\bf Open access to data and code}
    \item[] Question: Does the paper provide open access to the data and code, with sufficient instructions to faithfully reproduce the main experimental results, as described in supplemental material?
    \item[] Answer: \answerYes{} % Replace by \answerYes{}, \answerNo{}, or \answerNA{}.
    \item[] Justification: The experiments are fully reproducible using the provided code. Minor variations may arise from proprietary model availability or inherent stochasticity, but these do not affect our overall conclusions.

\item {\bf Experimental setting/details}
    \item[] Question: Does the paper specify all the training and test details (e.g., data splits, hyperparameters, how they were chosen, type of optimizer, etc.) necessary to understand the results?
    \item[] Answer: \answerYes{} % Replace by \answerYes{}, \answerNo{}, or \answerNA{}.
    \item[] Justification: We do not train our own models, but prompts and generation configurations for inference are provided in Appendix~\ref{app:details_on_eval}. 

\item {\bf Experiment statistical significance}
    \item[] Question: Does the paper report error bars suitably and correctly defined or other appropriate information about the statistical significance of the experiments?
    \item[] Answer: \answerYes{} % Replace by \answerYes{}, \answerNo{}, or \answerNA{}.
    \item[] Justification: Table~\ref{tab:main-results}, Table~\ref{tab:multi_modality} all provide means and standard deviation of multiple independent trials and bootstrapping. 

\item {\bf Experiments compute resources}
    \item[] Question: For each experiment, does the paper provide sufficient information on the computer resources (type of compute workers, memory, time of execution) needed to reproduce the experiments?
    \item[] Answer: \answerNA{} % Replace by \answerYes{}, \answerNo{}, or \answerNA{}.
    \item[] Justification: Our experiments are conducted entirely through APIs. We use the official provider for each model if available; if not, we use OpenRouter\footnote{\url{https://openrouter.ai/}}. Total API expenditures amount to approximately \$5,000.
    
\item {\bf Code of ethics}
    \item[] Question: Does the research conducted in the paper conform, in every respect, with the NeurIPS Code of Ethics \url{https://neurips.cc/public/EthicsGuidelines}?
    \item[] Answer: \answerYes{} % Replace by \answerYes{}, \answerNo{}, or \answerNA{}.
    \item[] Justification:

\item {\bf Broader impacts}
    \item[] Question: Does the paper discuss both potential positive societal impacts and negative societal impacts of the work performed?
    \item[] Answer: \answerNA{} % Replace by \answerYes{}, \answerNo{}, or \answerNA{}.
    \item[] Justification: We don't see negative societal impact in this work.
    
\item {\bf Safeguards}
    \item[] Question: Does the paper describe safeguards that have been put in place for responsible release of data or models that have a high risk for misuse (e.g., pretrained language models, image generators, or scraped datasets)?
    \item[] Answer: \answerYes{} % Replace by \answerYes{}, \answerNo{}, or \answerNA{}.
    \item[] Justification: We do not train custom models. Our sharing policy for the \benchmark{} dataset varies with the copyright status of each original paper; see Appendix~\ref{app:spot} for details.

\item {\bf Licenses for existing assets}
    \item[] Question: Are the creators or original owners of assets (e.g., code, data, models), used in the paper, properly credited and are the license and terms of use explicitly mentioned and properly respected?
    \item[] Answer: \answerYes{} % Replace by \answerYes{}, \answerNo{}, or \answerNA{}.
    \item[] Justification: All assets used in the paper are mentioned through footnotes or references.

\item {\bf New assets}
    \item[] Question: Are new assets introduced in the paper well documented and is the documentation provided alongside the assets?
    \item[] Answer: \answerYes{} % Replace by \answerYes{}, \answerNo{}, or \answerNA{}.
    \item[] Justification: See Section~\ref{sec:benchmark} and Appendix~\ref{app:spot}.

\item {\bf Crowdsourcing and research with human subjects}
    \item[] Question: For crowdsourcing experiments and research with human subjects, does the paper include the full text of instructions given to participants and screenshots, if applicable, as well as details about compensation (if any)? 
    \item[] Answer: \answerYes{} % Replace by \answerYes{}, \answerNo{}, or \answerNA{}.
    \item[] Justification: An image of the annotation platform is available in Appendix~\ref{app:spot}.

\item {\bf Institutional review board (IRB) approvals or equivalent for research with human subjects}
    \item[] Question: Does the paper describe potential risks incurred by study participants, whether such risks were disclosed to the subjects, and whether Institutional Review Board (IRB) approvals (or an equivalent approval/review based on the requirements of your country or institution) were obtained?
    \item[] Answer: \answerNA{} % Replace by \answerYes{}, \answerNo{}, or \answerNA{}.
    \item[] Justification: Paper does not involve crowdsourcing nor research with human subjects.

\item {\bf Declaration of LLM usage}
    \item[] Question: Does the paper describe the usage of LLMs if it is an important, original, or non-standard component of the core methods in this research? Note that if the LLM is used only for writing, editing, or formatting purposes and does not impact the core methodology, scientific rigorousness, or originality of the research, declaration is not required.
    %this research? 
    \item[] Answer: \answerNo{} % Replace by \answerYes{}, \answerNo{}, or \answerNA{}.
    \item[] Justification: LLM services such as ChatGPT are used to draft and edit the manuscript and to generate code for visualizations.

\end{enumerate}

\input{section/appendix}

\end{document}

%% file: section/introduction.tex
% Over the past year, large language models (LLMs) have evolved from text generators to AI Co-Scientists~\citep{gottweis2025towards} that propose hypotheses, synthesize literature, plan experiments, and even draft manuscripts~\citep{luo2025llm4sr, yang2025transforming} \stella{The wording here accepts implicitly that they're functional at these tasks. Let's reconsider the phraseology}. 

From simple next‐token predictors~\citep{radford2018improving, brown2020language}, large language models (LLMs) have evolved to exhibit graduate‐level STEM proficiency~\citep{guo2025r, rein2024gpqa, feng2025physics}, generate hypotheses~\citep{si2024can, park2024can}, synthesize literature~\citep{he2025pasa}, and draft manuscripts~\citep{jain2024generative}. Such advances have driven interest in their deployment as “AI Co-Scientists”~\citep{gottweis2025towards,lu2024ai}, proving to be viable options in the "generative" role of scientific research. They have rediscovered established findings~\citep{penades2025ai} and generated novel hypotheses worthy of investigation across diverse fields~\citep{m2024augmenting, pan2025quantum, deepmind2025alphaevolve}. However, despite their widespread usage as ``generators'' in the forward pass of scientific research, their utility in the backward pass of academic verification or as \textbf{\underline{verifiers}} remains underexplored, a blind spot in which most systems lean on LLM judges~\citep{zheng2023judging} without validation on their credibility in reviewing scientific research. Prior research on factual verification has primarily focused on everyday knowledge tasks~\citep{chen2019tabfact, bekoulis2021review, zhang2025poly}, reference‐based claim checking~\citep{ortega2025sciclaims, kumar2025sciclaimhunt}, or computer‐science disciplines alone~\citep{siegel2024core, dycke2022nlpeer, baumgartner2025peerqa}. This limits the potential applicability of the proposed benchmarks as evaluation tools for verification systems in AI-driven science research.

% Prior work on automated review typically draws on data from leading machine learning venues~\citep{dycke2022nlpeer, gao2024reviewer2, du2024llms} and evaluates models based on how well they replicate human critiques using ROUGE-L or semantic-similarity metrics~\citep{lin2023moprd, zeng2024scientific}. These efforts concentrate on high-level attributes~\citep{kang2018dataset}, overlooking deep, substantive flaws that can invalidate results and trigger errata or retractions. They also focus exclusively on computer science, leaving the vast majority of scientific fields untouched.

In this paper, we introduce \benchmark{} (\textbf{\underline{S}}cientific \textbf{\underline{P}}aper Err\textbf{\underline{o}}r De\textbf{\underline{t}}ection), a complex multi-modal academic error verification benchmark, comprising 83 up-to-date manuscripts spanning ten scientific fields with multiple \emph{human-annotated} errors. Given large-scale multi-modal inputs with 12,000 text tokens and 18 images on average, multi-modal LLMs (MLLMs) are tasked with generatively identifying more than one error with varying difficulties in a single paper: \eg, factual inconsistencies, figure duplications, and mathematical errors. We only select papers published \emph{from} 2024, minimizing the potential contamination with parametric knowledge during evaluation \citep{bejan-etal-2023-make}. It should be noted that, whereas prior evaluation suites focus on sentence‐level fact checks of everyday knowledge~\citep{thorne-etal-2018-fever, wadden-etal-2020-fact} or on reproducing noisy peer‐review feedback~\citep{lin2023moprd, shin2025automatically}, \benchmark{} extends verification to the full complexity of frontier-level scientific research. This paper is mainly divided into three parts.

\begin{enumerate}[left=1pt]
    \item \textbf{\benchmark{} Benchmark Design Principles} (Section \ref{sec:benchmark}): We detail our efforts of multiple automated filtering, author verifications, and human annotations, highlighting our commitment to include confirmed, noncontroversial errors across diverse scientific subdomains.
    
    \item \textbf{Model Evaluation and Analysis} (Section~\ref{sec:results}) We present evaluation results, demonstrating that even the state-of-the-art models struggle on \benchmark{}. Specifically, OpenAI's o3 \citep{openai2025o3o4mini} and Llama-4-Maverick \citep{meta2025llama4} achieved 18.4\% and 0.9\% at $\mathrm{pass}@1$ to name a few. Furthermore, model confidence approaches zero when repeated over eight independent trials, questioning their reliability. We also observe that proprietary reasoning models suffer in detecting figure-related errors, highlighting shortcomings in their multi-modal capabilities. Such results cast serious concerns, revealing significant gap between current AI capabilities and the demands of rigorous scientific verification.

    \item \textbf{Expert-led Case Studies} (Section~\ref{sec:case_study}) We present expert-led case studies in mathematics and materials science, analyzing model outputs to diagnose their failures. Our observations show that models struggle with long-tail knowledge likely absent in web data and extremely long contexts. We also note that, without fully spelled-out derivations, models fail to understand some calculations and overlook domain-specific conventions, making student-like errors. 
    % This highlights multiple avenues of potential improvement for future research.
\end{enumerate}

\begin{figure}[!t]
    \centering
    \includegraphics[width=\linewidth]{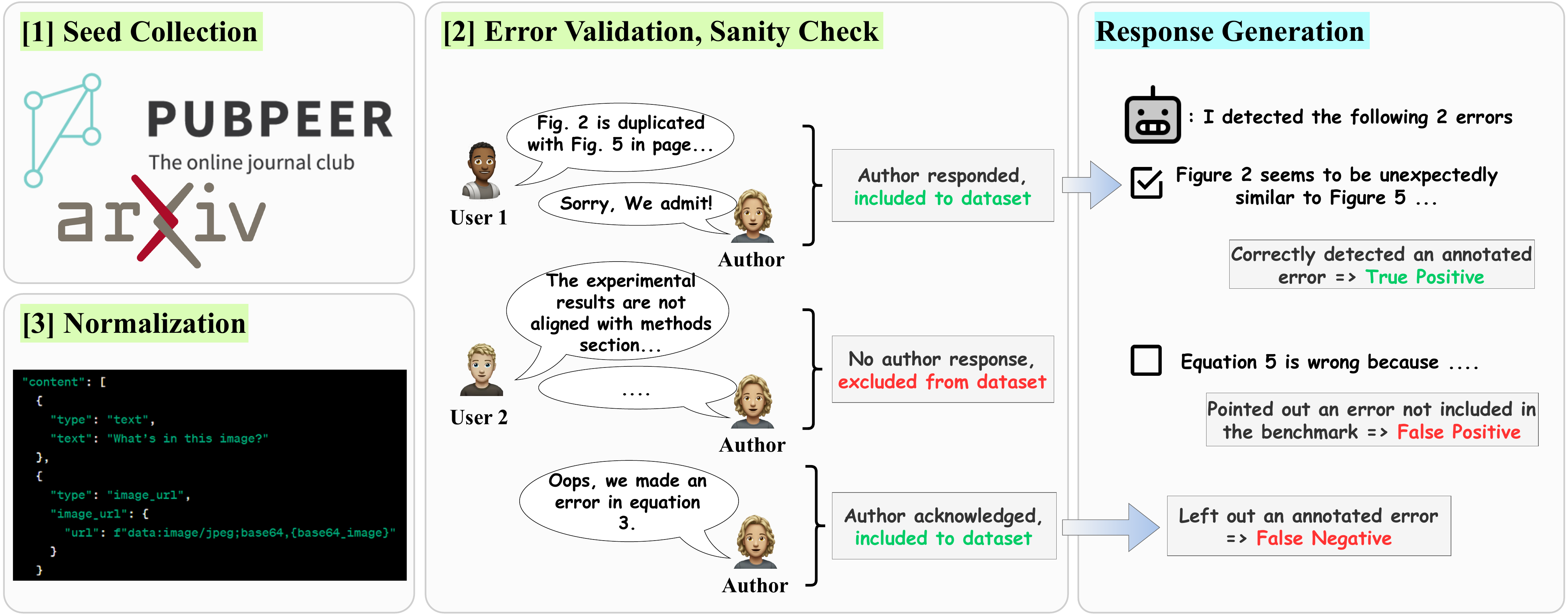}
    \vspace{-0.1in}
    \caption{\footnotesize \textbf{Overview of \benchmark.} Green indicates benchmark construction process, from seed collection through validation to normalization; blue indicates evaluation, where LLM outputs are compared to ground-truth errors and classified as true positives, false positives, or false negatives.
}
    \vspace{-0.1in}
    \label{fig:benchmark}
\end{figure}

%% file: section/benchmark.tex
In this section, we introduce a detailed overview of \benchmark{}, a complex multi-modal academic verification benchmark with cross-validated scientific manuscripts. We ensure credibility in the error annotations through a cross-validation process between human experts in each field and proprietary language models (Section \ref{subsec:construction}). Spanning over ten different fields and six error types (Section \ref{subsec:stats}), we introduce evaluation protocols mainly based on precision, recall, and $\mathrm{pass}@K$ (Section \ref{subsec:protocol}). 
% Figure~\ref{fig:benchmark} illustrates the overall process.

\subsection{Data Curation}\label{subsec:construction}

\paragraph{Stage 1 - Seed Collection}~We source our seed manuscripts from two major repositories: (1) \textsc{WithdrarXiv}~\citep{rao2024WithdrarXiv} and (2) PubPeer\footnote{\url{https://pubpeer.com/}}. First, we extract entries annotated as “factual/methodological/other critical errors” from \textsc{WithdrarXiv}, a dataset of 14,000 papers and their associated retraction comments. Second, we crawl PubPeer, an anonymous post-publication peer review website, where users flag methodological flaws, image manipulations, and other scientific concerns. Following \cite{ortega2022classification}, we query initial searches using alphabets, extract high-frequency keywords from the returned paper titles, re-query using those keywords, and scrape each paper’s metadata (title, authors, venue) alongside the entire comments. We briefly attempted to include medRxiv and bioRxiv, but having retrieved only 1 and 13 papers, respectively, we dropped them due to the low yield.

\paragraph{Stage 2 - Automated Filtering}~We apply two GPT-4o \citep{openai2024gpt4ocard}\footnote{Using version gpt-4o-2024-08-06} filtering passes. The first retains comment–manuscript pairs that unambiguously pinpoint a specific section, figure, equation, or table, reducing our pool to 1,855 \textsc{WithdrarXiv} and 25,378 PubPeer samples. The second pass removes reports that require external artifacts (e.g., duplicated images across papers or errors detectable only via external datasets or code). Finally, to avoid overlap with GPT-4o’s training cutoff
% and focus on recent work
, we filter for papers published after 2024, yielding 58 \textsc{WithdrarXiv} and 215 PubPeer samples.

\paragraph{Stage 3 - Error Validation by Original Authors}~For remaining manuscripts, we only retain those the \emph{original authors directly confirmed}. Specifically, we only retain PubPeer comments followed by an explicit author response acknowledging the mistake and treat \textsc{WithdrarXiv} self‐retractions as definitive evidence of a critical error. In all cases where the author themselves admits the problem, we take this acknowledgment as confirmation of a genuine error. While some errors may appear to be evident, we do not include any error with explicit acknowledgment from the original authors, as many of the work cover ungoing areas of research, which remain unsettled in the scientific discourse.

% \emph{How can we guarantee that flagged issues reflect genuine errors rather than unsettled scientific debate?} Recent scientific research may include provisional or controversial claims, making the notion of a “universal fact” inherently ambiguous; accordingly, we require direct author confirmation for every flagged issue. Specifically, we only retain PubPeer comments followed by an explicit author response acknowledging the mistake, and we treat \textsc{WithdrarXiv} self‐retractions as definitive evidence of a critical error. In all cases where the author themselves admits the problem, we take this acknowledgment as confirmation of a genuine error.

\paragraph{Stage 4 - Sanity Check from Human Annotators}~We further apply a two-stage human validation with mutually exclusive annotators. First, with part of the authors as human annotators, we manually validate if remaining flagged issues fulfill three conditions: (1) self-contained, (2) identifiable, and (3) explicitly acknowledged by the original authors. For those which satisfy the conditions, We retrieve the archived PDF to verify that the error remains visible, then document a concise description of the problem, quote the author’s acknowledgement verbatim, and assign both an error category and a severity rating—proxied by the form of the author’s response (erratum versus retraction). Afterwards, the second group conducted a comprehensive audit of all annotations to ensure consistent application of these standards. The final \benchmark{} benchmark comprises 83 manuscripts with 91 annotated errors. Although modest in size, our dataset aligns with recent trends toward compact, high-quality benchmarks: MT-Bench (80 items)~\citep{zheng2023judging}, GPQA-D (198 items)~\citep{rein2024gpqa}, AIME 2024/2025 (30 items each)~\citep{maa2024aime}, USAMO 2025 (6 items)~\citep{petrov2025proof} and PaperBench (20 items)~\citep{starace2025paperbench}.

% In Stage 1, an author confirms that each flagged issue is self-contained, unambiguously identifiable, and explicitly acknowledged by the original authors. We retrieve the archived PDF to verify that the error remains visible, then document a concise description of the problem, quote the author’s acknowledgement verbatim, and assign both an error category and a severity rating—proxied by the form of the author’s response (erratum versus retraction). In Stage 2, one author who didn't participate in Stage 1 conducted a comprehensive audit of all annotations to ensure consistent application of these standards. The final \benchmark{} benchmark comprises 83 manuscripts with 91 annotated errors. Although modest in size, our dataset aligns with recent trends toward compact, high-quality benchmarks: MT-Bench (80 items)~\citep{zheng2023judging}, GPQA-D (198 items)~\citep{rein2024gpqa}, AIME 2024/2025 (30 items each)~\citep{maa2024aime}, USAMO 2025 (6 items)~\citep{petrov2025proof} and PaperBench (20 items)~\citep{starace2025paperbench}.
% \stella{It took me a bit to understand what ``author reviewer'' meant. I would say ``In Stage 1, an author confirmed...'' and then ``in Stage 2, an author who didn't participate in Stage 1 conducted...''}

\paragraph{Stage 5 - Normalization}~We normalize final manuscripts in PDF format into text and image sets to best evaluate the comprehension of the target models. While prior benchmarks in manuscript error detection~\citep{baumgartner2025peerqa} and AI-assisted science~\citep{seo2025paper2code} have relied on raw PDFs or text-only inputs, this approach offloads document understanding to OCR and parsing modules rather than the LLM itself, thereby conflating upstream parser failures with downstream model errors. Instead, we process all the documents for usage. We first employ Llama-Parse\footnote{\url{https://www.llamaindex.ai/llamaparse}} to convert each PDF into Markdown and capture high-fidelity screenshots of every figure, table, and equation. In pilot experiments, OCR failures, particularly in mathematical expressions, led downstream models to misinterpret formatting artifacts as errors. To address this, we introduce a refinement stage. For each page, the initial OCR text and screenshots (one full-page image plus isolated equations and paragraphs, roughly eight images per page) are sent to GPT-4.1 for correction. Finally, we conduct a manual audit of all processed pages to ensure that every flagged error remains visible and accurately represented in the OCR output.

% \begin{figure}[t]
%     \centering
%     \includegraphics[width=\textwidth]{figures/model_detection_rate_heatmap.png}
%     \caption{\footnotesize place holder. summarizes the 91 errors we’ve annotated across 83 manuscripts, breaking them down by error type, research domain, severity, manuscript length, and figure count.}
%     \label{fig:sdfs}
% \end{figure}

\subsection{Benchmark Statistics}\label{subsec:stats}

\input{tables/types}

% \begin{itemize}
% \item \textbf{Data inconsistency:} mismatched values between text, tables, and figures
% \item \textbf{Equation/proof:} incorrect derivations
% \item \textbf{Experiment Setup:} missing controls or misreported protocols
% \item \textbf{Figure duplication:} reused or manipulated images
% \item \textbf{Reagent-identity:} mislabeled or incorrect materials
% \item \textbf{Statistical-reporting:} misused statistical values or inappropriate tests
% \end{itemize}

\paragraph{Error Types} We derive the six categories in Table~\ref{tab:general_stats} inductively from our annotations rather than setting a priori. As we review each error, we group similar cases. This is to capture the true distribution of errors existing in manuscripts. During this process, figure-duplication instances initially overwhelmed the dataset, so we filtered based on severity and paper category to prevent a single type from dominating.

\begin{wrapfigure}{r}{0.45\textwidth}
    \vspace{-0.2in}
    \centering
    \includegraphics[width=\linewidth]{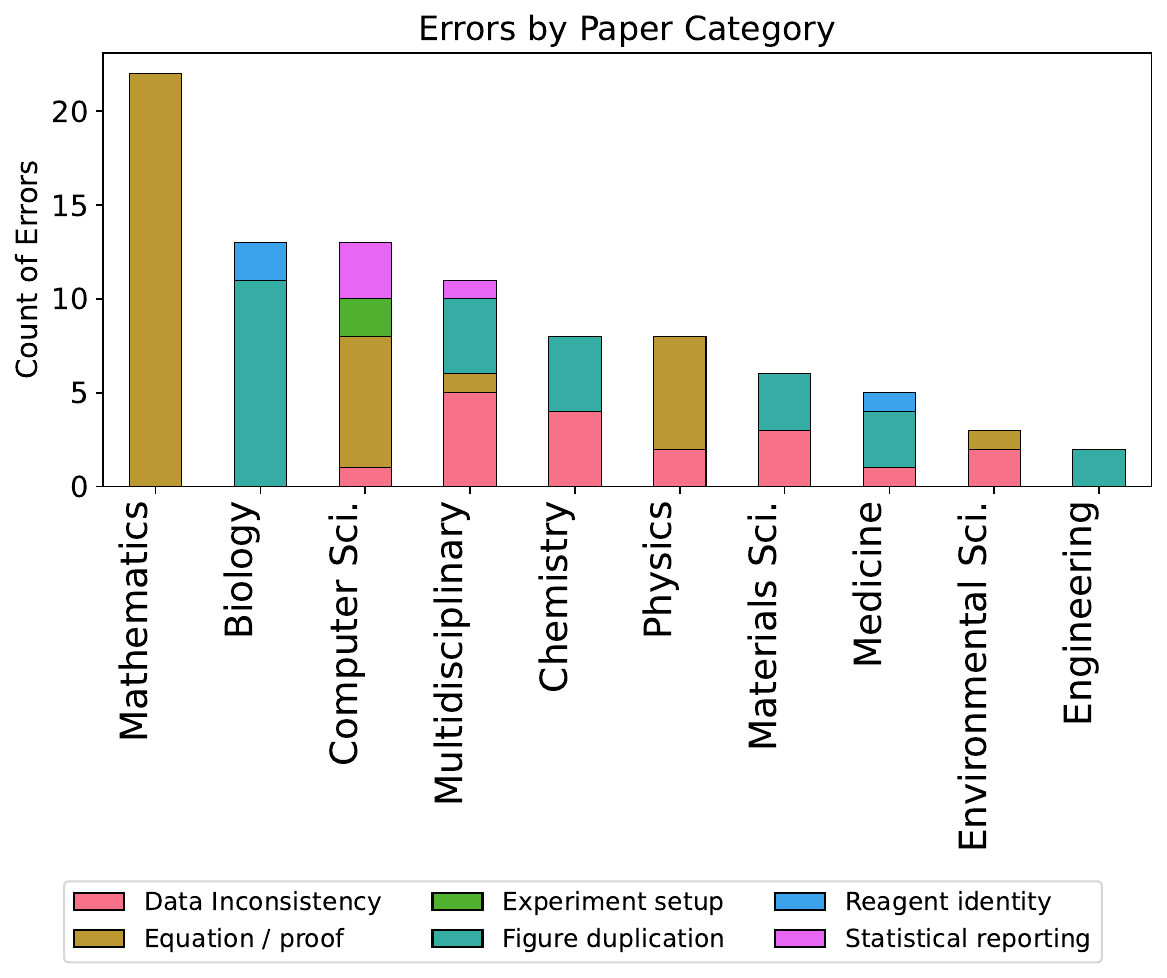}
    \vspace{-0.1in}
    \caption{\footnotesize \textbf{Distribution of annotated errors by research domain and error type.}}
    \vspace{-0.2in}
    \label{fig:error_types}
\end{wrapfigure}

\paragraph{Paper Subjects} We present general statistics in Table~\ref{tab:general_stats}. We classify each paper into ten research domains: Mathematics, Physics, Biology, Chemistry, Materials Science, Medicine, Environmental Science, Engineering, Computer Science, and  Multidisciplinary, based on its journal venue or arXiv subject. In Figure~\ref{fig:error_types}, we observe clear domain patterns: mathematics, computer science, and physics papers skew toward equation/proof flaws; biology toward figure-duplication. 76 manuscripts out of 83 contain a single error, six contain two, and one paper has the maximum of three annotated errors. We proxy error severity by the authors’ post-publication response: 59 errors were addressed via errata, while 32 led to full retractions. Retractions are concentrated mostly in equation/proof cases. Manuscripts span 1k–46k tokens and include 0–80 figures, creating a long context, multimodal, and figure-rich benchmark far exceeding the scale and complexity of existing error-detection datasets. Although longer papers tend to include more figures, the relationship is weak (Pearson’s $r = 0.19$), highlighting diverse presentation styles across fields.

\subsection{Evaluation protocol}\label{subsec:protocol}

We provide the full paper as interleaved text and image data, followed by the prompt to return every error with each error’s location (section, figure, equation, or table), accompanied by a description. The output is prompted to be a structured JSON format (see Appendix~\ref{app:details_on_eval} for an example).

\paragraph{Evaluation Metric} We mainly evaluate verification performance through precision, recall, and $\mathrm{pass}@K$. A predicted error is counted as a true positive (TP) only when the model’s reported location matches a benchmark annotation and an LLM confirms they indicate the same error\footnote{GPT-4.1 is used to compare predicted error descriptions against benchmark annotations as a similarity check~\citep{ni2024mixeval}; the LLM does not evaluate the errors’ correctness or severity.}. All other predictions, including those at non-annotated locations or those matching an annotated location but with a different description, are considered false positives (FP), and any benchmark annotation the model fails to predict is a false negative (FN). We treat the error annotations included in \benchmark{} as exhaustive: any model-reported error not matching an annotation is counted as a false positive. Although models could, in principle, flag genuine errors outside our annotations, through case studies later in this paper, we notice such cases are highly unlikely. To summarize model performance, we report Precision and Recall: 
\begin{equation}
    \mathrm{Precision}
    =\frac{\mathrm{TP}}{\mathrm{TP}+\mathrm{FP}},\qquad
    \mathrm{Recall}
    =\frac{\mathrm{TP}}{\mathrm{TP}+\mathrm{FN}}.
\end{equation}
\textit{Precision} quantifies the proportion of the model’s flagged errors that match benchmark annotations, penalizing unexpected predictions, and is most appropriate when false positives impose significant review overhead or undermine confidence in the tool’s outputs. \textit{Recall} quantifies the proportion of annotated errors the model successfully identifies, penalizing missed detections. In practice, users concerned about model hallucinations or the impact of unannotated flags should focus on Precision. In contrast, those seeking comprehensive error coverage, or who doubt the exhaustiveness of our annotations, should emphasize Recall.

Following \cite{kulal2019spoc} and \cite{chen2021evaluating}, to capture how error detection improves with multiple attempts, for \(K\) runs per paper, we define
\begin{equation}
    \mathrm{pass}@K
    = \frac{1}{\sum_{i=1}^N |G_i|}
      \sum_{i=1}^N \sum_{g \in G_i}
        \mathbf{1}\!\Bigl[\exists\,s\in\{1,\dots,K\}:\,g \in p_i[s]\Bigr],
\end{equation}

where $G_i$ is the set of annotated errors in paper $i$ and $p_i[s]$ the set predicted in the $s$-th run.  With 83 papers and 91 total errors we generate $N=8$ independent runs per paper.  For each $\mathrm{pass}@K$ we draw $K$ runs without replacement from the eight, repeat this resampling $B=1000$ times, and report the mean and standard deviation of the resulting bootstrap distribution for $K\in\{1,4\}$.

 % \stella{Pass @ \emph{K} is valuable in a context where...}

%% file: tables/types.tex
\begin{table}[]
\centering
\fontsize{8}{9.5}\selectfont
\caption{\footnotesize \textbf{Overview of \benchmark{}.} \textit{Left}: High-level statistics—83 manuscripts, 91 errors from 47 paper sources; tokens per manuscript (mean $\pm$ std., range) and images per manuscript (mean $\pm$ std., range). All token counts were computed using the GPT-4o~\citep{hurst2024gpt} tokenizer from \texttt{tiktoken}~\citep{openai2025tiktoken}. \textit{Right}: Six error categories with concise descriptions and instance counts in parentheses.
 }
\vspace{.1in}
\label{tab:general_stats}
\begin{tabular}{l||ll}
\toprule
\textbf{Benchmark Statistics} & \textbf{Category} & \textbf{Descriptions} \\
\midrule
\textit{\textbf{General}} & \multirow{2}{*}{Equation / proof (37)} & \multirow{2}{*}{Incorrect mathematical derivations} \\
\quad Total Manuscripts: $83$ &  &  \\
\quad Total Errors: $91$ & \multirow{2}{*}{Figure duplication (27)} & \multirow{2}{*}{Reused or manipulated images} \\
\quad Total Paper Sources: $47$ &  &  \\
\textit{\textbf{Tokens}} & \multirow{2}{*}{Data inconsistency (18)} & \multirow{2}{*}{Mismatched values between text, tables, and figures} \\
\quad Avg (std): $12,887_{{\scriptsize 7,421}}$ &  &  \\
\quad Max / Min: $46,441 / 1,207$ & \multirow{2}{*}{Statistical Reporting (4)} & \multirow{2}{*}{Misused statistical values or inappropriate tests} \\
\textit{\textbf{Images}} &  &  \\
\quad Avg (std): $17.5_{{\scriptsize 20.1}}$ & \multirow{2}{*}{Reagent Identity (3)} & \multirow{2}{*}{Mislabeled or incorrect materials} \\
    \quad Max / Min : $80 / 0$  &  &  \\
\textit{\textbf{Error severity}} & \multirow{2}{*}{Experiment Setup (2)} & \multirow{2}{*}{Missing controls or misreported protocols} \\
\quad Errata / Retract: $59 / 32$ &  &  \\ 
\bottomrule
\end{tabular}
\end{table}

%% file: section/results.tex
In the following sections, we evaluate six proprietary models: OpenAI o3~\citep{openai2025o3o4mini}, GPT-4.1~\citep{openai_gpt41}, Google Gemini 2.5 Pro~\citep{google2025gemini2_5_pro}, Gemini 2.0 Flash Lite~\citep{google_cloud_gemini_flash_lite}, Anthropic Claude 3.7 Sonnet:Thinking~\citep{anthropic_claude3_7_sonnet}, and Claude 3.7 Sonnet and four open models: Qwen 2.5-VL-72B/32B-Instruct~\citep{bai2025qwen2}, and Llama 4 Maverick/Scout~\citep{meta2025llama4}. We select the most capable models per family and observe that these models already score near zero in \benchmark{}. Accordingly, as smaller models are unlikely to perform any better, we do not include them in our evaluations. All models are accessed via APIs, and each call is retried up to three times; those that still fail or are cut off due to length limits are marked incorrect.

\input{tables/main}
\subsection{Main Results}

\begin{wrapfigure}{r}{0.45\textwidth}
    \vspace{-0.2in}
    \centering
    \includegraphics[width=\linewidth]{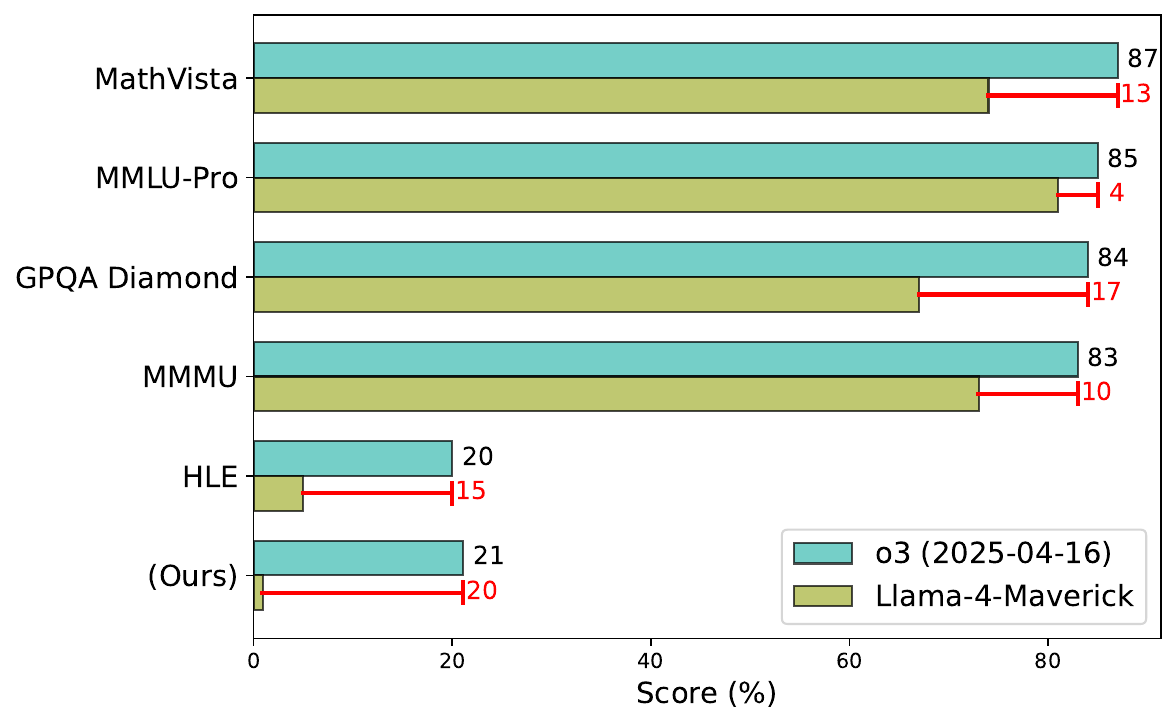}
    \caption{\footnotesize \textbf{Performance of o3 and Llama-4-Maverick across six challenging STEM benchmarks.} The short red horizontal lines mark the gap $\Delta = \text{o3} - \text{Llama-4-Maverick}$ for each benchmark.}
    \vspace{-0.2in}
    \label{fig:benchmark_perf}
% \vspace{-7mm}
\end{wrapfigure}

Table~\ref{tab:main-results} compares ten multi-modal LLMs on \benchmark{}. o3 achieves the highest scores, with 6.1\%\,$\pm$\,1.3 precision, 21.1\%\,$\pm$\,4.4 recall, and a 37.8\%  \(\mathrm{pass}@4\). It is followed by Gemini-2.5-Pro (3.1\%, 10.1\%, 25.9\%), Claude-3.7-Sonnet:Thinking (3.0\%, 6.0\%, 18.6\%), and GPT-4.1 (2.8\%, 6.0\%, 17.8\%).  The lighter proprietary variants, Gemini-2.0-Flash-Lite and the non-Thinking Claude-3.7-Sonnet, score marginally above zero. Surprisingly, open-source models such as Qwen2.5-VL-72B-Instruct and Llama-4-Maverick, which match proprietary models on existing multi-modal benchmarks like MMMU~\citep{yue2024mmmu} or MathVista~\citep{lu2023mathvista}, perform far worse on \benchmark{}. As shown in Figure~\ref{fig:benchmark_perf}, (1) the performance gap between o3 and Llama-4-Maverick is widest on \benchmark{} (ours) ($\Delta=20.2\ \mathrm{pp}$), and (2) \benchmark{} is the only benchmark where Llama-4-Maverick’s score collapses to near zero (0.9 \%). While neither proprietary nor open-source models fully satisfy the requirements of practical deployments of error-detecting AI systems, open-source models lag far behind in domain-specific rigor and robust error-detection capabilities essential for scientific applications.

% \stella{This paragraph makes me wonder what the performance of unimodal text models are. Also, if we claim the gap is much larger than other benchmarks we should demonstrate that in a figure or table. Currently assessing that claim requires looking at data not found in this paper.}

\paragraph{A New Challenging Benchmark for STEM.}

Figure~\ref{fig:benchmark_perf} illustrates the performance of o3 on six benchmarks: MathVista~\citep{lu2023mathvista}, MMLU-Pro~\citep{wang2024mmlu}, GPQA Diamond~\citep{rein2024gpqa}, MMMU~\citep{yue2024mmmu}, HLE~\citep{phan2025humanity} and \benchmark{} (recall). o3 exceeds 80\% on the first four benchmarks, demonstrating robust general reasoning and code understanding. However, performance drops to roughly 20\% on HLE, a curated set of frontier, research-level academic questions, and remains similarly low on \benchmark{} (21.1\%). This drop in performance underscores the difficulty of spotting errors in lengthy scientific text and figures.

\paragraph{Reasoning Models Excel at Equations but Falter on Figures}

\begin{figure}[!t]
    \centering
    \includegraphics[width=\linewidth]{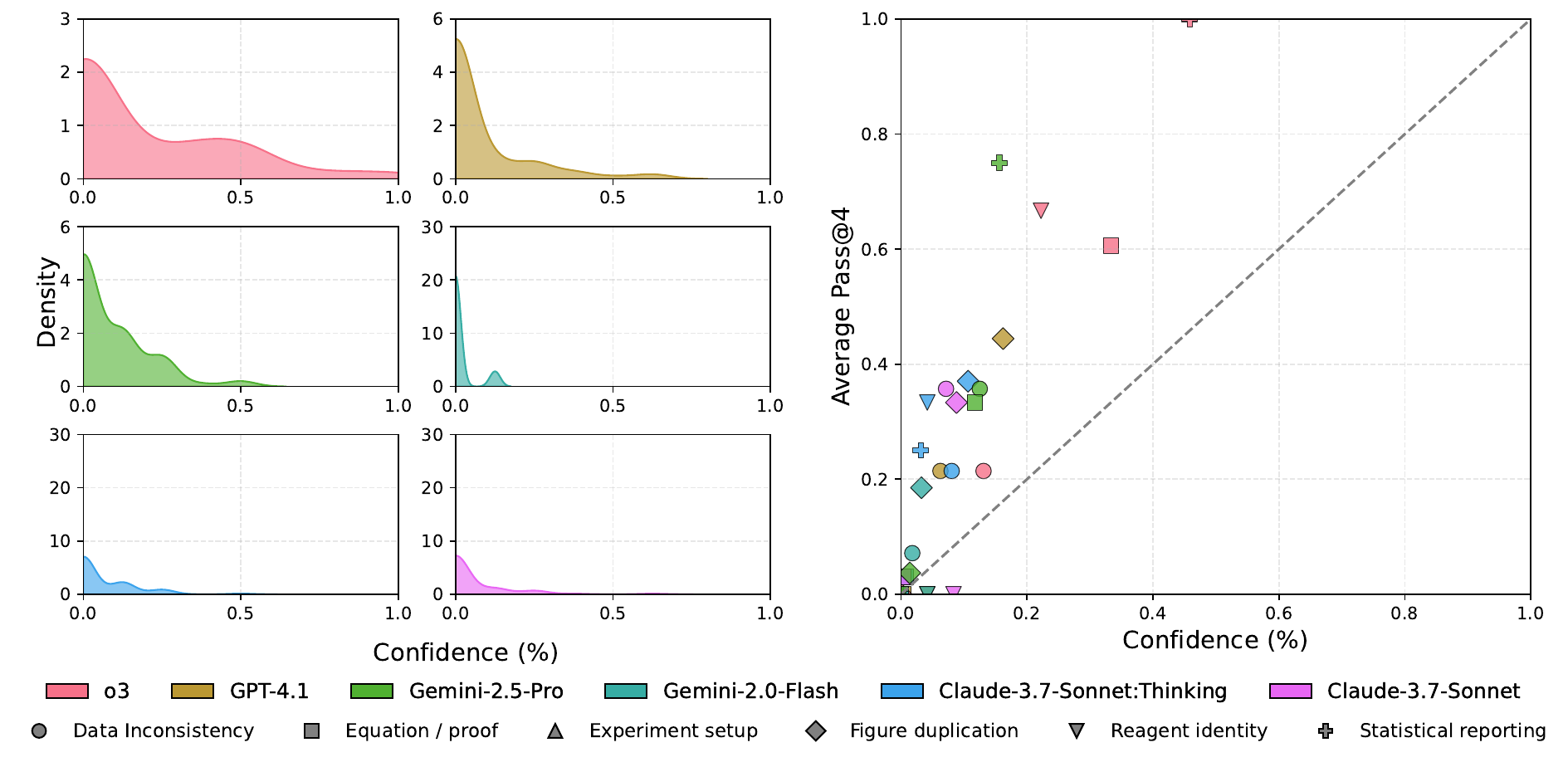}
    \caption{\footnotesize \textbf{Category-specific performance and calibration of six LLMs on \benchmark{}.} \textit{Left}: Kernel density estimates of each model’s reported confidence; all six models predominantly express very low confidence. \textit{Right}: Scatter plot of mean reported confidence (see Appendix~\ref{app:confidence_for_passk} for further details) versus $\mathrm{pass}@4$ for each model (color), broken down by error type (shape). The dashed diagonal marks perfect calibration.
}
\vspace{-5mm}
    \label{fig:confidence}
\end{figure}

The right panel of Figure~\ref{fig:confidence} presents the performance of six models across each category. In the Equation/Proof category, o3 leads with a 62.6\% (\(\mathrm{pass}@4\)), followed by Gemini-2.5-Pro at 36.4\%, while all other models remain below 5\%, underscoring o3’s superior mathematical reasoning. Surprisingly, GPT-4.1 achieves a 44.4\% in the Figure Duplication category, outperforming Claude-3.7-Sonnet\:Thinking (33.3\%), o3 (0\%), and Gemini-2.5-Pro (0\%), revealing a weakness in figure analysis in reasoning models. 
% \stella{This is a very cool result.}

\subsection{Unreliability of Miscalibrated Models.} 

Alongside \(\mathrm{pass}@4\), calibration~\citep{guo2017calibration, ovadia2019can} indicates how much we should trust a model’s predictions. In error detection, where false positives can incur substantial time and labor, knowing when to trust a model is crucial. For each error category, we assess calibration by comparing the model’s actual performance, measured as its average $\mathrm{pass}@4$ rate, with its self-estimated confidence. For details on how the confidence is derived see Appendix~\ref{app:confidence_for_passk}.  % \stella{How does this assess calibration? Also, I don't see this in Figure 5.}

However, Figure~\ref{fig:confidence} (right) shows that confidence correlates only weakly with \(\mathrm{pass}@4\), and the left panel reveals that most models report very low confidence, clustering near zero. Across 498 model–instance evaluations (83 instances × six models), we observe only two cases (both from o3) of full confidence, highlighting the widespread difficulty of reliably detecting errors in scientific manuscripts. These findings demonstrate substantial variability across categories and reaffirm that current LLMs remain unreliable for scientific error detection. 

% \stella{How is confidance calculated? Make sure to include that explicitly. Also, what's the correct way to compute confidance for pass@k? I'm not sure, maybe we should look this up.}

\subsection{Impact of Multi-Modality in Detecting Scientific Errors}

To isolate the impact of visual inputs, we create a text-only subset of \benchmark{} by removing all instances from the figure-duplication and any data-inconsistency category that necessitate figures for comprehension. This yields 48 instances in which errors can be detected using text alone. Table~\ref{tab:multi_modality} compares model performance on the selected instances under multimodal and text-only conditions. The left panel reports each model’s accuracy on these 48 cases with figures included (extracted from the runs of Table~\ref{tab:main-results}); the right panel shows performance after stripping out all figures. In the text-only setting, we add three unimodal LLMs: DeepSeek-R1~\citep{guo2025deepseek}, DeepSeek-V3~\citep{liu2024deepseek}, and Qwen3-235B-A22B~\citep{yang2025qwen3technicalreport}.

\input{tables/modailty}

% \stella{How often does a human need the image to decide? Is the image supporting general understanding or is it crucial to understanding the errors?}

We observe two key findings. First, most models improve in recall and \(\mathrm{pass}@4\) when removing images, suggesting that figures usually act as distractors rather than helpful context. The exceptions are o3 and Gemini-2.5-Pro, which see a modest drop without visual inputs. This indicates that they have been leveraging figures to understand the paper rather than treating them as mere auxiliary signals. Second, the divide between proprietary and open models is vast in the multi-modal setting, proprietary systems maintain substantial recall (e.g., o3 at 34.6 \%, Gemini-2.5-Pro at 13.7 \%) and \(\mathrm{pass}@4\), whereas open-source models collapse to near zero. Under text-only conditions, while proprietary systems still lead, the performance margin is comparably smaller. 
% \stella{Don't assume the reader knows which are openly licensed models.}

%% file: tables/main.tex
\begin{table}[t!]
\centering
\fontsize{8}{9.5}\selectfont
\caption{\footnotesize \textbf{Performance of ten models on the \benchmark{} dataset.} The Think column denotes the use of test-time scaling. Precision, Recall, $\mathrm{pass}@1$ and $\mathrm{pass}@4$ (all in \%) are reported as mean and standard deviation (in parentheses) over eight independent trials. The highest value in each column is \textbf{bolded}, and the second-highest is \underline{underlined}. Detailed evaluation results are available in Appendix~\ref{app:detailed_resutls}.}
\vspace{0.1in}
% \stella{Some of these have only one number in subscript, others have two. What's the difference?}
% The colored squares denote each model’s features:
% $ \coloredblock{prop}$  Proprietary; $ \coloredblock{think} $ Thinking mode enabled; $ \coloredblock{text}$ Text Only.}
\label{tab:main-results}
\begin{tabular}{lccccc}
\toprule
\textbf{Models} & \textbf{Think} & \textbf{Precision (\%)} & \textbf{Recall (\%)} & \textbf{pass@1 (\%)} & \textbf{pass@4 (\%)} \\
\midrule
o3 {\scriptsize (2025-04-16)}                     & \cmark & $\mathbf{6.1_{1.3}}$  & $\mathbf{21.1_{4.4}}$  & $\mathbf{18.4_{2.1}}$ & $\mathbf{37.8_{1.8}}$ \\
GPT-4.1 {\scriptsize (2025-04-14)}                 & \xmark & $2.8_{0.8}$            & $6.0_{1.6}$            & $6.6_{1.7}$ & $17.8_{1.5}$         \\
Gemini-2.5-Pro {\scriptsize (preview-03-25)}       & \cmark & $\underline{3.1_{1.7}}$            & $\underline{10.1_{5.6}}$           & $\underline{7.8_{3.8}}$ & $\underline{25.9_{4.0}}$         \\
Gemini-2.0-Flash-Lite {\scriptsize (001)}         & \xmark & $1.0_{0.8}$            & $1.6_{1.3}$            & $1.5_{1.0}$ & $6.0_{1.5}$            \\
Claude-3.7-Sonnet {\scriptsize (20250219:Think)}   & \cmark & $3.0_{1.3}$            & $6.0_{2.4}$            & $5.5_{1.7}$ & $18.6_{2.8}$         \\
Claude-3.7-Sonnet {\scriptsize (20250219)}         & \xmark & $3.2_{1.5}$            & $5.8_{2.7}$            & $4.5_{1.9}$ & $14.1_{1.6}$         \\
\midrule
Qwen2.5-VL-72B-Instruct                            & \xmark & $0.6_{1.2}$            & $0.4_{0.7}$            &  $0.4_{0.6}$ & $1.7_{1.0}$            \\
Qwen2.5-VL-32B-Instruct           & \xmark &  $1.9_{2.0}$ &   $1.9_{1.7}$     & $2.0_{1.5}$ &  $5.6_{1.6}$ \\
Llama-4-Maverick   & \xmark &     $2.0_{2.6}$      &   $0.9_{1.2}$   & $0.9_{1.0}$ &  $3.3_{1.2}$   \\
Llama-4-Scout           & \xmark &  $0.8_{1.0}$ & $1.9_{2.3}$  & $1.8_{2.0}$ &  $7.2_{3.1}$   \\
\bottomrule
\end{tabular}
\end{table}

%% file: tables/modailty.tex
\begin{table}[t!]
  \centering
  \fontsize{8}{9.5}\selectfont
  \caption{\footnotesize \textbf{Multi-modality ablation for 13 models}: recall and pass@4 (in \%) are reported as mean (std) over eight independent trials. The left panel shows each model’s performance with multi-modal inputs; the right panel shows performance on the text-only subset of \benchmark{} (48 figure-independent instances), including additional unimodal LLMs (DeepSeek-R1, DeepSeek-V3, Qwen3-235B-A22B). The highest value in each column is \textbf{bolded}, and the second-highest is \underline{underlined}. Detailed evaluation results are available in Appendix~\ref{app:detailed_resutls}.}

  \vspace{0.1in}
  \label{tab:multi_modality}
  \begin{tabular}{l c c c c c}
    \toprule
      & & \multicolumn{2}{c}{\textbf{Multi-Modal}} 
        & \multicolumn{2}{c}{\textbf{Text-Only}} \\
    \cmidrule(r){3-4}\cmidrule(l){5-6}
    \textbf{Models} 
      & \textbf{Think}
      & \textbf{Recall (\%)} & \textbf{pass@4 (\%)} 
      & \textbf{Recall (\%)} & \textbf{pass@4 (\%)} \\
    \midrule
    o3 {\scriptsize (2025-04-16)}                       & \cmark & $34.6_{7.1}$ & $61.1_{2.9}$ & $25.7_{7.1}$ & $56.2_{4.2}$ \\
    GPT-4.1 {\scriptsize (2025-04-14)}                  & \xmark & $0.5_{0.9}$  & $2.0_{1.4}$  & $8.4_{2.5}$  & $19.8_{2.7}$ \\
    Gemini-2.5-Pro {\scriptsize (preview-03-25)}        & \cmark & $13.7_{8.6}$ & $34.8_{6.1}$ & $6.9_{3.1}$  & $17.0_{2.8}$ \\
    Gemini-2.0-Flash-Lite {\scriptsize (001)}          & \xmark & $0.4_{0.9}$  & $2.1_{1.3}$  & $1.9_{1.4}$  & $8.0_{1.8}$  \\
    Claude-3.7-Sonnet {\scriptsize (20250219:Think)}   & \cmark & $2.9_{2.3}$  & $8.5_{1.7}$  & $5.0_{2.3}$  & $17.0_{3.1}$ \\
    Claude-3.7-Sonnet {\scriptsize (20250219)}         & \xmark & $1.9_{2.1}$  & $4.8_{1.5}$  & $5.8_{3.1}$  & $15.2_{2.8}$ \\
    \midrule
    DeepSeek-R1                                         & \cmark & –            & –            & $14.8_{3.8}$ & $38.6_{3.3}$ \\
    DeepSeek-V3 {\scriptsize (0324)}                    & \xmark & –            & –            & $1.9_{1.1}$  & $6.7_{2.1}$  \\
    Qwen3-235B-A22B                                     & \cmark & –            & –            & $15.4_{6.2}$ & $38.2_{3.1}$ \\
    \midrule
    Qwen2.5-VL-72B-Instruct                             & \xmark & $0.0_{0.0}$  & $0.0_{0.0}$  & $4.7_{2.2}$  & $11.2_{2.5}$ \\
    Qwen2.5-VL-32B-Instruct                             & \xmark & $0.4_{0.8}$  & $1.7_{0.9}$  & $1.1_{1.5}$  & $3.0_{1.2}$  \\
    Llama-4-Maverick                                   & \xmark& $0.5_{0.9}$  & $2.1_{1.4}$  & $0.8_{1.0}$  & $3.5_{1.5}$  \\
    Llama-4-Scout                                      & \xmark & $0.4_{0.8}$  & $2.0_{1.4}$  & $1.6_{2.1}$  & $5.9_{2.5}$  \\
    \bottomrule
  \end{tabular}
\end{table}

%% file: section/case_study.tex
We select two withdrawn manuscripts, each from mathematics and materials science, for a qualitative review. A domain expert evaluated each paper, either a researcher with relevant publications or a PhD-trained postdoc in the field. Reviewers are provided the LLM-flagged “errors” from o3 and Gemini 2.5 Pro alongside the official withdrawal notices. They are asked to verify whether the model has missed any benchmarked errors (i.e., true positives mislabeled as false negatives). Moreover, they are required to assess each flagged issue that falls outside our annotations to determine if any presumed false positives correspond to valid flaws. We consulted the original authors to verify the disputed issues whenever a reviewer remained uncertain.\footnote{Due to space constraints, we show only excerpts of model responses and one case study per domain; for the complete results, see Appendix \ref{app:case_study}.}

\subsection{Mathematics}
\citet{petersen2024} studies the configuration spaces of points in algebraic varieties with a multiplicative decomposition, and discusses some applications such as the cohomology of moduli stacks of hyperelliptic curves. It was withdrawn because of a gap that lies in the core arguments of Theorem 1.8 and Theorem 1.13 which invalidates the bulk of the paper.

%Figure~\ref{fig:o3-petersen-response} is a snippet of o3’s feedback targeting Section 3.3. It is wrong because the extra “$-\ell$” is necessary. It is worth noting that this particular Hodge-theoretic interpretation of Eichler-Shimura isomorphism is exceedingly rare in internet data. It may also appear differently than how the Eichler-Shimura isomorphism is presented in common internet sources, whose adaptation requires genuine understanding of multiple subjects, such as variation of Hodge structures and the theory of modular forms. But the way o3 challenged the twist still appears to be unfounded and unexplainable.

Both o3 and Gemini-2.5-Pro exclusively flag issues in Section 3. Ironically, this is the only part of the manuscript {\bf{not}} affected by the actual mathematical gap. o3 criticizes the calculation of $H^k(M_{1,1}, V_\ell)$ in Section 3.3, claiming that the use of Eichler-Shimura isomorphism is wrong without giving too much details. But there is no mistake in this part and the claim is hallucinated. It is worth noting that applying Eichler-Shimura isomorphism in a similar context as the paper is exceedingly rare in internet data. The formula may also appear differently than how the Eichler-Shimura isomorphism is presented in common internet sources, whose equivalence requires genuine understanding of multiple subjects, such as variation of Hodge structures and the theory of modular forms.

\begin{figure}[ht]
  \centering
  \begin{tcolorbox}[colback=gray!10, colframe=blue!5, width=\textwidth,fontupper=\small]
    Eichler--Shimura is quoted as 
    \(H^1(M_{1,1},V_\ell)\cong S_{\ell+2}\oplus\mathbb{Q}(-1-\ell)\),
    but the Tate (Eisenstein) summand should be \(\mathbb{Q}(-1)\), independent of \(\ell\). 
    The extra “\(-\ell\)” is a mis-twist.
  \end{tcolorbox}
  \caption{\footnotesize o3’s feedback on \cite{petersen2024}.}
  \label{fig:o3-petersen-response}
\end{figure}

On the other hand, Gemini-2.5-Pro tackles Theorem 3.8 as shown in Figure \ref{fig:g2.5-petersen-response}. Despite doing reasonable calculations, the model makes a false claim on $H^0(A_{1,1}, V(2,2))(-1)$, resembling mistakes students could make after recollecting a wrong fact.

\begin{figure}[ht]
  \centering
  \begin{tcolorbox}[colback=gray!10, colframe=blue!5, width=\textwidth,fontupper=\small]
    The stated result \(H^1(M_2, V(2,2)) = \mathbb{Q}(-3)\) contradicts the Gysin‐sequence derivation (Sec.~3.1), which yields \(H^1(A_2, V(2,2)) = 0\), together with \(H^0(A_{1,1}, V(2,2))(-1) = \mathbb{Q}(-3)\) (Sec.~3.5) and the non‐vanishing/isomorphism of the Gysin map \(G: H^0(A_{1,1}, V(2,2))(-1)\to H^2(A_2)\), thereby implying \(H^1(M_2, V(2,2)) = 0\).
  \end{tcolorbox}
  \caption{\footnotesize Gemini-2.5-Pro’s feedback on \cite{petersen2024}.}
  \label{fig:g2.5-petersen-response}
\end{figure}

%Interestingly, in both cases, the models exclusively flag issues in Section 3—ironically, the only part of the manuscript {\bf{not}} affected by the actual mathematical gap. Section 3 is also the section containing the most details, and we suspect that LLMs might be drawn to that.

\subsection{Material science}

In our materials science case study, we select \cite{ye2025superacid}, which uses TfOH to synthesize covalent organic frameworks (COFs). However, it mislabels several samples in its figures. These errors are easier to spot than those in the mathematics papers and would be obvious to any attentive reader.

\begin{figure}[ht]
  \centering
  \begin{tcolorbox}[colback=gray!10, colframe=blue!5, width=\textwidth,fontupper=\small]
    Optimal TfOH concentration is stated as 0.006–0.016 M, yet all optimisation and standard syntheses use 0.2 M TfOH. Internal inconsistency invalidates the universality of the claimed optimum.
  \end{tcolorbox}
  \caption{\footnotesize o3’s feedback on \cite{ye2025superacid}.}
  \label{fig:o3-ye-response}
\end{figure}

Errors pointed out by o3 exemplify failures in both long-range context comprehension and multi-hop reasoning. For instance, in Figure~\ref{fig:o3-ye-response}, o3 does not realize that the optimal concentration value reported by the authors (0.006–0.016 M) is the concentration of the final mixture, while the the second value (0.2M) is the concentration of the acid before being added to the final mixture. This misunderstanding likely arises because the optimal concentration in the final mixture is mentioned only once, and the explicit calculation is not shown throughout the manuscript. As a result, o3, having seen references only to the concentration before mixture, fails to infer the relationship between the two values.

\begin{figure}[ht]
  \centering
  \begin{tcolorbox}[colback=gray!10, colframe=blue!5, width=\textwidth ,fontupper=\small]
    (A) There is a contradiction in the indexing of PXRD peaks for TAPPy-TFPPy-COF (Figure 6H). The peaks are initially assigned to facets including (020): 'TAPPy-TFPPy-COF displayed peaks [...] \\ \\
    (B) The BET surface area for the scaled-up TFPPy-PDA-COF is reported as `$1606\ \mathrm{cm}^2\,\mathrm{g}^{-1}$`. The correct unit is `$\mathrm{m}^2\,\mathrm{g}^{-1}$.` This unit error misrepresents the surface area by a factor of 10,000, constituting a fundamental data-reporting mistake.

  \end{tcolorbox}
  \caption{\footnotesize Two of Gemini 2.5 Pro’s feedback on \cite{ye2025superacid}.}
  \label{fig:gemini-ye-response}
\end{figure}

In (A) of Figure~\ref{fig:gemini-ye-response}, Gemini 2.5 Pro seems to make a "reading" mistake, attributing the second facet pair to TAPPy-TFPPy-COF when it in fact describes TAPPy-BPTC-COF. Notably, however, in (B), it notices a potential error in the units, where a certain compound was assigned a surface area 10000x smaller than all the other compounds in the same family. Because the authors do not mention this extreme property of this material, we suspect that this is a real typo. While not severe, \textbf{this error is the only instance} in which we observe an LLM identifying an unannotated but genuine error.

%% file: section/related_works.tex
\paragraph{AI Co-Scientists}

Recent breakthroughs have pushed LLMs to PhD-level performance on STEM benchmarks~\citep{rein2024gpqa}, driving efforts to embed them as generators of the scientific forward pass, encompassing hypothesis generation~\citep{si2024can}, experimental planning~\citep{seo2025paper2code}, and manuscript drafting~\citep{jain2024generative}.  Such systems, or AI Co-Scientists~\citep{lu2024ai, deepmind2025alphaevolve}, employ agent-based pipelines that mirror the stages of scientific research. However, concurrent works often omit a rigorous backward pass or “verification” and instead rely on LLM judges~\citep{zheng2023judging}. Yet, prior studies demonstrate that LLM judges may fail on complex tasks~\citep{son2024llm}, allowing factual and methodological errors to remain undetected. This compromises the reliability of AI-driven research. It should be noted that verifiability has long been central to scaling AI progress: self‐supervised learning employs next‐token prediction as a provable training objective~\citep{jernite2017discourse}; instruction‐tuning leverages LLM-generated instruction–response pairs as reliable signals~\citep{wang2022self}; and reinforcement learning uses verifiable rewards~\citep{guo2025deepseek} for alignment. Likewise, we posit that robust scientific verification must underpin reliable LLM‐driven scientific research. 

\paragraph{Automating Scientific Verification}
Two research strands, fact verification and automated peer review generation, may appear related to \benchmark{}, but each has critical limitations. Prior fact verification benchmarks~\citep{thorne-etal-2018-fever,wadden-etal-2020-fact} concentrate on claims at the sentence level and rely on text inputs only to assess consistency with reference documents. Automated peer review systems draw almost entirely on computer science publications~\citep{gao2024reviewer2, dycke2022nlpeer, baumgartner2025peerqa}, restricting their disciplinary coverage. These approaches measure success by matching past reviews via metrics such as ROUGE~\citep{zeng2024scientific} rather than detecting errors. They also overlook the inherent noise in peer review reports~\citep{cortes2021inconsistency, bonavia2023noise} and seldom apply adequate quality control or validate ground truth. Our work sets apart from previous efforts by applying expert and automated validation to distill only genuine mistakes into \benchmark{}, additionally, we package full, multimodal papers into models at inference mirroring real-world academic verifications.

%% file: section/conclusion.tex
In this paper, we introduce \benchmark{}, a multimodal error‐detection benchmark that captures the full complexity of frontier-level scientific research. Each instance averages 12,000 text tokens and 18 images, posing a significant challenge for current large language models: OpenAI's o3 and Google's Gemini 2.5 Pro achieve $\mathrm{pass}@1$ scores of only 18.4 \% and 7.3 \%, respectively. Our expert‐led case studies further show that these models fall short in long‐tail domain knowledge and implicit multi‐step calculations. Together with the rise of interest in AI Co-Scientists, these results highlight the need for further research in robust verification systems to ensure reliability in AI-driven research workflows.

%% file: section/appendix.tex
\clearpage

\section{Limitations}\label{app:limitations}

\paragraph{Benchmark Coverage} 
By prioritizing copyright compliance(Appendix~\ref{app:spot}), contamination prevention, and annotation accuracy, \benchmark{} remains relatively modest in size. We leave the expansion of this effort to create larger, more diverse benchmarks that span additional scientific disciplines and error categories to future works. 
\paragraph{Annotation Validity and Evaluation Protocol} 
All errors in \benchmark{} are verified by explicit author acknowledgments or retraction notices, but the complexity of scientific manuscripts means some true errors may be unannotated. Our case studies reveal that false negatives can arise from the following cases:
\begin{enumerate}
    \item The author’s note contains an error location that does not sufficiently cover all the affected results.
    \item There exist smaller errors unrelated to the main technical error in the preprint.
\end{enumerate}
Conversely, false positives may occur when:
\begin{enumerate}
    \item An LLM correctly points out a theorem that contains an error, but the content in the LLM’s response is still irrelevant. 
\end{enumerate}
We therefore recommend a secondary expert review, particularly for domains with complex logical dependencies or deep specialization, to validate and refine model-flagged errors.

\section{Additional Analysis}

\subsection{Impact of Context Length in Detecting Scientific Errors}

% \begin{figure}[h]
%   \centering
%   % Left: wrap‐style figure
%   \begin{minipage}{0.45\textwidth}
%     \centering
%     \includegraphics[width=\linewidth]{figures/segments_delta.pdf}
%     \captionof{figure}{\footnotesize \textbf{Impact of context length on error detection.} Each bar shows $\Delta$ = (segment-only − full-paper) for precision and recall across five models (o3, Gemini 2.5 Pro, Claude 3.7-Sonnet:Thinking, Qwen 2.5-VL-72B-Instruct, Llama-4-Maverick).}
%     \label{fig:long_delta}
%   \end{minipage}%
%   \hfill
%   % Right: scaling test figure
%   \begin{minipage}{0.45\textwidth}
%     \centering
%     \includegraphics[width=\linewidth]{figures/scaling_test.pdf}
%     \captionof{figure}{\footnotesize \textbf{Performance of o4-mini with varying reasoning effort.} Performance is reported from three independent trials.}
%     \label{fig:scaling_test}
%   \end{minipage}
% \end{figure}

\begin{wrapfigure}{r}{0.45\textwidth}
    \vspace{-0.2in}
    \centering
    \includegraphics[width=\linewidth]{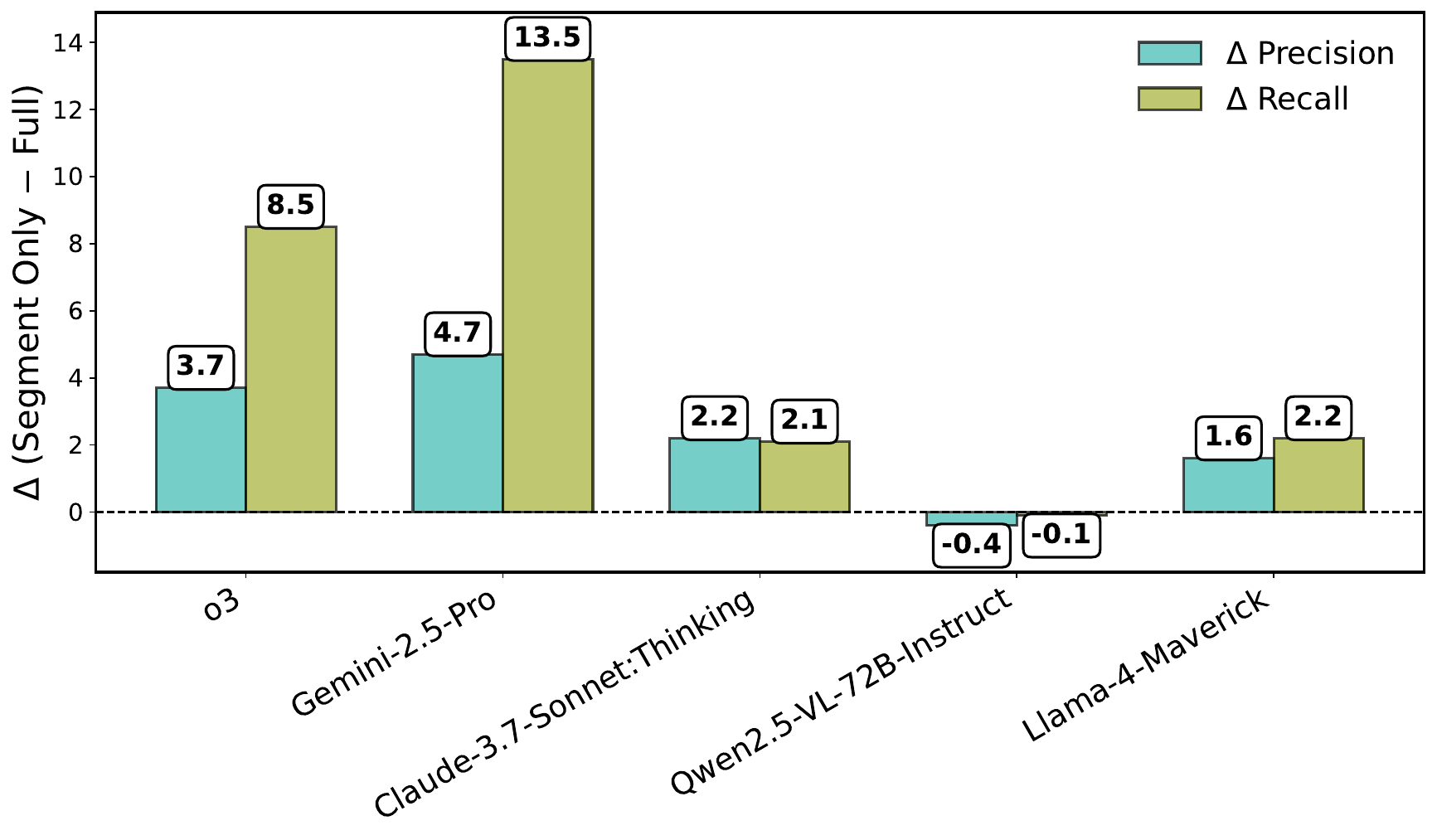}
    \caption{\footnotesize \textbf{Impact of context length on error detection.} Each bar shows $\Delta$ = (segment-only - full-paper) for precision and recall across five models (o3, Gemini 2.5 Pro, Claude 3.7-Sonnet:Thinking, Qwen 2.5-VL-72B-Instruct, Llama-4-Maverick). }
    \vspace{-0.1in}
    \label{fig:long_delta}
% \vspace{-7mm}
\end{wrapfigure}

In Table~\ref{tab:main-results}, we evaluate each model on complete manuscripts, which can span up to 140,000 characters and 90 figures. However, LLMs still struggle to synthesize long-context information~\citep{vodrahalli2024michelangelo}; to isolate the effect of context length on error detection, we extract the page containing the ground-truth error and rerun the same detection prompt on this shorter segment. Comparing the \texttt{full-paper} and \texttt{segment-only} settings decouples long-context processing from core error-detection ability. For this ablation, we conduct experiments on a subset of 36 instances, excluding the Equation/Proof category—mathematical papers often rely on global notation and prior results, making single sections insufficient—and we omit any errors that span multiple sections.

Figure~\ref{fig:long_delta} plots $\Delta$ = (segment-only – full-paper) for precision and recall. Gemini-2.5-Pro leads with gains of +4.7 precision and +13.5 recall. o3 follows at +3.7/+8.5, then Claude-3.7-Sonnet at +2.2/+2.1, and Llama-4-Maverick at +1.6/+2.2—showing that long‐context processing often masks their true error‐detection performance. o3’s smaller gains reflect the removal of Equation/Proof cases, its original strength. Qwen2.5-VL-72B-Instruct shows almost no change (–0.4 precision, –0.1 recall), indicating a fundamental limit in its error‐detection capability rather than a context‐length issue.

% \stella{It would be a good idea to look at \textit{which specific problems} the model gets right in each context. The interpretation of these numbers will differ if when going from long to short context a model starts getting some wrong.}

\subsection{Impact of Test-Time Scaling in Detecting Scientific Errors}

Test-time scaling involves adjusting the inference budget~\citep{jones2021scaling}, such as the depth of reasoning or number of solution paths explored~\citep{son2025linguistic}, to boost model performance on complex tasks. This approach is widely adopted in STEM and reasoning benchmarks~\citep{snell2024scaling}, where allocating more computational effort to inference has been shown to yield higher performance. We use OpenAI’s o4‐mini series~\citep{openai2025o3o4mini} for our experiments and vary the “reasoning effort” parameter across low, medium, and high settings.\footnote{While recent work has demonstrated similar budget controlling strategies for open models~\citep{muennighoff2025s1}, the full-size MLLMs (Llama-4-Maverick totaling 402B parameters) were too large to host for multi‐thousand‐token generations.}

\begin{wrapfigure}{r}{0.45\textwidth}
    \vspace{-0.2in}
    \centering
    \includegraphics[width=\linewidth]{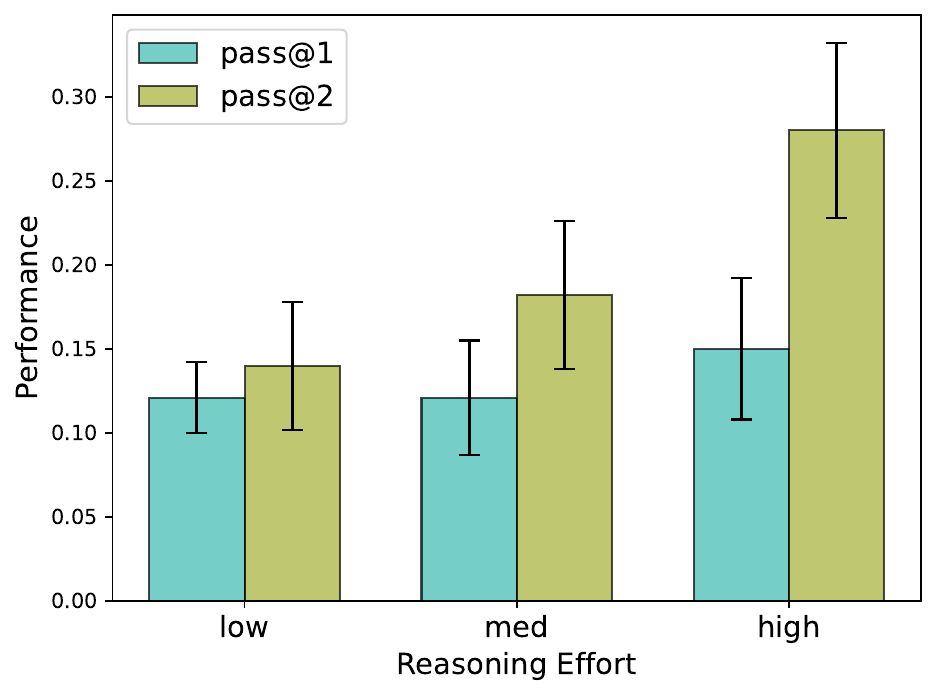}
    \caption{\footnotesize \textbf{Performance of o4-mini with varying reasoning effort.} Performance is reported from three independent trials.}
    \label{fig:scaling_test}
\vspace{-7mm}
\end{wrapfigure}

In Figure \ref{fig:scaling_test}, we see that o4-mini’s error-detection performance increases almost linearly with higher reasoning effort, demonstrating that scaling computation at test time effectively boosts accuracy. This finding is consistent with how specialized “Thinking” modes (e.g., Claude 3.7-Sonnet:Thinking VSClaude 3.7-Sonnet) and reasoning-trained models (DeepSeek-R1 vs. DeepSeek-V3) deliver similar boosts, also in line with recent error-detection literature~\citep{ahuja2025finding}.

\section{Estimating Confidence for \texorpdfstring{$\mathrm{pass}@K$}{pass@K}}
\label{app:confidence_for_passk}
Given \(N\) papers with ground‐truth error sets \(G_1,\dots,G_N\), we define the \(\mathrm{pass}@K\) metric as

\begin{equation}
    \mathrm{pass}@K 
    = \frac{1}{\sum_{i=1}^N |G_i|}
  \sum_{i=1}^N \sum_{g\in G_i}
    \mathbf{1}\!\Bigl[\exists\,s\in\{1,\dots,K\}:\;g\in p_i[s]\Bigr],
\end{equation}

where \(p_i[s]\) denotes the set of errors predicted in the \(s\)th run for paper \(i\).  This captures the fraction of all ground‐truth errors detected in at least one of \(K\) independent attempts.

\subsection{Unbiased Per‐Error Confidence}

To assign each ground‐truth error \(g\in G_i\) a \emph{confidence} score, we perform \(n\) independent runs (here \(n=8\)) and let \(c_{i,g}\) be the number of runs in which \(g\) is detected.  The probability that all \(K\) fresh attempts miss \(g\) is

\begin{equation}
\frac{\binom{n - c_{i,g}}{K}}{\binom{n}{K}},
\end{equation}

So one minus this quantity is the probability of \(\ge1\) success~\citep{chen2021evaluating}.  Hence the unbiased estimator for the pass@\(K\) probability of error \(g\) is

\begin{equation}
\hat p_{i,g}
= 1 \;-\; \frac{\binom{n - c_{i,g}}{K}}{\binom{n}{K}}.
\label{eq:error_confidence}
\end{equation}

\subsection{Aggregating Confidence and Calibration}

We then aggregate these per‐error confidences into an overall self‐estimated confidence:

\begin{equation}
\mathrm{Confidence}
= \frac{1}{\sum_{i=1}^N |G_i|}
  \sum_{i=1}^N \sum_{g\in G_i} \hat p_{i,g}.
\end{equation}

\section{Case Study}
\label{app:case_study}
\subsection{Mathematics}
We selected \cite{petersen2024} and \cite{park2024}, both in the area of algebraic geometry in pure mathematics. \cite{petersen2024} studies the configuration spaces of points in algebraic varieties with a multiplicative decomposition, and discusses some applications such as the cohomology of moduli stacks of hyperelliptic curves. It was withdrawn because of a gap that lies in the core arguments of Theorem 1.8 and Theorem 1.13. On the other hand, \cite{park2024} studies projective hypersurfaces with real coefficients and presents an explicit description of the complex conjugation of singular cohomology in terms of algebraic de Rham cohomology. The authors' withdrawal note on arXiv suggests that a hidden assumption on Hodge numbers was accidentally introduced in the early set-up of section 2.2, thus making Theorem 2.3 only valid under the additional assumption. We run our evaluation using OpenAI o3 and Gemini 2.5 Pro over the two pre-prints, once for each combination.

\begin{figure}[ht]
  \centering
\begin{tcolorbox}[colback=gray!10, colframe=blue!5, width=\textwidth,fontupper=\small]
\begin{itemize}[leftmargin=1em, labelsep=0.5em, itemsep=0.5ex, topsep=1ex]
  \item \textbf{Section 2.3, proof of Theorem 2.3:}  
    They set 
    \[
      C\bigl(\varphi_C(U_k),\varphi_C(U_k)\bigr)
      = \int_X \varphi_C(U_k)\wedge\varphi_C(U_k)
    \]
    and claim it equals the nonzero quantity \(U_{k,\gamma}\); however, for even \(n\), \(\varphi_C(U_k)\) has odd degree \(n-1\), so 
    \(\varphi_C(U_k)\wedge\varphi_C(U_k)=0\)
    identically, making the equality impossible and invalidating Lemma 2.5 and Theorem 2.3.
  \item \textbf{Example 2.4:}  
    Using the erroneous formula without complex conjugation yields \(\mu=0\), which would force \(\varphi_C([1])=0\), contradicting that \([1]\) represents the holomorphic 1-form. This example exposes the same flaw as in the main proof.
  \item \textbf{Theorem 2.3 (implicit assumption):}  
    Formula (2.10) divides by \(U_{k,\gamma}\), but the proof provides no argument that \(U_{k,\gamma}\neq0\). In fact, for odd-degree forms \(U_{k,\gamma}\) vanishes, so the construction of \(\tilde f_{ji}(x)\) is undefined.
\end{itemize}
\end{tcolorbox}
  \caption{\footnotesize o3’s initial feedback on \cite{park2024}.}
  \label{fig:o3-park-response-0}
\end{figure}

\begin{figure}[ht]
  \centering
\begin{tcolorbox}[colback=gray!10, colframe=blue!5, width=\textwidth,fontupper=\small]
\begin{itemize}[leftmargin=1em,labelsep=0.5em,itemsep=0.5ex,topsep=1ex]
  \item \textbf{Section 2.5:} Corollary 2.7 incorrectly describes the R-span of the algebraic representatives for the real primitive cohomology \(H^{n-1}_{\mathrm{prim}}(X,\mathbb{R})\). The basis elements given, \([\omega_{j_i}]\) and \([\sqrt{-1}\,\omega_{j_i}]\), correspond under \(\varphi_C\) to \(\Re(\omega_k)\) and \(i\,\Re(\omega_k)\). The correct basis for the real cohomology \(H(X,\mathbb{R})\) should correspond to \(\Re(\omega_k)\) and \(\Im(\omega_k)\), where \(\omega_k = \varphi_C(U_k)\). This makes the explicit description of \((\varphi_C^{-1}\circ P)\bigl(H^{n-1}_{\mathrm{prim}}(X,\mathbb{R})\bigr)\) scientifically invalid.
  \item \textbf{Example 2.8:} This example illustrates Corollary 2.7 for elliptic curves and thus inherits the same flaw. The stated R-span \(\mathbb{R}[\dots] + \sqrt{-1}\,\mathbb{R}[\dots]\) describes an incorrect subspace related only to the real part of the basis forms, not the full real cohomology \(H^1(X,\mathbb{R})\).
\end{itemize}
\end{tcolorbox}
  \caption{\footnotesize Gemini-2.5-Pro’s initial feedback on \cite{park2024}.}
  \label{fig:g25-park-response-0}
\end{figure}

Since we have already analyzed the LLM responses for \cite{petersen2024} in Section \ref{sec:case_study}, here we only 
describe some additional analysis on \cite{park2024}. Our first attempt at running our pipeline introduced many OCR errors, such as missing conjugations in an integral. Both o3 and Gemini-2.5-Pro reacted to different OCR errors and identified them as mistakes (Figures~\ref{fig:o3-park-response-0} and \ref{fig:g25-park-response-0}).

After an improvement (inclusion of GPT-4.1 mentioned in Section~\ref{sec:benchmark}), surprisingly, o3 claimed that there are no errors in \cite{park2024}.

As to Gemini 2.5 Pro, it proposed several potential mistakes (Figure~\ref{fig:g25-park-response-1}). First, it argued that in the statement of Theorem 2.3, ``the existence of a polynomial $g_{j_i}$ such that $r_{f_{j_i},g_{j_i}} \neq 0$" was never justified. Although this may be a fair writing feedback, this does not constitute an error. Aside from another OCR problem where the LLM saw an incorrect mathematical equation, another argument it made was that in Example 2.8, the basis of the cohomology on the right-hand side was unconventional and $\mathbb{R}\cdot[\omega]+\mathbb{R}\cdot(i[\omega])$ was better. This was made out of context, and the LLM might have forgotten that this example was an application of Corollary 2.7.

\begin{figure}[ht]
  \centering
\begin{tcolorbox}[colback=gray!10, colframe=blue!5, width=\textwidth,fontupper=\small]
\begin{itemize}[leftmargin=1em, labelsep=0.5em, itemsep=0.5ex, topsep=1ex]
  \item \textbf{Section 2.2, Theorem 2.3:}  
    The theorem requires the existence of a polynomial \(g_{ji}\) such that
    \[
      r_{f_{ji},g_{ji}} \neq 0
    \]
    (hence \(f_{ji}\notin \mathrm{Jac}(G)\)). However, the paper never justifies that the chosen basis elements 
    \[
      U_k = y^i f_{ji}
    \]
    (relative to \(\mathrm{Jac}(S)\)) can always be represented by polynomials \(f_{ji}\) lying outside \(\mathrm{Jac}(G)\). This hidden assumption is crucial for formula (2.10) to be well-defined and for the theorem to hold.
  \item \textbf{Section 2.2, Example 2.4:}  
    The text claims the constant \(\mu\) is proportional to
    \[
      \bigl(\!\int_{\gamma_1}\!\omega\bigr)\,\bigl(\!\int_{\gamma_2}\!\omega\bigr)
      -\bigl(\!\int_{\gamma_2}\!\omega\bigr)\,\bigl(\!\int_{\gamma_1}\!\omega\bigr).
    \]
    But by Definition 2.2 and the pairing \(C(\omega,\overline{\omega})\), it should be
    \[
      \bigl(\!\int_{\gamma_1}\!\omega\bigr)\,\bigl(\!\int_{\gamma_2}\!\overline{\omega}\bigr)
      -\bigl(\!\int_{\gamma_2}\!\omega\bigr)\,\bigl(\!\int_{\gamma_1}\!\overline{\omega}\bigr),
    \]
    —the presented formula omits the complex-conjugate terms, indicating a calculation error.
  \item \textbf{Section 2.4:}  
    For odd \(n\), the paper defines
    \[
      U_{k,\gamma} = PX \;\cdot\; P \;\cdot\; PX^T.
    \]
    But since \(U_{k,\gamma}\) comes from the pairing \(C(\phi_C(U_k),\overline{\phi_C(U_k)})\) and \(PX\) is complex, the correct expression uses the conjugate transpose:
    \[
      U_{k,\gamma} = PX \;\cdot\; P \;\cdot\; \overline{PX}\!^T,
    \]
    so using \(PX^T\) is mathematically incorrect.
  \item \textbf{Section 2.5, Example 2.8:}  
    The basis for the real cohomology is written as
    \[
      \mathbb{R}\bigl[(1 + \mu\,y^d(x))/2\bigr]
      + \sqrt{-1}\,\mathbb{R}\bigl[(1 + \mu\,y^d(x))/2\bigr],
    \]
    which is unconventional and obscures the structure. A clearer description is
    \[
      \mathbb{R}\cdot[\omega]
      + \mathbb{R}\cdot\bigl(i[\omega]\bigr),
    \]
    explicitly showing the real subspace spanned by \([\omega]\) and its \(i\)-multiple.
\end{itemize}
\end{tcolorbox}
  \caption{\footnotesize Gemini-2.5-Pro’s regenerated feedback on \cite{park2024}.}
  \label{fig:g25-park-response-1}
\end{figure}

\clearpage

\subsection{Material science}

We select \cite{ye2025superacid} and \cite{altuijri2024impacts}, two recent materials-science studies. \cite{ye2025superacid} investigates the synthesis of covalent organic frameworks (COFs), a class of porous materials with diverse applications, using a novel superacid-catalyzed process that enables the fabrication of new types of COF architectures. Several errors were pointed out about this article, mostly due to mislabeling COFs either in the text or in the figures, but none of them impact the conclusions of the article. \cite{altuijri2024impacts} proposes a method to increase the hydrophobicity and electrical properties of certain types of films by incidence with low-energy argon beams. This work had a wrong caption, claiming to have used a different beam intensity on a certain panel when that panel was at a different magnification. Both these errors are easier to spot than the ones found in the mathematics articles. While some of the errors in  \cite{ye2025superacid} require some domain knowledge in COFs, most of the other errors can be spotted by paying close attention to the text and figures.

\begin{figure}[ht]
  \centering
\begin{tcolorbox}[colback=gray!10, colframe=blue!5, width=\textwidth,fontupper=\small]
\begin{itemize}[leftmargin=1em, labelsep=0.5em, itemsep=0.75ex, topsep=1ex]
  \item \textbf{Section 2.2, Model Reactions (Figures 1C–G):}  
    Quantitative yields of individual species are derived from a single UV absorbance at 240 nm without extinction‐coefficient calibration or spectral deconvolution, even though all species overlap strongly in this region. The reported 92 \%, 82 \%, etc. compositions are therefore not experimentally supported.
  \item \textbf{Section 2.2 / Figure 1H:}  
    Assumes acetals are unaffected by $\text{NaHCO}_3$ neutralization; no control provided. If acetals hydrolyze under basic conditions, the calculated “protected” fractions are invalid, undermining the mechanistic interpretation.
  \item \textbf{Section 2.3:}  
    Conclusion that barriers < 20 kcal $mol^{-1}$ imply “full reversibility at room temperature” ignores solvent and entropic contributions; a 19 kcal mol$^{-1}$ barrier can correspond to multi‐hour timescales. The stated kinetic implication is not justified by the DFT data.
  \item \textbf{Abstract \& Section 2.5:}  
    Optimal TfOH concentration is stated as 0.006–0.016 M, yet all optimization and “standard” syntheses use 0.2 M TfOH. Internal inconsistency invalidates the universality of the claimed optimum.
  \item \textbf{Section 2.4 / Figure 2:}  
    Higher PXRD peak counts are used as evidence of greater crystallinity, but raw intensity is instrument‐ and packing‐dependent. Without normalization or an internal standard, counts cannot be compared between samples, so the claim of superior crystallinity is unsupported.
  \item \textbf{Section 2.7, “nine new COFs” claim:}  
    Authors conclude AcOH “cannot” give these COFs after testing only a single AcOH condition per monomer pair. Lack of systematic screening means the impossibility claim is unsubstantiated.
  \item \textbf{Section 2.11, Photocatalysis:}  
    Performance deemed “exceptional” based solely on DPBF bleaching time, with no quantum yield, turnover number, or photon‐normalized rate. Comparisons to literature therefore lack a common metric, invalidating the superiority claim.
\end{itemize}
\end{tcolorbox}
  \caption{\footnotesize o3’s regenerated feedback on \cite{ye2025superacid}.}
  \label{fig:o3-ye-response-0}
\end{figure}

Most of the errors OpenAI's o3 pointed out on \cite{ye2025superacid} were on the experimental setup and the conclusions that the authors derived from them. Although the specific points raised by o3 seem to be plausible mistakes, all of them either stem from misunderstandings by o3 or from ignoring information. For example in Figure~\ref{fig:o3-ye-response-0}, the claim that the UV-derived yields in Section 2.2 (Figures 1C–G) lack support is a hallucination: the authors validate those compositions with 1H NMR spectroscopy, which produces results similar to the UV measurements, so it is incorrect to argue that the yields are unsupported. In Section 2.3, o3 contends that barriers below 20 kcal mol$^{-1}$  cannot guarantee “full reversibility” at room temperature. While entropic factors do affect reaction rates, barriers under 20 kcal mol$^{-1}$ are generally regarded as low enough for reversibility, and multi-hour timescales still allow dynamic equilibrium. This error likely stems from o3’s unfamiliarity with practical conventions in kinetic chemistry. Finally, for photocatalysis performance (Section 2.11), o3 questions the “exceptional” activity because no quantum yields or turnover numbers are provided. However, o3 does not have access to the supplementary information and thus is hallucinating that no DPBF bleaching comparisons exist, Table S10 demonstrates that these COFs outperform those in the literature.

\begin{figure}[ht]
  \centering
\begin{tcolorbox}[colback=gray!10, colframe=blue!5, width=\textwidth,fontupper=\small]
\begin{itemize}[leftmargin=1em, labelsep=0.5em, itemsep=0.75ex, topsep=1ex]
  \item \textbf{Section 2.3 and Figure 1I (related to Section 2.2(2)):}  
    The DFT‐calculated relative energy of the acetal product P1 (+7.4 kcal/mol compared to benzaldehyde) indicates it is thermodynamically highly unfavorable, directly contradicting the experimental 81 \% yield reported in Section 2.2(2), which would require a negative $\Delta$G (approx. –0.85 kcal/mol). This discrepancy undermines the validity of the DFT calculations as an explanation for the experimentally observed acetal protection.

  \item \textbf{Section 2.5(1) and Figures 3A, 3C, 3E:}  
    The text states that low-boiling point solvents (MeOH, EtOH, n-propanol) are 'unsuitable' for reactions at 120 °C due to their boiling points being below this temperature. However, Figure 3 A,C,E present PXRD data for COFs synthesized in these solvents, presumably at 120 °C. The manuscript fails to specify whether these reactions were performed in sealed vessels. If not, solvent evaporation would lead to uncontrolled and non-reproducible reaction conditions, invalidating the results obtained with these specific solvents.

  \item \textbf{Section 2.6:}  
    The BET surface area for the scaled-up TFPPy-PDA-COF is reported as “1606 cm\textsuperscript{2} g\textsuperscript{–1}.” The correct unit is m\textsuperscript{2}/g. This unit error misrepresents the surface area value by a factor of 10,000 and is a fundamental error in data reporting.

  \item \textbf{Abstract; Introductory paragraph; Section 2.7; Figure 4C:}  
    The paper claims “nine new imine-linked frameworks,” but Figure 4C highlights ten COFs (\#5, \#9, \#13, \#14, \#15, \#18, \#23, \#24, \#25, \#28) as newly synthesized or inaccessible by traditional methods, creating a numerical inconsistency.

  \item \textbf{Unnumbered paragraph on PXRD for Figure 6H:}  
    There is a contradiction in the indexing of PXRD peaks for TAPPy-TFPPy-COF (Figure 6H). The peaks are initially assigned to facets including (020): 'TAPPy-TFPPy-COF displayed peaks at 5.00°, 6.90°, 10.08°, 15.24°, and 23.22°...assigned to the (110), (020), (220), (330), and (001) facets...'. A subsequent concluding sentence for this group of COFs states related peaks 'were attributed to the (100), (110), (220), (330), and (001) facets...'. The differing indices for the second peak ((020) vs. (100)) for the same data create ambiguity and undermine the reliability of the structural characterization from PXRD for this COF.
\end{itemize}
\end{tcolorbox}
  \caption{\footnotesize Gemini-2.5-Pro’s regenerated feedback on \cite{ye2025superacid}.}
  \label{fig:oㅎ25-ye-response-0}
\end{figure}

Gemini 2.5 Pro also misunderstand the authors, in it feedback regarding Section 2.7 (Figure 4C) is related to the fact that the authors claim that a certain reagent is not suitable for the production of a COF and they show how it does not produce such a good results, because of this Gemini claims that the reagent can actually be used because the authors used it in a figure. Contrary to o3, it does find a real mistake by noticing that one of the characterizations of a certain COF contained the wrong unit, making the measurement 10000 times smaller.

\begin{figure}[ht]
  \centering
\begin{tcolorbox}[colback=gray!10, colframe=blue!5, width=\textwidth,fontupper=\small]
\begin{itemize}[leftmargin=1em, labelsep=0.5em, itemsep=0.75ex, topsep=1ex]
  \item \textbf{Methods – SRIM simulation / Figure 2:}  
    Authors discuss ion penetration up to 1 µm for 4 keV $\text{Ar}^+$, whereas SRIM and the plotted axis (0–1000 Å) show a stopping range below 20 nm. Conclusions about bulk (µm-scale) modification rest on an impossible depth estimate.
  \item \textbf{Results – EDX (Figures 3 \& 4):}  
    Claim that irradiation changed elemental composition is based on C, O, N shifts under 2 wt \%, well inside normal EDX error; no statistics provided, so the asserted chemical change is unsupported.
  \item \textbf{Results – Contact angle / Work of adhesion (Eq.~[1], Figures 7 \& 8):}  
    Reported \(W_a\) values do not satisfy \(W_a = \gamma_\ell(1+\cos\theta)\); for example, \(\theta=62.1^\circ\) for water gives 106 mJ m\(^{-2}\), but the table lists 74.2 mJ m\(^{-2}\). All subsequent surface-energy conclusions use these erroneous numbers.
  \item \textbf{Results – Surface free energy (Figure 9):}  
    Owens–Wendt requires two liquids to solve for \(\gamma_s^{\mathrm d}\) and \(\gamma_s^{\mathrm p}\). With only two liquids the solution at each fluence is unique, yet the authors report independent trends for both components, indicating a mathematical inconsistency.
  \item \textbf{Results – Electrical conductivity (Figure 10, text):}  
    Two mutually contradictory pristine conductivities are given (\(3.8\times10^{-7}\) S cm\(^{-1}\) vs. \(1.1\times10^{-8}\) S cm\(^{-1}\)). The inconsistency invalidates the reported enhancements due to irradiation.
  \item \textbf{Results – Potential barrier calculation (Figure 11, Eq.~[4]):}  
    Formula \(W_m = -4k_BT/m\) sets an upper magnitude of 0.104 eV at 300 K, yet values up to 1.07 eV are reported and the sign is ignored; the dielectric analysis is therefore mathematically impossible.
\end{itemize}
\end{tcolorbox}
  \caption{\footnotesize o3’s regenerated feedback on \cite{altuijri2024impacts}.}
  \label{fig:o3-altuijri-response-0}
\end{figure}

Figure~\ref{fig:o3-altuijri-response-0}, are feedbacks of o3 on \cite{altuijri2024impacts}. Here, it seems like many of o3’s flagged issues arise from misinterpretation rather than true errors. Its EDX objection ignores that the authors corroborate composition changes with complementary measurements and that small weight‐percent shifts demand statistical treatment. Regarding contact angle, o3 uses diiodomethane data instead of water, miscomputing \(W_a\). The supposed Owens–Wendt inconsistency stems from a misunderstanding that two liquids yield a unique solution per fluence, so the reported parallel trends do not imply mathematical error. Finally, the potential barrier formula comes from literature conventions, and Figure 11 presents a related quantity rather than misapplying the sign.

\begin{figure}[ht]
  \centering
\begin{tcolorbox}[colback=gray!10, colframe=blue!5, width=\textwidth,fontupper=\small]

\begin{itemize}[leftmargin=1em,labelsep=0.5em,itemsep=0.75ex,topsep=1ex]
\item \textbf{Figure 2a:}  
    The text states SRIM simulation data is shown for depths of 0 to 1 µm, but the x-axis of Figure 2a (“Target Depth”) only extends to 1000 Å (0.1 µm). This misrepresents the range of simulation data presented visually.

  \item \textbf{Results and Discussion – Contact Angle, Work of Adhesion, Surface Free Energy (Figures 7, 8, 9):}  
    Table I (containing contact angle and work of adhesion data) and Table II (containing surface free energy data) are referenced but not provided. This prevents independent verification of the data plotted in Figures 7–9 and the calculations derived from them, undermining the scientific validity of these results.

  \item \textbf{Results and Discussion – Work of Adhesion (Figure 8):}  
    The work of adhesion \(W_a\) values reported in the text (and presumably plotted in Figure 8) are inconsistent with calculations based on contact angle data from Figure 7 and standard liquid surface tension values (e.g., water at $4\times10^{14}\,\mathrm{ions/cm}^2$: reported $W_a = 74.2\,\mathrm{mJ/m}^2$, calculated $W_a \approx 106.4\,\mathrm{mJ/m}^2$) . The liquid surface tension values ( \(\gamma_\ell\)) used for calculations are not stated, and the discrepancy suggests significant errors in calculation or data, invalidating the reported \(W_a\) values.

  \item \textbf{Results and Discussion – Figure 10 and related text:}  
    AC conductivity \(\sigma_{\mathrm{ac}}\) values at 50 Hz reported in the text for irradiated samples differ significantly (e.g.\ for \(4\times10^{14}\) ions·cm\textsuperscript{–2}: text \(3.8\times10^{-7}\ \mathrm{S\,cm^{-1}}\) vs.\ Fig. 10 data \(\sim3.74\times10^{-8}\ \mathrm{S\,cm^{-1}}\)) in a summary paragraph preceding the electrical characterization. This contradiction undermines the baseline characterization of the material.

  \item \textbf{Results and Discussion – Electrical Properties:}  
    The AC conductivity \(\sigma_{\mathrm{ac}}\) for unirradiated (pristine) CA/PANI at 50 Hz is reported inconsistently: as \(1.1\times10^{-8}\ \mathrm{S\,cm^{-1}}\) in the Figure 10 discussion (matching the figure) and as \(3.8\times10^{-7}\ \mathrm{S\,cm^{-1}}\) in a preceding summary paragraph.

  \item \textbf{Results and Discussion – Figure 11 and Equation [4]:}  
    The exponent \(m\) for calculating \(W_m\) is derived from a narrow low-frequency range in Figure 11 (\(\ln\omega\approx5.8\text{–}6.5\)), which is not justified and may not be representative of the model used. The validity of conclusions based on \(W_m\) is therefore questionable

  \item \textbf{Results and Discussion – Equation [4]:}  
    The formula
    \[
      W_m = -\frac{4k_B T}{m}
    \]
    used for calculating the maximum potential barrier is unconventional for hopping conduction models, and its scientific basis or appropriateness for this system is not established via the provided citation [40] or derivation. Using an potentially incorrect or unsubstantiated formula invalidates the calculated \(W_m\) values.

  \item \textbf{Results and Discussion – Calculation of \(W_m\):}  
    The text states that the estimated \(W_m\) “decreased from 0.116 eV…to 1.03 eV.” However, the authors’ reported values \((0.116,\ 0.11,\ 1.07,\ 1.03)\) eV show a non-monotonic behavior (slight decrease, then large increase, then slight decrease). This is a misrepresentation of their calculated results.
    
\end{itemize}
\end{tcolorbox}
  \caption{\footnotesize Gemini-2.5-Pro’s regenerated feedback on \cite{altuijri2024impacts}.}
  \label{fig:g25-altuijri-response-0}
\end{figure}

In Figure~\ref {fig:g25-altuijri-response-0} we observe multiple errors where Gemini-2.5-Pro fails to identify visual inputs correctly; for instance, it claims that Tables I and II referenced in the discussion of Figures 7–9 are missing, even though those tables are present, indicating it has failed to identify the context. Additionally, in Figure 10, it misreads a log-scaled plot as linear and thereby hallucinates values.

\subsection{Environmental Science}

\begin{figure}[ht]
  \centering
\begin{tcolorbox}[colback=gray!10, colframe=blue!5, width=\textwidth,fontupper=\small]

\begin{itemize}[leftmargin=1em,labelsep=0.5em,itemsep=0.75ex,topsep=1ex]
  \item \textbf{Section 3.1.1 / Table 1:}  
    Temperature “predictions” obtained from \(T = \tfrac{PM}{\rho R}\) are tautological because \(\rho = \tfrac{PM}{RT}\) by definition; using measured \(P\), \(M\), and \(\rho\) always reproduces the measured \(T\). This provides no evidence against the greenhouse effect yet is used as the cornerstone of the paper’s conclusions.

  \item \textbf{Section 3.1.2 (Equations 3–4 and Figure 1):}  
    The forcing equation is applied with \(\log_{10}\) instead of the natural log, reducing \(\Delta F\) by \(\approx2.3\times\) and yielding a spurious warming of \(0.64\)–\(0.81\,^\circ\mathrm{C}\) for CO\(_2\) doubling. Correct use of \(\ln\) gives \(1.5\)–\(1.9\,^\circ\mathrm{C}\), invalidating the claim that future warming is negligible.

  \item \textbf{Section 3.1.2 (Climate-sensitivity discussion):}  
    Derived ECS \(\approx0.5\,^\circ\mathrm{C}\) ignores water-vapour, lapse-rate, and cloud feedbacks without justification;  hence the stated upper bound on warming lacks the physical components known to amplify $\text{CO}_2$ forcing.

  \item \textbf{Section 3.1.4 (Residence time of CO\(_2\)):}  
  Paper conflates 3–5 yr molecular turnover time with the multi‑century adjustment time of a concentration perturbation, leading to the unsupported claim that anthropogenic CO\(_2\) ‘quickly disappears’ and is climatically irrelevant.
  
  \item \textbf{Section 3.1.4 (Anthropogenic share argument):}  
    Using the fact that anthropogenic emissions are \(\sim4\%\) of gross annual flux to assert they cannot raise atmospheric CO\(_2\) ignores that natural sources and sinks are nearly balanced; failure to perform a mass‑balance analysis renders the conclusion scientifically invalid.

  \item \textbf{Section 3.1 (General claim of “no greenhouse effect”):}  
    Assertion that adiabatic compression or thermal conductivity alone explains the 33 K surface–space temperature difference disregards radiative‐transfer energy balance and provides no quantitative mechanism, contradicting well‑established physics without offering an alternative that reproduces observed fluxes.
\end{itemize}
\end{tcolorbox}
  \caption{\footnotesize o3’s regenerated feedback on \cite{simpson2024scientific}.}
  \label{fig:o3-simpson-response-0}
\end{figure}

% We selected \cite{simpson2024scientific} in environmental science for its comprehensive evaluation of human contributions to atmospheric CO$_2$. In a data-driven analysis, \cite{simpson2024scientific} argues that human-induced CO$_2$ emissions are negligible compared to natural sources, thereby questioning the validity of the Greenhouse Gas Hypothesis. However, the paper contains a significant miscalculation in Section 3.1.2, ``Measurement of Infrared Absorption of the Earth's Atmosphere.'' Specifically, the author incorrectly applies Equation~(3) from the IPCC: $F = 5.35 \ln\left(\frac{C_t}{C_0}\right)$ by using the base-10 logarithm ($\log_{10}$) instead of the natural logarithm ($\ln$), leading to erroneous numerical values. Additionally, the paper has been challenged for several arguable errors, including questionable core assumptions and misuse of the ideal gas law.

% Regarding the miscalculation, o3 correctly identifies the target error: ``With $\lambda = 0.4$--$0.5\, \mathrm{K\, (W\, m^{-2})^{-1}}$, the correct $\Delta T$ for doubling CO$_2$ (400$\rightarrow$800 ppm) is 1.5--1.9~$^\circ$C, not 0.64--0.81~$^\circ$C as stated,'' highlighting a mismatch between the scale of results and the correct calculation basis. On the other hand, Gemini2.5 fails to point out this specific error, despite generating seven potential factual faults in the paper.

% Regarding the arguable errors, both o3 and gemini makes several claims that seem valid, such that the core assumption of the ideal gas law in the paper may not be suitable for estimating global-scale climate effects.

We selected \cite{simpson2024scientific} in environmental science, the paper conductsa comprehensive evaluation of human contributions to atmospheric CO$_2$. In a data-driven analysis, \cite{simpson2024scientific} argues that human-induced CO$_2$ emissions are negligible compared to natural sources, thereby questioning the validity of the Greenhouse Gas Hypothesis. However, the original version of the paper contains a miscalculation in Section 3.1.2, ``Measurement of Infrared Absorption of the Earth's Atmosphere.'' Specifically, the author incorrectly applies Equation~(3) from the IPCC: $F = 5.35 \ln\left(\frac{C_t}{C_0}\right)$ by using the base-10 logarithm ($\log_{10}$) instead of the natural logarithm ($\ln$), leading to erroneous numerical values. Regarding the miscalculation, in Figure~\ref{fig:o3-simpson-response-0}, o3 correctly locates the target error: ``\(\log_{10}\) instead of the natural log, reducing \(\Delta F\) by \(\approx2.3\times\) and yielding a spurious warming of \(0.64\)–\(0.81\,^\circ\mathrm{C}\) for CO\(_2\) doubling. Correct use of \(\ln\) gives \(1.5\)–\(1.9\,^\circ\mathrm{C}\),'' highlighting a mismatch between the scale of results and the correct calculation basis. However, its subsequent claim that this error “invalidates the assertion of negligible future warming” seems overstated. As the authors acknowledge, projected temperature increases remain modest. In short, the magnitude of these miscalculations is insufficient to overturn the paper’s broader argument about limited warming. 

On the other hand, Gemini2.5 fails to point out the specific error.

% Regarding the arguable errors, both o3 and gemini makes several claims that seem valid, such that the core assumption of the ideal gas law in the paper may not be suitable for estimating global-scale climate effects.

% \subsection{Quantum physics}
% We selected 

\section{Additional Details on \benchmark{}}\label{app:spot}

\paragraph{License \& Copyright}

\benchmark{} comprises 83 manuscripts published across 28 venues (including arXiv). Of these, 62 (74.7 \%) are openly accessible under a CC license; we publicly share our fully preprocessed versions via the Hugging Face Hub. The remaining 21 (25.3 \%) are paywalled, so we do not redistribute them directly. Organizations with institutional access to Springer Nature or Elsevier can apply our preprocessing pipeline to generate their own versions.

\begin{figure}[ht]
  \centering
  
    \includegraphics[width=0.45\linewidth]{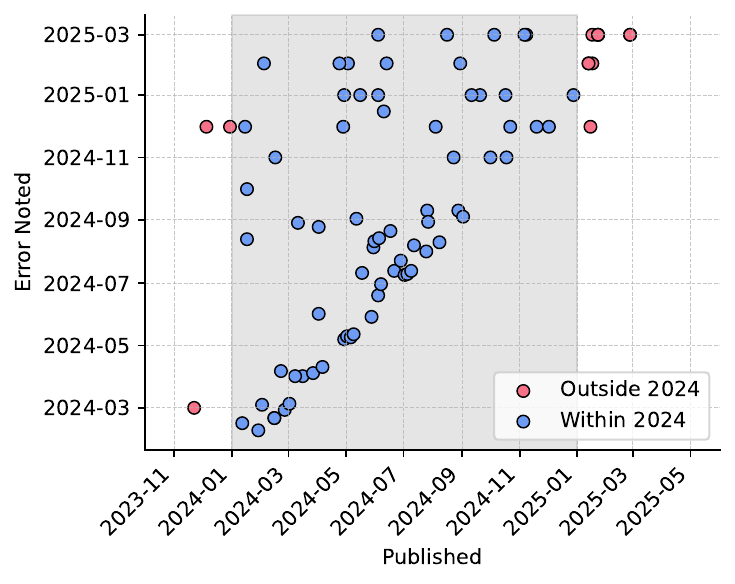}
    % \caption{place holder}
    \caption{\footnotesize \textbf{Publication dates against first error‐notice dates for the 83 manuscripts.} Each point denotes one paper; blue markers note papers published in 2024, while red markers are those otherwise.}

    % \vspace{-0.1in}
    \label{fig:date}
\vspace{-7mm}
\end{figure}

\paragraph{Date} To minimize contamination against parametric knowledge~\citep{bejan-etal-2023-make}, we aim to include only papers published from 2024 onward. As Figure~\ref{fig:date} shows, the bulk of our corpus dates to 2024, with ten papers from 2025 and three that originally appeared before 2023. Those three early manuscripts passed our automated filters because revisions were submitted after 2024; we retained them since their first error notices appeared in March 2024, minimizing any chance that models were exposed to the original withdrawal details during training.

\paragraph{Annotation} We use human annotators in Section~\ref{sec:benchmark} during the benchmark creation process. Details on the annotator guideline are available in Figure~\ref{fig:annotation-guideline}, a sample image of the platform in Figure~\ref{fig:annotation-platform}.

\begin{figure}[ht]
  \centering
  
\begin{tcolorbox}[colback=gray!15, colframe=black!40]
A lightweight Streamlit app for labeling errors discussed on \textbf{PubPeer} or papers \textbf{withdrawn from arXiv}. Contributors review randomly selected papers, answer guided questions, and append their work to \texttt{annotations.csv}.

\subsection*{Getting Started}
\subsubsection*{Prerequisites}
\begin{itemize}
  \item Python \(\geq\) 3.8
  \item \href{https://streamlit.io/}{Streamlit}
  \item pandas
\end{itemize}

\subsubsection*{Installation}
\begin{verbatim}
# 1 Clone the repo
git clone https://github.com/guijinSON/ai4s_r2.git
cd ai4s_r2

# 2 Install dependencies
pip install streamlit pandas
\end{verbatim}

\subsubsection*{Dataset}
\texttt{retracted\_machine\_filtered\_final.csv} ships with the repository—no additional download required.

\subsection*{Usage}
\begin{verbatim}
streamlit run streamlit_sample.py
\end{verbatim}
\begin{enumerate}
  \item \textbf{Shuffle Sample} loads a random paper.
  \item Complete the six annotation questions in the right panel.
  \item Click \textbf{Save Annotation} to append to \texttt{annotations.csv}.
  \item Repeat until 3--5 rows are completed.
  \item Click \textbf{Submit} to send \texttt{annotations.csv} to the maintainer.
\end{enumerate}
\end{tcolorbox}

  \caption{\footnotesize Guideline provided to annotators.}
  \label{fig:annotation-guideline}
\end{figure}

\begin{figure}[ht]
  \centering
    \includegraphics[width=\textwidth]{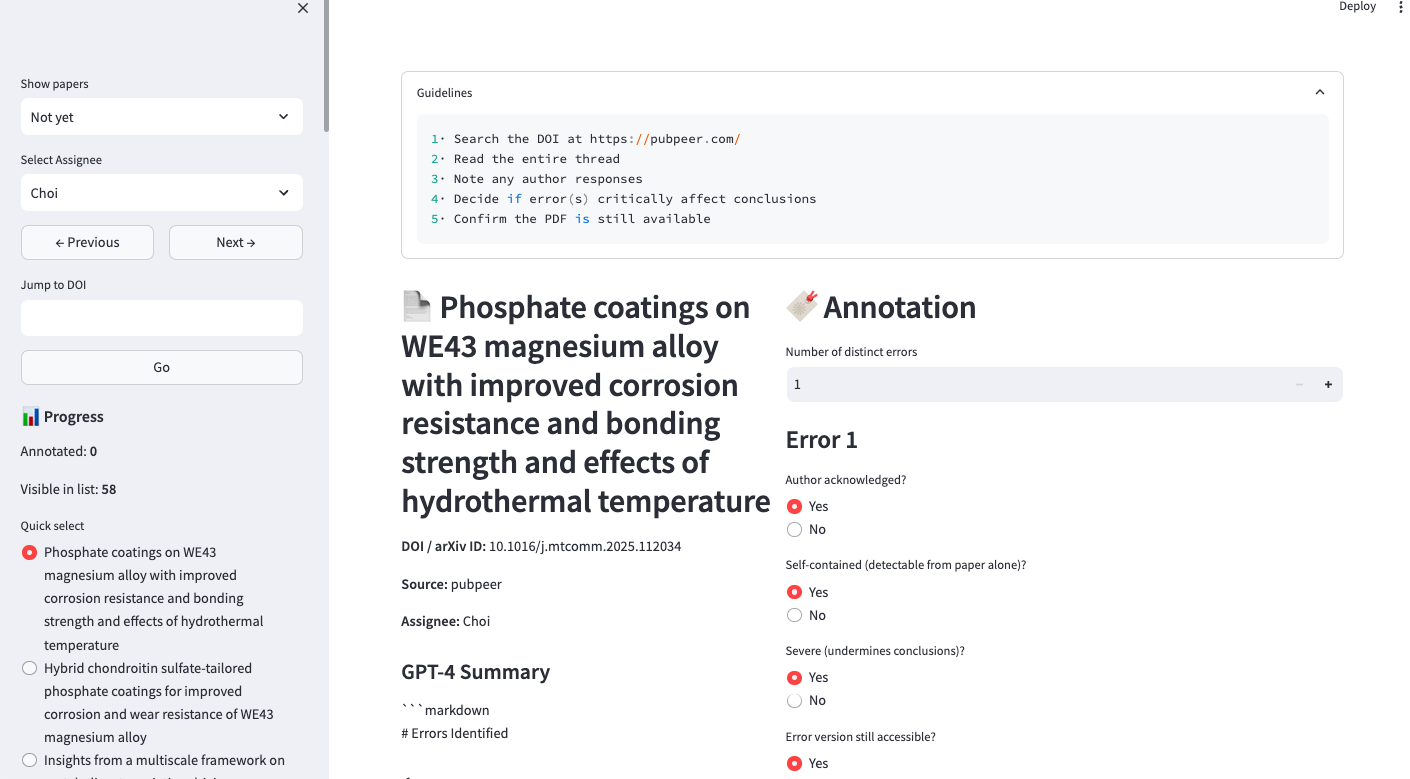}
    \caption{\footnotesize Example image of annotation platform.}
  \label{fig:annotation-platform}
\end{figure}

\section{Additional Details on Evaluation}\label{app:details_on_eval}
Evaluation consists of two phases. In the first phase, the target LLM is prompted to identify potential errors in each paper using our “Generation Prompt.” In the second phase, we employ GPT-4.1 to align and compare the model’s candidates against the ground-truth annotations with our “Evaluation Prompt.” In the remainder of this section, we specify details on generation configurations and present the full text of both prompts.

\subsection{Generation Configurations}
For each model, we adopt the provider’s recommended parameters when available; otherwise, we use a sampling temperature of 0.6, top-p of 0.95, a repetition penalty of 1.0, and enforce a minimum of 8 and a maximum of 8192 tokens.

\subsection{Prompts}

\input{tables/prompts}

\newpage

\section{Detailed Results}\label{app:detailed_resutls}

\begin{enumerate}
    \item In Table~\ref{tab:passk-o3} to \ref{tab:passk-openrouter_meta-llama_llama-4-s} we present detailed results of each model from Table~\ref{tab:main-results}.
    \item In Table~\ref{tab:passk-text-o3} to \ref{tab:passk-text-openrouter_meta-llama_llama-4-s} we present detailed results of the text-only evaluation from Table~\ref{tab:multi_modality}.
\end{enumerate}

\input{tables/add_results}
\input{tables/add_results_text}

%% file: tables/prompts.tex
\begin{tcolorbox}[colback=gray!15,colframe=black!40,title=Generation Prompt]
You are a \textbf{scientific‐rigor auditor}. You will receive the parsed contents of a research paper. Your job is to identify only those errors or flaws that directly undermine the scientific validity of the paper’s methods, analyses, or conclusions. Your sole focus is identifying flaws, such as errors in experimental design, data integrity, calculations, statistical inference, or reproducibility, that directly call into question the validity of a specific claim, paragraph, or the paper. Do not report issues purely presentational, rhetorical, stylistic, or related to citation practices.
\\
---
\\
After you’ve done a \textbf{detailed walkthrough} of the paper, output exactly in this format—no extra keys or commentary:
\\ \\ 
``` \\
<analysis> \\
\{detailed walk‑through of how you checked each section/figure and why you flagged (or did not flag) any flaw\} \\ 
</analysis> \\
\\
<response> \\
\begin{verbatim}
{
    "has_error": <true | false>,
    "errors": [
        {
            "location": "Section 2.1",
            "description": "Claim that ‘all X are Y’ is ..."
        },
        {
            "location": "Figure 3",
            "description": "X-Axis labeled ‘Time (s)’ but units ..."
        }
        // …more entries…
    ]
}
\end{verbatim}
</response>\\
\\
- Do not include other keys or prose outside these two tagged blocks.  \\
- Do not report stylistic or citation issues.  \\
- Be as precise as possible about where (section ID or figure/table) and what the scientific flaw is. \\
- Each description of the errors must be rooted in a scientific rationale explaining why they are 'wrong' (not how they could be improved). \\ 

Begin your analysis now.
\end{tcolorbox}

\begin{tcolorbox}[colback=gray!15,colframe=black!40,title=Evaluation Prompt]
You are an expert LLM-as-a-Judge. You will receive a JSON object with two arrays:\\
\\
1. "annotations": the ground‐truth errors (each has "location" and "description"). \\ 
2. "predictions": the model’s reported errors (same format).\\
\\
\textbf{Task}\\
1. Compare each prediction against each annotation.  \\
2. A match occurs only when both "location" and "description" are identical.  \\
3. Your output should be generated in the following format:\\
\\
<analysis>\\
Analysis and comparison of each prediction and annotation.\\
</analysis>\\
<response>\\
\begin{verbatim}
{
  "matches": [
      {
      "location": the location of the matched object, which should be
      based on the annotated location,
      "description": your explanation on why you think it is a match.
      },
        {
      "location": ... ,
      "description": ... 
      },
  ]
}
\end{verbatim}
</response>\\

Be rigorous in considering matches; the location may be slightly differently named, but the description must match overall. 
\end{tcolorbox}

%% file: tables/add_results.tex
\begin{table}[h]
  \centering
  \fontsize{8}{9.5}\selectfont
  \caption{\footnotesize Mean and standard deviation of $\mathrm{pass}@K$ for o3 ($K \in \{1,2,4\}$) by error category (left) and paper category (right). Detailed evaluations results for Table~\ref{tab:main-results}.}
  \label{tab:passk-o3}
  \begin{tabular}{@{} lccc @{\quad} lccc @{} }
    \toprule
    \multicolumn{4}{c}{\textbf{Error Category}} & \multicolumn{4}{c}{\textbf{Paper Category}} \\
    \cmidrule(r){1-4} \cmidrule(l){5-8}
    \multicolumn{1}{c}{Category} & pass@1 & pass@2 & pass@4 & \multicolumn{1}{c}{Category} & pass@1 & pass@2 & pass@4 \\
    \midrule
    Data Inconsistency & $13.1_{8.3}$ & $19.4_{7.0}$ & $25.7_{4.3}$ & Biology & $5.1_{5.8}$ & $9.4_{5.9}$ & $14.5_{2.5}$ \\
    Equation / proof & $33.6_{3.8}$ & $51.6_{6.1}$ & $67.5_{3.7}$ & Chemistry & $0.0_{0.0}$ & $0.0_{0.0}$ & $0.0_{0.0}$ \\
    Experiment setup & $0.0_{0.0}$ & $0.0_{0.0}$ & $0.0_{0.0}$ & Computer Science & $21.0_{11.5}$ & $36.1_{12.1}$ & $55.9_{9.5}$ \\
    Figure duplication & $0.0_{0.0}$ & $0.0_{0.0}$ & $0.0_{0.0}$ & Engineering & $0.0_{0.0}$ & $0.0_{0.0}$ & $0.0_{0.0}$ \\
    Reagent identity & $22.0_{25.1}$ & $40.7_{25.5}$ & $62.7_{10.8}$ & Environmental Science & $5.1_{12.0}$ & $10.7_{15.6}$ & $22.1_{15.8}$ \\
    Statistical reporting & $45.7_{17.2}$ & $62.8_{17.9}$ & $88.4_{12.5}$ & Materials Science & $14.2_{6.0}$ & $16.7_{0.0}$ & $16.7_{0.0}$ \\
    & & & & Mathematics & $34.3_{5.3}$ & $53.3_{6.0}$ & $67.6_{2.5}$ \\
    & & & & Medicine & $0.0_{0.0}$ & $0.0_{0.0}$ & $0.0_{0.0}$ \\
    & & & & Multidisciplinary & $20.0_{0.0}$ & $20.0_{0.0}$ & $20.0_{0.0}$ \\
    & & & & Physics & $33.7_{20.7}$ & $56.3_{17.6}$ & $79.0_{7.1}$ \\
    \bottomrule
  \end{tabular}
\end{table}

\begin{table}[h]
  \centering
  \fontsize{8}{9.5}\selectfont
  \caption{\footnotesize Mean and standard deviation of $\mathrm{pass}@K$ for GPT-4.1 ($K \in \{1,2,4\}$) by error category (left) and paper category (right). Detailed evaluations results for Table~\ref{tab:main-results}.}
  \label{tab:passk-gpt4_1}
  \begin{tabular}{@{} lccc @{\quad} lccc @{} }
    \toprule
    \multicolumn{4}{c}{\textbf{Error Category}} & \multicolumn{4}{c}{\textbf{Paper Category}} \\
    \cmidrule(r){1-4} \cmidrule(l){5-8}
    \multicolumn{1}{c}{Category} & pass@1 & pass@2 & pass@4 & \multicolumn{1}{c}{Category} & pass@1 & pass@2 & pass@4 \\
    \midrule
    Data Inconsistency & $6.4_{5.6}$ & $11.4_{6.6}$ & $19.2_{6.1}$ & Biology & $21.4_{6.5}$ & $34.8_{7.5}$ & $48.9_{7.3}$ \\
    Equation / proof & $0.4_{1.0}$ & $0.8_{1.3}$ & $1.5_{1.5}$ & Chemistry & $2.1_{5.5}$ & $4.0_{7.1}$ & $8.2_{8.3}$ \\
    Experiment setup & $0.0_{0.0}$ & $0.0_{0.0}$ & $0.0_{0.0}$ & Computer Science & $0.0_{0.0}$ & $0.0_{0.0}$ & $0.0_{0.0}$ \\
    Figure duplication & $16.3_{4.9}$ & $27.6_{5.7}$ & $41.1_{5.6}$ & Engineering & $12.9_{21.9}$ & $23.1_{24.9}$ & $39.3_{20.5}$ \\
    Reagent identity & $4.5_{11.4}$ & $8.6_{14.6}$ & $16.5_{16.7}$ & Environmental Science & $0.0_{0.0}$ & $0.0_{0.0}$ & $0.0_{0.0}$ \\
    Statistical reporting & $0.0_{0.0}$ & $0.0_{0.0}$ & $0.0_{0.0}$ & Materials Science & $14.6_{9.8}$ & $26.8_{11.3}$ & $41.5_{9.4}$ \\
    & & & & Mathematics & $0.0_{0.0}$ & $0.0_{0.0}$ & $0.0_{0.0}$ \\
    & & & & Medicine & $5.9_{16.1}$ & $11.3_{20.9}$ & $25.2_{25.0}$ \\
    & & & & Multidisciplinary & $12.7_{8.4}$ & $22.9_{8.7}$ & $36.4_{7.4}$ \\
    & & & & Physics & $0.0_{0.0}$ & $0.0_{0.0}$ & $0.0_{0.0}$ \\
    \bottomrule
  \end{tabular}
\end{table}

\begin{table}[h]
\centering
  \fontsize{8}{9.5}\selectfont
  \caption{\footnotesize Mean and standard deviation of $\mathrm{pass}@K$ for Gemini-2.5-Pro ($K \in \{1,2,4\}$) by error category (left) and paper category (right). Detailed evaluations results for Table~\ref{tab:main-results}.}
  \label{tab:passk-openrouter_google_gemini-2_5}
  \begin{tabular}{@{} lccc @{\quad} lccc @{} }
    \toprule
    \multicolumn{4}{c}{\textbf{Error Category}} & \multicolumn{4}{c}{\textbf{Paper Category}} \\
    \cmidrule(r){1-4} \cmidrule(l){5-8}
    \multicolumn{1}{c}{Category} & pass@1 & pass@2 & pass@4 & \multicolumn{1}{c}{Category} & pass@1 & pass@2 & pass@4 \\
    \midrule
    Data Inconsistency & $12.6_{7.0}$ & $22.3_{7.6}$ & $36.5_{6.9}$ & Biology & $2.0_{3.4}$ & $3.9_{4.4}$ & $7.6_{5.1}$ \\
    Equation / proof & $11.8_{7.0}$ & $21.9_{7.9}$ & $38.4_{7.3}$ & Chemistry & $8.6_{8.3}$ & $15.5_{9.5}$ & $25.9_{8.7}$ \\
    Experiment setup & $0.0_{0.0}$ & $0.0_{0.0}$ & $0.0_{0.0}$ & Computer Science & $8.2_{7.1}$ & $16.7_{9.7}$ & $31.7_{11.5}$ \\
    Figure duplication & $1.5_{2.6}$ & $2.8_{3.4}$ & $5.5_{3.9}$ & Engineering & $0.0_{0.0}$ & $0.0_{0.0}$ & $0.0_{0.0}$ \\
    Reagent identity & $0.0_{0.0}$ & $0.0_{0.0}$ & $0.0_{0.0}$ & Environmental Science & $8.3_{14.4}$ & $15.6_{16.6}$ & $26.4_{13.6}$ \\
    Statistical reporting & $15.2_{24.3}$ & $31.6_{31.5}$ & $57.7_{32.0}$ & Materials Science & $3.9_{7.1}$ & $7.5_{8.3}$ & $13.0_{6.9}$ \\
    & & & & Mathematics & $11.3_{8.5}$ & $21.1_{9.7}$ & $38.0_{8.6}$ \\
    & & & & Medicine & $0.0_{0.0}$ & $0.0_{0.0}$ & $0.0_{0.0}$ \\
    & & & & Multidisciplinary & $11.2_{5.9}$ & $20.3_{7.7}$ & $32.8_{8.8}$ \\
    & & & & Physics & $14.6_{14.5}$ & $25.7_{14.9}$ & $39.7_{12.8}$ \\
    \bottomrule
  \end{tabular}
\end{table}

\begin{table}[ht]
\centering
  \fontsize{8}{9.5}\selectfont
  \caption{\footnotesize Mean and standard deviation of $\mathrm{pass}@K$ for Gemini-2.0-Flash-Lite-001 ($K \in \{1,2,4\}$) by error category (left) and paper category (right). Detailed evaluations results for Table~\ref{tab:main-results}.}
  \label{tab:passk-openrouter_google_gemini-2_0-flash-lite-001}
  \begin{tabular}{@{} lccc @{\quad} lccc @{} }
    \toprule
    \multicolumn{4}{c}{\textbf{Error Category}} & \multicolumn{4}{c}{\textbf{Paper Category}} \\
    \cmidrule(r){1-4} \cmidrule(l){5-8}
    \multicolumn{1}{c}{Category} & pass@1 & pass@2 & pass@4 & \multicolumn{1}{c}{Category} & pass@1 & pass@2 & pass@4 \\
    \midrule
    Data Inconsistency & $1.8_{3.1}$ & $3.5_{4.0}$ & $7.0_{4.7}$ & Biology & $3.7_{3.8}$ & $7.6_{4.9}$ & $15.4_{5.7}$ \\
    Equation / proof & $0.0_{0.0}$ & $0.0_{0.0}$ & $0.0_{0.0}$ & Chemistry & $2.1_{5.5}$ & $4.2_{7.2}$ & $8.1_{8.3}$ \\
    Experiment setup & $0.0_{0.0}$ & $0.0_{0.0}$ & $0.0_{0.0}$ & Computer Science & $0.0_{0.0}$ & $0.0_{0.0}$ & $0.0_{0.0}$ \\
    Figure duplication & $3.2_{2.9}$ & $6.3_{3.7}$ & $12.9_{4.2}$ & Engineering & $0.0_{0.0}$ & $0.0_{0.0}$ & $0.0_{0.0}$ \\
    Reagent identity & $3.9_{10.7}$ & $8.7_{14.7}$ & $16.9_{16.7}$ & Environmental Science & $0.0_{0.0}$ & $0.0_{0.0}$ & $0.0_{0.0}$ \\
    Statistical reporting & $0.0_{0.0}$ & $0.0_{0.0}$ & $0.0_{0.0}$ & Materials Science & $2.1_{5.5}$ & $4.1_{7.2}$ & $8.3_{8.3}$ \\
    & & & & Mathematics & $0.0_{0.0}$ & $0.0_{0.0}$ & $0.0_{0.0}$ \\
    & & & & Medicine & $3.1_{8.2}$ & $6.6_{11.0}$ & $12.6_{12.5}$ \\
    & & & & Multidisciplinary & $3.7_{4.8}$ & $7.2_{6.3}$ & $14.9_{7.4}$ \\
    & & & & Physics & $0.0_{0.0}$ & $0.0_{0.0}$ & $0.0_{0.0}$ \\
    \bottomrule
  \end{tabular}
\end{table}

\begin{table}[ht]
\centering
  \fontsize{8}{9.5}\selectfont
  \caption{\footnotesize Mean and standard deviation of $\mathrm{pass}@K$ for Claude-3.7-Sonnet:Thinking ($K \in \{1,2,4\}$) by error category (left) and paper category (right). Detailed evaluations results for Table~\ref{tab:main-results}.}
  \label{tab:passk-claude_think}
  \begin{tabular}{@{} lccc @{\quad} lccc @{} }
    \toprule
    \multicolumn{4}{c}{\textbf{Error Category}} & \multicolumn{4}{c}{\textbf{Paper Category}} \\
    \cmidrule(r){1-4} \cmidrule(l){5-8}
    \multicolumn{1}{c}{Category} & pass@1 & pass@2 & pass@4 & \multicolumn{1}{c}{Category} & pass@1 & pass@2 & pass@4 \\
    \midrule
    Data Inconsistency & $8.0_{4.2}$ & $15.7_{5.8}$ & $29.3_{6.9}$ & Biology & $13.4_{7.3}$ & $23.9_{9.9}$ & $41.2_{10.6}$ \\
    Equation / proof & $0.8_{1.3}$ & $1.6_{1.8}$ & $3.0_{2.0}$ & Chemistry & $2.1_{5.5}$ & $4.1_{7.2}$ & $8.3_{8.3}$ \\
    Experiment setup & $0.0_{0.0}$ & $0.0_{0.0}$ & $0.0_{0.0}$ & Computer Science & $0.0_{0.0}$ & $0.0_{0.0}$ & $0.0_{0.0}$ \\
    Figure duplication & $10.6_{3.5}$ & $19.7_{4.9}$ & $34.9_{6.0}$ & Engineering & $6.2_{16.5}$ & $12.0_{21.4}$ & $24.8_{25.0}$ \\
    Reagent identity & $3.9_{10.8}$ & $7.5_{13.9}$ & $16.8_{16.7}$ & Environmental Science & $16.4_{23.2}$ & $33.1_{29.8}$ & $59.8_{29.3}$ \\
    Statistical reporting & $3.1_{8.3}$ & $6.1_{10.7}$ & $12.5_{12.5}$ & Materials Science & $10.4_{11.4}$ & $20.8_{14.3}$ & $38.4_{14.5}$ \\
    & & & & Mathematics & $0.6_{1.6}$ & $1.3_{2.2}$ & $2.5_{2.5}$ \\
    & & & & Medicine & $6.2_{10.8}$ & $12.1_{14.2}$ & $24.9_{16.4}$ \\
    & & & & Multidisciplinary & $10.0_{10.0}$ & $19.2_{13.2}$ & $35.8_{14.9}$ \\
    & & & & Physics & $0.0_{0.0}$ & $0.0_{0.0}$ & $0.0_{0.0}$ \\
    \bottomrule
  \end{tabular}
\end{table}

\begin{table}[ht]
\centering
  \fontsize{8}{9.5}\selectfont
  \caption{\footnotesize Mean and standard deviation of $\mathrm{pass}@K$ for Claude-3.7-Sonnet ($K \in \{1,2,4\}$) by error category (left) and paper category (right). Detailed evaluations results for Table~\ref{tab:main-results}.}
  \label{tab:passk-openrouter_anthropic_claude-3_7-sonnet_}
  \begin{tabular}{@{} lccc @{\quad} lccc @{} }
    \toprule
    \multicolumn{4}{c}{\textbf{Error Category}} & \multicolumn{4}{c}{\textbf{Paper Category}} \\
    \cmidrule(r){1-4} \cmidrule(l){5-8}
    \multicolumn{1}{c}{Category} & pass@1 & pass@2 & pass@4 & \multicolumn{1}{c}{Category} & pass@1 & pass@2 & pass@4 \\
    \midrule
    Data Inconsistency & $7.0_{6.2}$ & $13.6_{7.7}$ & $25.4_{8.1}$ & Biology & $10.6_{5.3}$ & $17.8_{5.5}$ & $27.3_{5.0}$ \\
    Equation / proof & $0.4_{1.0}$ & $0.7_{1.3}$ & $1.5_{1.5}$ & Chemistry & $4.2_{7.2}$ & $8.2_{9.3}$ & $16.4_{10.9}$ \\
    Experiment setup & $0.0_{0.0}$ & $0.0_{0.0}$ & $0.0_{0.0}$ & Computer Science & $0.0_{0.0}$ & $0.0_{0.0}$ & $0.0_{0.0}$ \\
    Figure duplication & $8.8_{4.8}$ & $15.5_{4.8}$ & $25.1_{4.0}$ & Engineering & $0.0_{0.0}$ & $0.0_{0.0}$ & $0.0_{0.0}$ \\
    Reagent identity & $8.6_{14.6}$ & $15.4_{16.6}$ & $26.1_{13.8}$ & Environmental Science & $3.9_{10.8}$ & $7.5_{13.9}$ & $16.8_{16.7}$ \\
    Statistical reporting & $0.0_{0.0}$ & $0.0_{0.0}$ & $0.0_{0.0}$ & Materials Science & $8.2_{8.3}$ & $15.9_{10.2}$ & $30.1_{11.2}$ \\
    & & & & Mathematics & $0.0_{0.0}$ & $0.0_{0.0}$ & $0.0_{0.0}$ \\
    & & & & Medicine & $3.4_{8.5}$ & $6.4_{10.9}$ & $12.8_{12.5}$ \\
    & & & & Multidisciplinary & $13.8_{11.2}$ & $25.4_{12.7}$ & $42.6_{10.9}$ \\
    & & & & Physics & $0.0_{0.0}$ & $0.0_{0.0}$ & $0.0_{0.0}$ \\
    \bottomrule
  \end{tabular}
\end{table}

\begin{table}[ht]
  \centering
  \fontsize{8}{9.5}\selectfont
  \caption{\footnotesize Mean and standard deviation of $\mathrm{pass}@K$ for Qwen2.5-VL-72B-instruct ($K \in \{1,2,4\}$) by error category (left) and paper category (right). Detailed evaluations results for Table~\ref{tab:main-results}.}
  \label{tab:passk-openrouter_qwen_qwen-2_5-vl-72b-instruct}
  \begin{tabular}{@{} lccc @{\quad} lccc @{} }
    \toprule
    \multicolumn{4}{c}{\textbf{Error Category}} & \multicolumn{4}{c}{\textbf{Paper Category}} \\
    \cmidrule(r){1-4} \cmidrule(l){5-8}
   \multicolumn{1}{c}{Category} & pass@1 & pass@2 & pass@4 & \multicolumn{1}{c}{Category} & pass@1 & pass@2 & pass@4 \\
    \midrule
    Data Inconsistency & $0.4_{1.7}$ & $0.9_{2.4}$ & $1.8_{3.1}$ & Biology & $1.5_{3.1}$ & $3.0_{4.1}$ & $6.2_{5.7}$ \\
    Equation / proof & $0.0_{0.0}$ & $0.0_{0.0}$ & $0.0_{0.0}$ & Chemistry & $0.9_{3.9}$ & $2.1_{5.5}$ & $4.3_{7.3}$ \\
    Experiment setup & $0.0_{0.0}$ & $0.0_{0.0}$ & $0.0_{0.0}$ & Computer Science & $0.0_{0.0}$ & $0.0_{0.0}$ & $0.0_{0.0}$ \\
    Figure duplication & $1.0_{1.6}$ & $1.9_{2.2}$ & $3.9_{3.0}$ & Engineering & $3.1_{12.1}$ & $6.6_{16.9}$ & $12.8_{21.8}$ \\
    Reagent identity & $0.0_{0.0}$ & $0.0_{0.0}$ & $0.0_{0.0}$ & Environmental Science & $0.0_{0.0}$ & $0.0_{0.0}$ & $0.0_{0.0}$ \\
    Statistical reporting & $0.0_{0.0}$ & $0.0_{0.0}$ & $0.0_{0.0}$ & Materials Science & $0.0_{0.0}$ & $0.0_{0.0}$ & $0.0_{0.0}$ \\
    & & & & Mathematics & $0.0_{0.0}$ & $0.0_{0.0}$ & $0.0_{0.0}$ \\
    & & & & Medicine & $0.0_{0.0}$ & $0.0_{0.0}$ & $0.0_{0.0}$ \\
    & & & & Multidisciplinary & $0.0_{0.0}$ & $0.0_{0.0}$ & $0.0_{0.0}$ \\
    & & & & Physics & $0.0_{0.0}$ & $0.0_{0.0}$ & $0.0_{0.0}$ \\
    \bottomrule
  \end{tabular}
\end{table}

\begin{table}[ht]
\centering
  \fontsize{8}{9.5}\selectfont
  \caption{\footnotesize Mean and standard deviation of $\mathrm{pass}@K$ for Qwen2.5-VL-32B-instruct ($K \in \{1,2,4\}$) by error category (left) and paper category (right). Detailed evaluations results for Table~\ref{tab:main-results}.}
  \label{tab:passk-openrouter_qwen_qwen2_5-vl-32b-instruct}
  \begin{tabular}{@{} lccc @{\quad} lccc @{} }
    \toprule
    \multicolumn{4}{c}{\textbf{Error Category}} & \multicolumn{4}{c}{\textbf{Paper Category}} \\
    \cmidrule(r){1-4} \cmidrule(l){5-8}
   \multicolumn{1}{c}{Category} & pass@1 & pass@2 & pass@4 & \multicolumn{1}{c}{Category} & pass@1 & pass@2 & pass@4 \\
    \midrule
    Data Inconsistency & $0.9_{2.4}$ & $1.8_{3.1}$ & $3.5_{3.6}$ & Biology & $9.7_{8.5}$ & $16.4_{8.3}$ & $25.5_{8.1}$ \\
    Equation / proof & $0.0_{0.0}$ & $0.0_{0.0}$ & $0.0_{0.0}$ & Chemistry & $6.4_{11.6}$ & $12.0_{14.0}$ & $21.1_{13.3}$ \\
    Experiment setup & $0.0_{0.0}$ & $0.0_{0.0}$ & $0.0_{0.0}$ & Computer Science & $0.0_{0.0}$ & $0.0_{0.0}$ & $0.0_{0.0}$ \\
    Figure duplication & $4.7_{4.5}$ & $7.9_{4.9}$ & $12.2_{4.8}$ & Engineering & $0.0_{0.0}$ & $0.0_{0.0}$ & $0.0_{0.0}$ \\
    Reagent identity & $8.0_{14.3}$ & $16.1_{16.7}$ & $26.6_{13.4}$ & Environmental Science & $0.0_{0.0}$ & $0.0_{0.0}$ & $0.0_{0.0}$ \\
    Statistical reporting & $0.0_{0.0}$ & $0.0_{0.0}$ & $0.0_{0.0}$ & Materials Science & $0.0_{0.0}$ & $0.0_{0.0}$ & $0.0_{0.0}$ \\
    & & & & Mathematics & $0.0_{0.0}$ & $0.0_{0.0}$ & $0.0_{0.0}$ \\
    & & & & Medicine & $0.0_{0.0}$ & $0.0_{0.0}$ & $0.0_{0.0}$ \\
    & & & & Multidisciplinary & $0.0_{0.0}$ & $0.0_{0.0}$ & $0.0_{0.0}$ \\
    & & & & Physics & $0.0_{0.0}$ & $0.0_{0.0}$ & $0.0_{0.0}$ \\
    \bottomrule
  \end{tabular}
\end{table}

\begin{table}[ht]
\centering
  \fontsize{8}{9.5}\selectfont
  \caption{\footnotesize Mean and standard deviation of $\mathrm{pass}@K$ for Llama-4-Maverick ($K \in \{1,2,4\}$) by error category (left) and paper category (right). Detailed evaluations results for Table~\ref{tab:main-results}.}
  \label{tab:passk-openrouter_meta-llama_llama-4-m}
  \begin{tabular}{@{} lccc @{\quad} lccc @{} }
    \toprule
    \multicolumn{4}{c}{\textbf{Error Category}} & \multicolumn{4}{c}{\textbf{Paper Category}} \\
    \cmidrule(r){1-4} \cmidrule(l){5-8}
   \multicolumn{1}{c}{Category} & pass@1 & pass@2 & pass@4 & \multicolumn{1}{c}{Category} & pass@1 & pass@2 & pass@4 \\
    \midrule
    Data Inconsistency & $0.8_{2.3}$ & $1.9_{3.1}$ & $3.6_{3.6}$ & Biology & $2.9_{5.4}$ & $5.4_{6.4}$ & $9.9_{6.1}$ \\
    Equation / proof & $0.0_{0.0}$ & $0.0_{0.0}$ & $0.0_{0.0}$ & Chemistry & $2.2_{5.7}$ & $4.3_{7.3}$ & $8.5_{8.3}$ \\
    Experiment setup & $0.0_{0.0}$ & $0.0_{0.0}$ & $0.0_{0.0}$ & Computer Science & $0.0_{0.0}$ & $0.0_{0.0}$ & $0.0_{0.0}$ \\
    Figure duplication & $1.9_{2.7}$ & $3.6_{3.2}$ & $6.6_{3.1}$ & Engineering & $0.0_{0.0}$ & $0.0_{0.0}$ & $0.0_{0.0}$ \\
    Reagent identity & $4.1_{11.0}$ & $8.0_{14.2}$ & $16.5_{16.7}$ & Environmental Science & $0.0_{0.0}$ & $0.0_{0.0}$ & $0.0_{0.0}$ \\
    Statistical reporting & $0.0_{0.0}$ & $0.0_{0.0}$ & $0.0_{0.0}$ & Materials Science & $4.2_{7.2}$ & $8.7_{9.7}$ & $16.7_{11.1}$ \\
    & & & & Mathematics & $0.0_{0.0}$ & $0.0_{0.0}$ & $0.0_{0.0}$ \\
    & & & & Medicine & $0.0_{0.0}$ & $0.0_{0.0}$ & $0.0_{0.0}$ \\
    & & & & Multidisciplinary & $0.0_{0.0}$ & $0.0_{0.0}$ & $0.0_{0.0}$ \\
    & & & & Physics & $0.0_{0.0}$ & $0.0_{0.0}$ & $0.0_{0.0}$ \\
    \bottomrule
  \end{tabular}
\end{table}

\begin{table}[ht]
   \centering
  \fontsize{8}{9.5}\selectfont
  \caption{\footnotesize Mean and standard deviation of $\mathrm{pass}@K$ for Llama-4-Scout ($K \in \{1,2,4\}$) by error category (left) and paper category (right). Detailed evaluations results for Table~\ref{tab:main-results}.}
  \label{tab:passk-openrouter_meta-llama_llama-4-s}
  \begin{tabular}{@{} lccc @{\quad} lccc @{} }
    \toprule
    \multicolumn{4}{c}{\textbf{Error Category}} & \multicolumn{4}{c}{\textbf{Paper Category}} \\
    \cmidrule(r){1-4} \cmidrule(l){5-8}
   \multicolumn{1}{c}{Category} & pass@1 & pass@2 & pass@4 & \multicolumn{1}{c}{Category} & pass@1 & pass@2 & pass@4 \\
    \midrule
    Data Inconsistency & $2.6_{3.4}$ & $4.9_{4.1}$ & $9.2_{4.2}$ & Biology & $6.5_{6.0}$ & $12.8_{7.9}$ & $25.3_{9.6}$ \\
    Equation / proof & $0.0_{0.0}$ & $0.0_{0.0}$ & $0.0_{0.0}$ & Chemistry & $6.1_{8.0}$ & $11.5_{9.5}$ & $21.4_{9.7}$ \\
    Experiment setup & $0.0_{0.0}$ & $0.0_{0.0}$ & $0.0_{0.0}$ & Computer Science & $0.0_{0.0}$ & $0.0_{0.0}$ & $0.0_{0.0}$ \\
    Figure duplication & $3.6_{4.9}$ & $7.0_{6.4}$ & $14.0_{7.6}$ & Engineering & $6.2_{16.5}$ & $12.2_{21.5}$ & $24.9_{25.0}$ \\
    Reagent identity & $3.9_{10.7}$ & $8.7_{14.7}$ & $16.9_{16.7}$ & Environmental Science & $0.0_{0.0}$ & $0.0_{0.0}$ & $0.0_{0.0}$ \\
    Statistical reporting & $3.1_{8.3}$ & $6.1_{10.7}$ & $12.5_{12.5}$ & Materials Science & $0.0_{0.0}$ & $0.0_{0.0}$ & $0.0_{0.0}$ \\
    & & & & Mathematics & $0.0_{0.0}$ & $0.0_{0.0}$ & $0.0_{0.0}$ \\
    & & & & Medicine & $3.1_{8.3}$ & $6.1_{10.7}$ & $12.5_{12.5}$ \\
    & & & & Multidisciplinary & $1.2_{3.3}$ & $2.4_{4.3}$ & $5.0_{5.0}$ \\
    & & & & Physics & $0.0_{0.0}$ & $0.0_{0.0}$ & $0.0_{0.0}$ \\
    \bottomrule
  \end{tabular}
\end{table}

%% file: tables/add_results_text.tex
\begin{table}[ht]
  \centering
  \fontsize{8}{9.5}\selectfont
  \caption{\footnotesize Mean and standard deviation of $\mathrm{pass}@K$ for o3 ($K \in \{1,2,4\}$) by error category (left) and paper category (right). Detailed evaluation results for text-only evaluation of Table~\ref{tab:multi_modality}.}
  \label{tab:passk-text-o3}
  \begin{tabular}{@{} lccc @{\quad} lccc @{} }
    \toprule
    \multicolumn{4}{c}{\textbf{Error Category}} & \multicolumn{4}{c}{\textbf{Paper Category}} \\
    \cmidrule(r){1-4} \cmidrule(l){5-8}
   \multicolumn{1}{c}{Category} & pass@1 & pass@2 & pass@4 & \multicolumn{1}{c}{Category} & pass@1 & pass@2 & pass@4 \\
    \midrule
    Data Inconsistency & $14.6_{5.5}$ & $25.2_{8.3}$ & $38.0_{9.7}$ & Biology & $6.2_{16.5}$ & $13.1_{22.0}$ & $25.1_{25.0}$ \\
    Equation / proof & $24.6_{4.9}$ & $41.3_{5.5}$ & $62.8_{5.7}$ & Computer Science & $24.0_{7.8}$ & $42.8_{8.9}$ & $66.8_{7.5}$ \\
    Experiment setup & $6.7_{17.0}$ & $12.9_{21.9}$ & $24.8_{25.0}$ & Environmental Science & $12.0_{21.4}$ & $24.1_{25.0}$ & $40.0_{20.0}$ \\
    Reagent identity & $8.3_{14.4}$ & $17.1_{19.0}$ & $33.0_{21.7}$ & Materials Science & $0.0_{0.0}$ & $0.0_{0.0}$ & $0.0_{0.0}$ \\
    Statistical reporting & $24.8_{17.8}$ & $44.1_{18.8}$ & $66.1_{12.4}$ & Mathematics & $21.8_{5.5}$ & $37.7_{7.9}$ & $58.9_{8.4}$ \\
    & & & & Medicine & $12.4_{33.0}$ & $25.1_{43.4}$ & $48.7_{50.0}$ \\
    & & & & Multidisciplinary & $33.3_{0.0}$ & $41.7_{14.5}$ & $49.6_{16.7}$ \\
    & & & & Physics & $27.4_{14.4}$ & $46.0_{14.5}$ & $67.6_{10.5}$ \\
    \bottomrule
  \end{tabular}
\end{table}

\begin{table}[ht]
  \centering
  \fontsize{8}{9.5}\selectfont
  \caption{\footnotesize Mean and standard deviation of $\mathrm{pass}@K$ for GPT-4.1 ($K \in \{1,2,4\}$) by error category (left) and paper category (right). Detailed evaluation results for text-only evaluation of Table~\ref{tab:multi_modality}.}
  \label{tab:passk-text-gpt-4.1}
  \begin{tabular}{@{} lccc @{\quad} lccc @{} }
    \toprule
    \multicolumn{4}{c}{\textbf{Error Category}} & \multicolumn{4}{c}{\textbf{Paper Category}} \\
    \cmidrule(r){1-4} \cmidrule(l){5-8}
   \multicolumn{1}{c}{Category} & pass@1 & pass@2 & pass@4 & \multicolumn{1}{c}{Category} & pass@1 & pass@2 & pass@4 \\
    \midrule
    Data Inconsistency & $8.6_{14.6}$ & $15.7_{16.6}$ & $26.1_{13.8}$ & Biology & $6.2_{16.5}$ & $13.1_{22.0}$ & $25.1_{25.0}$ \\
    Equation / proof & $6.0_{3.4}$ & $10.8_{3.6}$ & $18.1_{3.1}$ & Computer Science & $4.1_{4.2}$ & $7.9_{5.2}$ & $14.8_{5.4}$ \\
    Experiment setup & $0.0_{0.0}$ & $0.0_{0.0}$ & $0.0_{0.0}$ & Environmental Science & $19.6_{35.6}$ & $36.4_{42.2}$ & $64.6_{40.1}$ \\
    Reagent identity & $8.3_{14.4}$ & $16.7_{19.1}$ & $33.2_{21.9}$ & Materials Science & $0.0_{0.0}$ & $0.0_{0.0}$ & $0.0_{0.0}$ \\
    Statistical reporting & $15.6_{12.1}$ & $22.3_{7.8}$ & $25.0_{0.0}$ & Mathematics & $4.3_{3.9}$ & $8.0_{4.3}$ & $13.5_{3.8}$ \\
    & & & & Medicine & $12.4_{33.0}$ & $24.0_{42.7}$ & $49.5_{50.0}$ \\
    & & & & Multidisciplinary & $45.9_{16.2}$ & $71.3_{17.4}$ & $92.2_{14.1}$ \\
    & & & & Physics & $0.0_{0.0}$ & $0.0_{0.0}$ & $0.0_{0.0}$ \\
    \bottomrule
  \end{tabular}
\end{table}

\begin{table}[ht]
  \centering
  \fontsize{8}{9.5}\selectfont
  \caption{\footnotesize Mean and standard deviation of $\mathrm{pass}@K$ for Gemini-2.5-Pro ($K \in \{1,2,4\}$) by error category (left) and paper category (right). Detailed evaluation results for text-only evaluation of Table~\ref{tab:multi_modality}.}
  \label{tab:passk-text-openrouter_google_gemini-2_5}
  \begin{tabular}{@{} lccc @{\quad} lccc @{} }
    \toprule
    \multicolumn{4}{c}{\textbf{Error Category}} & \multicolumn{4}{c}{\textbf{Paper Category}} \\
    \cmidrule(r){1-4} \cmidrule(l){5-8}
   \multicolumn{1}{c}{Category} & pass@1 & pass@2 & pass@4 & \multicolumn{1}{c}{Category} & pass@1 & pass@2 & pass@4 \\
    \midrule
    Data Inconsistency & $4.1_{7.2}$ & $7.7_{8.3}$ & $13.0_{6.9}$ & Biology & $0.0_{0.0}$ & $0.0_{0.0}$ & $0.0_{0.0}$ \\
    Equation / proof & $7.6_{4.0}$ & $11.6_{3.9}$ & $16.2_{3.8}$ & Computer Science & $4.2_{5.9}$ & $8.2_{7.5}$ & $14.8_{8.5}$ \\
    Experiment setup & $0.0_{0.0}$ & $0.0_{0.0}$ & $0.0_{0.0}$ & Environmental Science & $12.9_{21.9}$ & $23.5_{25.0}$ & $38.6_{21.0}$ \\
    Reagent identity & $8.3_{14.4}$ & $15.4_{16.6}$ & $26.1_{13.8}$ & Materials Science & $0.0_{0.0}$ & $0.0_{0.0}$ & $0.0_{0.0}$ \\
    Statistical reporting & $12.1_{12.5}$ & $19.6_{10.3}$ & $24.7_{2.8}$ & Mathematics & $5.0_{2.5}$ & $7.1_{2.5}$ & $9.0_{2.0}$ \\
    & & & & Medicine & $24.8_{43.2}$ & $46.3_{49.9}$ & $78.2_{41.3}$ \\
    & & & & Multidisciplinary & $49.4_{16.7}$ & $73.6_{20.2}$ & $92.3_{14.1}$ \\
    & & & & Physics & $0.0_{0.0}$ & $0.0_{0.0}$ & $0.0_{0.0}$ \\
    \bottomrule
  \end{tabular}
\end{table}

\begin{table}[ht]
  \centering
  \fontsize{8}{9.5}\selectfont
  \caption{\footnotesize Mean and standard deviation of $\mathrm{pass}@K$ for Gemini-2.0-Flash-Lite-001 ($K \in \{1,2,4\}$) by error category (left) and paper category (right). Detailed evaluation results for text-only evaluation of Table~\ref{tab:multi_modality}.}
  \label{tab:passk-text-openrouter_google_gemini-2_0-flash-lite-001}
  \begin{tabular}{@{} lccc @{\quad} lccc @{} }
    \toprule
    \multicolumn{4}{c}{\textbf{Error Category}} & \multicolumn{4}{c}{\textbf{Paper Category}} \\
    \cmidrule(r){1-4} \cmidrule(l){5-8}
   \multicolumn{1}{c}{Category} & pass@1 & pass@2 & pass@4 & \multicolumn{1}{c}{Category} & pass@1 & pass@2 & pass@4 \\
    \midrule
    Data Inconsistency & $2.1_{5.5}$ & $4.0_{7.1}$ & $8.2_{8.3}$ & Biology & $12.9_{21.9}$ & $25.4_{28.6}$ & $49.9_{32.9}$ \\
    Equation / proof & $1.1_{1.5}$ & $2.1_{1.7}$ & $3.9_{1.8}$ & Computer Science & $2.1_{3.6}$ & $3.8_{4.2}$ & $6.5_{3.5}$ \\
    Experiment setup & $0.0_{0.0}$ & $0.0_{0.0}$ & $0.0_{0.0}$ & Environmental Science & $0.0_{0.0}$ & $0.0_{0.0}$ & $0.0_{0.0}$ \\
    Reagent identity & $12.5_{16.1}$ & $25.7_{20.9}$ & $50.1_{24.2}$ & Materials Science & $0.0_{0.0}$ & $0.0_{0.0}$ & $0.0_{0.0}$ \\
    Statistical reporting & $3.4_{8.5}$ & $6.5_{10.9}$ & $12.4_{12.5}$ & Mathematics & $0.6_{1.6}$ & $1.1_{2.1}$ & $2.5_{2.5}$ \\
    & & & & Medicine & $11.7_{32.2}$ & $26.2_{44.0}$ & $50.7_{50.0}$ \\
    & & & & Multidisciplinary & $8.6_{14.6}$ & $16.6_{19.3}$ & $33.0_{22.2}$ \\
    & & & & Physics & $0.0_{0.0}$ & $0.0_{0.0}$ & $0.0_{0.0}$ \\
    \bottomrule
  \end{tabular}
\end{table}

\begin{table}[ht]
  \centering
  \fontsize{8}{9.5}\selectfont
  \caption{\footnotesize Mean and standard deviation of $\mathrm{pass}@K$ for Claude-3.7-Sonnet:Thinking ($K \in \{1,2,4\}$) by error category (left) and paper category (right). Detailed evaluation results for text-only evaluation of Table~\ref{tab:multi_modality}.}
  \label{tab:passk-text-openrouter_anthropic_claude-3_7-sonnet:t}
  \begin{tabular}{@{} lccc @{\quad} lccc @{} }
    \toprule
    \multicolumn{4}{c}{\textbf{Error Category}} & \multicolumn{4}{c}{\textbf{Paper Category}} \\
    \cmidrule(r){1-4} \cmidrule(l){5-8}
   \multicolumn{1}{c}{Category} & pass@1 & pass@2 & pass@4 & \multicolumn{1}{c}{Category} & pass@1 & pass@2 & pass@4 \\
    \midrule
    Data Inconsistency & $6.2_{11.6}$ & $12.1_{14.0}$ & $21.1_{13.3}$ & Biology & $0.0_{0.0}$ & $0.0_{0.0}$ & $0.0_{0.0}$ \\
    Equation / proof & $4.6_{3.7}$ & $8.4_{4.6}$ & $15.1_{4.9}$ & Computer Science & $7.2_{6.5}$ & $12.4_{7.8}$ & $20.7_{8.9}$ \\
    Experiment setup & $0.0_{0.0}$ & $0.0_{0.0}$ & $0.0_{0.0}$ & Environmental Science & $12.4_{21.6}$ & $23.8_{25.0}$ & $39.1_{20.7}$ \\
    Reagent identity & $0.0_{0.0}$ & $0.0_{0.0}$ & $0.0_{0.0}$ & Materials Science & $6.2_{16.5}$ & $12.6_{21.7}$ & $24.3_{25.0}$ \\
    Statistical reporting & $18.4_{20.6}$ & $31.2_{19.9}$ & $44.3_{11.6}$ & Mathematics & $4.5_{3.9}$ & $8.0_{4.3}$ & $13.6_{3.7}$ \\
    & & & & Medicine & $0.0_{0.0}$ & $0.0_{0.0}$ & $0.0_{0.0}$ \\
    & & & & Multidisciplinary & $8.0_{14.3}$ & $15.8_{16.7}$ & $26.4_{13.5}$ \\
    & & & & Physics & $4.2_{7.2}$ & $7.7_{8.3}$ & $13.0_{6.9}$ \\
    \bottomrule
  \end{tabular}
\end{table}

\begin{table}[ht]
  \centering
  \fontsize{8}{9.5}\selectfont
  \caption{\footnotesize Mean and standard deviation of $\mathrm{pass}@K$ for Claude-3.7-Sonnet ($K \in \{1,2,4\}$) by error category (left) and paper category (right). Detailed evaluation results for text-only evaluation of Table~\ref{tab:multi_modality}.}
  \label{tab:passk-text-openrouter_anthropic_claude-3_7-sonnet_}
  \begin{tabular}{@{} lccc @{\quad} lccc @{} }
    \toprule
    \multicolumn{4}{c}{\textbf{Error Category}} & \multicolumn{4}{c}{\textbf{Paper Category}} \\
    \cmidrule(r){1-4} \cmidrule(l){5-8}
   \multicolumn{1}{c}{Category} & pass@1 & pass@2 & pass@4 & \multicolumn{1}{c}{Category} & pass@1 & pass@2 & pass@4 \\
    \midrule
    Data Inconsistency & $6.3_{8.1}$ & $11.6_{10.1}$ & $21.5_{10.0}$ & Biology & $6.2_{16.5}$ & $12.0_{21.4}$ & $24.8_{25.0}$ \\
    Equation / proof & $4.9_{2.1}$ & $8.9_{3.0}$ & $15.2_{3.4}$ & Computer Science & $6.3_{5.5}$ & $11.8_{6.9}$ & $21.3_{7.0}$ \\
    Experiment setup & $0.0_{0.0}$ & $0.0_{0.0}$ & $0.0_{0.0}$ & Environmental Science & $18.4_{24.1}$ & $34.4_{29.3}$ & $63.4_{29.3}$ \\
    Reagent identity & $4.1_{11.0}$ & $8.0_{14.2}$ & $16.5_{16.7}$ & Materials Science & $0.0_{0.0}$ & $0.0_{0.0}$ & $0.0_{0.0}$ \\
    Statistical reporting & $3.4_{8.5}$ & $6.5_{10.9}$ & $12.4_{12.5}$ & Mathematics & $3.8_{4.2}$ & $6.4_{4.2}$ & $9.9_{3.3}$ \\
    & & & & Medicine & $0.0_{0.0}$ & $0.0_{0.0}$ & $0.0_{0.0}$ \\
    & & & & Multidisciplinary & $8.9_{14.8}$ & $17.2_{18.9}$ & $33.5_{21.2}$ \\
    & & & & Physics & $0.0_{0.0}$ & $0.0_{0.0}$ & $0.0_{0.0}$ \\
    \bottomrule
  \end{tabular}
\end{table}

\begin{table}[ht]
  \centering
  \fontsize{8}{9.5}\selectfont
  \caption{\footnotesize Mean and standard deviation of $\mathrm{pass}@K$ for DeepSeek-R1 ($K \in \{1,2,4\}$) by error category (left) and paper category (right). Detailed evaluation results for text-only evaluation of Table~\ref{tab:multi_modality}.}
  \label{tab:passk-text-openrouter_deepseek_deepseek-r1}
  \begin{tabular}{@{} lccc @{\quad} lccc @{} }
    \toprule
    \multicolumn{4}{c}{\textbf{Error Category}} & \multicolumn{4}{c}{\textbf{Paper Category}} \\
    \cmidrule(r){1-4} \cmidrule(l){5-8}
   \multicolumn{1}{c}{Category} & pass@1 & pass@2 & pass@4 & \multicolumn{1}{c}{Category} & pass@1 & pass@2 & pass@4 \\
    \midrule
    Data Inconsistency & $12.1_{11.7}$ & $20.5_{11.5}$ & $30.2_{6.5}$ & Biology & $7.4_{17.8}$ & $14.3_{22.6}$ & $28.9_{24.7}$ \\
    Equation / proof & $16.1_{5.5}$ & $27.8_{6.0}$ & $41.5_{4.3}$ & Computer Science & $13.0_{9.7}$ & $23.8_{10.3}$ & $39.1_{7.7}$ \\
    Experiment setup & $0.0_{0.0}$ & $0.0_{0.0}$ & $0.0_{0.0}$ & Environmental Science & $15.7_{23.2}$ & $26.9_{24.9}$ & $41.9_{18.5}$ \\
    Reagent identity & $4.9_{11.8}$ & $9.5_{15.1}$ & $19.3_{16.5}$ & Materials Science & $0.0_{0.0}$ & $0.0_{0.0}$ & $0.0_{0.0}$ \\
    Statistical reporting & $28.3_{15.7}$ & $43.3_{19.8}$ & $61.3_{18.1}$ & Mathematics & $15.9_{4.1}$ & $26.9_{5.2}$ & $39.7_{5.0}$ \\
    & & & & Medicine & $0.0_{0.0}$ & $0.0_{0.0}$ & $0.0_{0.0}$ \\
    & & & & Multidisciplinary & $38.4_{20.8}$ & $55.9_{15.6}$ & $65.8_{5.3}$ \\
    & & & & Physics & $16.3_{18.0}$ & $28.1_{17.3}$ & $42.2_{9.7}$ \\
    \bottomrule
  \end{tabular}
\end{table}

\begin{table}[ht]
  \centering
  \fontsize{8}{9.5}\selectfont
  \caption{\footnotesize Mean and standard deviation of $\mathrm{pass}@K$ for DeepSeek-V3-0324 ($K \in \{1,2,4\}$) by error category (left) and paper category (right). Detailed evaluation results for text-only evaluation of Table~\ref{tab:multi_modality}.}
  \label{tab:passk-text-openrouter_deepseek_deepseek-chat}
  \begin{tabular}{@{} lccc @{\quad} lccc @{} }
    \toprule
    \multicolumn{4}{c}{\textbf{Error Category}} & \multicolumn{4}{c}{\textbf{Paper Category}} \\
    \cmidrule(r){1-4} \cmidrule(l){5-8}
   \multicolumn{1}{c}{Category} & pass@1 & pass@2 & pass@4 & \multicolumn{1}{c}{Category} & pass@1 & pass@2 & pass@4 \\
    \midrule
    Data Inconsistency & $7.0_{8.2}$ & $12.1_{7.5}$ & $16.2_{2.7}$ & Biology & $7.1_{17.5}$ & $14.2_{22.6}$ & $28.9_{24.7}$ \\
    Equation / proof & $1.3_{1.5}$ & $2.6_{2.0}$ & $5.2_{2.1}$ & Computer Science & $1.1_{2.8}$ & $2.2_{3.7}$ & $4.9_{4.1}$ \\
    Experiment setup & $0.0_{0.0}$ & $0.0_{0.0}$ & $0.0_{0.0}$ & Environmental Science & $29.3_{24.6}$ & $51.4_{27.0}$ & $76.8_{24.9}$ \\
    Reagent identity & $4.7_{11.6}$ & $9.5_{15.1}$ & $19.3_{16.5}$ & Materials Science & $0.0_{0.0}$ & $0.0_{0.0}$ & $0.0_{0.0}$ \\
    Statistical reporting & $0.0_{0.0}$ & $0.0_{0.0}$ & $0.0_{0.0}$ & Mathematics & $0.7_{1.7}$ & $1.4_{2.3}$ & $2.9_{2.5}$ \\
    & & & & Medicine & $0.0_{0.0}$ & $0.0_{0.0}$ & $0.0_{0.0}$ \\
    & & & & Multidisciplinary & $0.0_{0.0}$ & $0.0_{0.0}$ & $0.0_{0.0}$ \\
    & & & & Physics & $0.0_{0.0}$ & $0.0_{0.0}$ & $0.0_{0.0}$ \\
    \bottomrule
  \end{tabular}
\end{table}

\begin{table}[ht]
  \centering
  \fontsize{8}{9.5}\selectfont
  \caption{\footnotesize Mean and standard deviation of $\mathrm{pass}@K$ for Qwen3-235A-22B ($K \in \{1,2,4\}$) by error category (left) and paper category (right). Detailed evaluation results for text-only evaluation of Table~\ref{tab:multi_modality}.}
  \label{tab:passk-text-openrouter_qwen_qwen3-235b}
  \begin{tabular}{@{} lccc @{\quad} lccc @{} }
    \toprule
    \multicolumn{4}{c}{\textbf{Error Category}} & \multicolumn{4}{c}{\textbf{Paper Category}} \\
    \cmidrule(r){1-4} \cmidrule(l){5-8}
   \multicolumn{1}{c}{Category} & pass@1 & pass@2 & pass@4 & \multicolumn{1}{c}{Category} & pass@1 & pass@2 & pass@4 \\
    \midrule
    Data Inconsistency & $8.4_{8.3}$ & $15.8_{9.5}$ & $26.7_{8.2}$ & Biology & $0.0_{0.0}$ & $0.0_{0.0}$ & $0.0_{0.0}$ \\
    Equation / proof & $17.1_{6.9}$ & $28.1_{6.1}$ & $40.7_{3.3}$ & Computer Science & $17.9_{5.7}$ & $29.9_{7.4}$ & $43.2_{5.6}$ \\
    Experiment setup & $0.0_{0.0}$ & $0.0_{0.0}$ & $0.0_{0.0}$ & Environmental Science & $7.6_{18.0}$ & $16.0_{23.3}$ & $33.2_{23.6}$ \\
    Reagent identity & $5.7_{12.5}$ & $11.6_{15.9}$ & $22.3_{15.7}$ & Materials Science & $0.0_{0.0}$ & $0.0_{0.0}$ & $0.0_{0.0}$ \\
    Statistical reporting & $25.9_{20.6}$ & $44.4_{18.9}$ & $64.8_{12.3}$ & Mathematics & $12.7_{6.3}$ & $20.6_{6.3}$ & $31.2_{4.9}$ \\
    & & & & Medicine & $17.0_{37.6}$ & $34.8_{47.7}$ & $66.9_{47.1}$ \\
    & & & & Multidisciplinary & $45.2_{24.4}$ & $74.6_{25.0}$ & $98.2_{7.5}$ \\
    & & & & Physics & $16.6_{9.7}$ & $28.7_{12.8}$ & $43.2_{10.2}$ \\
    \bottomrule
  \end{tabular}
\end{table}

\begin{table}[ht]
  \centering
  \fontsize{8}{9.5}\selectfont
  \caption{\footnotesize Mean and standard deviation of $\mathrm{pass}@K$ for Qwen2.5-VL-72B-Instruct ($K \in \{1,2,4\}$) by error category (left) and paper category (right). Detailed evaluation results for text-only evaluation of Table~\ref{tab:multi_modality}.}
  \label{tab:passk-text-openrouter_qwen_qwen2_5-vl-72b-instruct}
  \begin{tabular}{@{} lccc @{\quad} lccc @{} }
    \toprule
    \multicolumn{4}{c}{\textbf{Error Category}} & \multicolumn{4}{c}{\textbf{Paper Category}} \\
    \cmidrule(r){1-4} \cmidrule(l){5-8}
   \multicolumn{1}{c}{Category} & pass@1 & pass@2 & pass@4 & \multicolumn{1}{c}{Category} & pass@1 & pass@2 & pass@4 \\
    \midrule
    Data Inconsistency & $18.9_{10.4}$ & $31.0_{9.1}$ & $42.2_{8.3}$ & Biology & $23.2_{38.1}$ & $41.6_{43.3}$ & $70.8_{36.6}$ \\
    Equation / proof & $0.4_{1.0}$ & $0.8_{1.3}$ & $1.8_{1.5}$ & Computer Science & $0.0_{0.0}$ & $0.0_{0.0}$ & $0.0_{0.0}$ \\
    Experiment setup & $0.0_{0.0}$ & $0.0_{0.0}$ & $0.0_{0.0}$ & Environmental Science & $34.4_{34.5}$ & $55.4_{31.7}$ & $79.6_{24.6}$ \\
    Reagent identity & $15.5_{25.4}$ & $27.7_{28.8}$ & $47.2_{24.4}$ & Materials Science & $8.2_{18.6}$ & $15.2_{23.0}$ & $28.1_{24.8}$ \\
    Statistical reporting & $6.9_{11.2}$ & $13.3_{12.5}$ & $21.2_{9.0}$ & Mathematics & $0.0_{0.0}$ & $0.0_{0.0}$ & $0.0_{0.0}$ \\
    & & & & Medicine & $0.0_{0.0}$ & $0.0_{0.0}$ & $0.0_{0.0}$ \\
    & & & & Multidisciplinary & $22.9_{15.5}$ & $41.3_{17.9}$ & $60.7_{12.8}$ \\
    & & & & Physics & $0.0_{0.0}$ & $0.0_{0.0}$ & $0.0_{0.0}$ \\
    \bottomrule
  \end{tabular}
\end{table}

\begin{table}[ht]
  \centering
  \fontsize{8}{9.5}\selectfont
  \caption{\footnotesize Mean and standard deviation of $\mathrm{pass}@K$ for Qwen2.5-VL-32B-Instruct ($K \in \{1,2,4\}$) by error category (left) and paper category (right). Detailed evaluation results for text-only evaluation of Table~\ref{tab:multi_modality}.}
  \label{tab:passk-text-openrouter_qwen_qwen2_5-vl-32b-instruct}
  \begin{tabular}{@{} lccc @{\quad} lccc @{} }
    \toprule
    \multicolumn{4}{c}{\textbf{Error Category}} & \multicolumn{4}{c}{\textbf{Paper Category}} \\
    \cmidrule(r){1-4} \cmidrule(l){5-8}
   \multicolumn{1}{c}{Category} & pass@1 & pass@2 & pass@4 & \multicolumn{1}{c}{Category} & pass@1 & pass@2 & pass@4 \\
    \midrule
    Data Inconsistency & $4.6_{7.4}$ & $8.9_{8.3}$ & $14.1_{6.0}$ & Biology & $0.0_{0.0}$ & $0.0_{0.0}$ & $0.0_{0.0}$ \\
    Equation / proof & $0.0_{0.0}$ & $0.0_{0.0}$ & $0.0_{0.0}$ & Computer Science & $0.0_{0.0}$ & $0.0_{0.0}$ & $0.0_{0.0}$ \\
    Experiment setup & $0.0_{0.0}$ & $0.0_{0.0}$ & $0.0_{0.0}$ & Environmental Science & $0.0_{0.0}$ & $0.0_{0.0}$ & $0.0_{0.0}$ \\
    Reagent identity & $4.7_{11.6}$ & $9.5_{15.1}$ & $19.3_{16.5}$ & Materials Science & $0.0_{0.0}$ & $0.0_{0.0}$ & $0.0_{0.0}$ \\
    Statistical reporting & $0.0_{0.0}$ & $0.0_{0.0}$ & $0.0_{0.0}$ & Mathematics & $0.0_{0.0}$ & $0.0_{0.0}$ & $0.0_{0.0}$ \\
    & & & & Medicine & $14.2_{34.9}$ & $28.5_{45.2}$ & $57.8_{49.4}$ \\
    & & & & Multidisciplinary & $9.1_{14.9}$ & $17.7_{16.6}$ & $28.3_{11.9}$ \\
    & & & & Physics & $0.0_{0.0}$ & $0.0_{0.0}$ & $0.0_{0.0}$ \\
    \bottomrule
  \end{tabular}
\end{table}

\begin{table}[ht]
  \centering
  \fontsize{8}{9.5}\selectfont
  \caption{\footnotesize Mean and standard deviation of $\mathrm{pass}@K$ for Llama-4-Maverick ($K \in \{1,2,4\}$) by error category (left) and paper category (right). Detailed evaluation results for text-only evaluation of Table~\ref{tab:multi_modality}.}
  \label{tab:passk-text-openrouter_meta-llama_llama-4-m}
  \begin{tabular}{@{} lccc @{\quad} lccc @{} }
    \toprule
    \multicolumn{4}{c}{\textbf{Error Category}} & \multicolumn{4}{c}{\textbf{Paper Category}} \\
    \cmidrule(r){1-4} \cmidrule(l){5-8}
   \multicolumn{1}{c}{Category} & pass@1 & pass@2 & pass@4 & \multicolumn{1}{c}{Category} & pass@1 & pass@2 & pass@4 \\
    \midrule
    Data Inconsistency & $0.0_{0.0}$ & $0.0_{0.0}$ & $0.0_{0.0}$ & Biology & $6.7_{17.0}$ & $13.9_{22.4}$ & $28.3_{24.8}$ \\
    Equation / proof & $0.8_{1.3}$ & $1.7_{1.8}$ & $3.5_{2.0}$ & Computer Science & $1.1_{2.8}$ & $2.2_{3.7}$ & $4.9_{4.1}$ \\
    Experiment setup & $0.0_{0.0}$ & $0.0_{0.0}$ & $0.0_{0.0}$ & Environmental Science & $0.0_{0.0}$ & $0.0_{0.0}$ & $0.0_{0.0}$ \\
    Reagent identity & $4.5_{11.4}$ & $9.3_{14.9}$ & $18.9_{16.5}$ & Materials Science & $0.0_{0.0}$ & $0.0_{0.0}$ & $0.0_{0.0}$ \\
    Statistical reporting & $0.0_{0.0}$ & $0.0_{0.0}$ & $0.0_{0.0}$ & Mathematics & $0.0_{0.0}$ & $0.0_{0.0}$ & $0.0_{0.0}$ \\
    & & & & Medicine & $0.0_{0.0}$ & $0.0_{0.0}$ & $0.0_{0.0}$ \\
    & & & & Multidisciplinary & $4.7_{11.6}$ & $9.7_{15.2}$ & $18.6_{16.6}$ \\
    & & & & Physics & $0.0_{0.0}$ & $0.0_{0.0}$ & $0.0_{0.0}$ \\
    \bottomrule
  \end{tabular}
\end{table}

\begin{table}[ht]
  \centering
  \fontsize{8}{9.5}\selectfont
  \caption{\footnotesize Mean and standard deviation of $\mathrm{pass}@K$ for Llama-4-Scout ($K \in \{1,2,4\}$) by error category (left) and paper category (right). Detailed evaluation results for text-only evaluation of Table~\ref{tab:multi_modality}.}
  \label{tab:passk-text-openrouter_meta-llama_llama-4-s}
  \begin{tabular}{@{} lccc @{\quad} lccc @{} }
    \toprule
    \multicolumn{4}{c}{\textbf{Error Category}} & \multicolumn{4}{c}{\textbf{Paper Category}} \\
    \cmidrule(r){1-4} \cmidrule(l){5-8}
   \multicolumn{1}{c}{Category} & pass@1 & pass@2 & pass@4 & \multicolumn{1}{c}{Category} & pass@1 & pass@2 & pass@4 \\
    \midrule
    Data Inconsistency & $6.9_{8.2}$ & $13.1_{9.7}$ & $23.9_{9.2}$ & Biology & $6.5_{16.8}$ & $13.1_{22.0}$ & $29.6_{24.6}$ \\
    Equation / proof & $0.8_{1.3}$ & $1.5_{1.5}$ & $2.6_{1.0}$ & Computer Science & $2.3_{3.7}$ & $4.1_{4.2}$ & $7.2_{2.8}$ \\
    Experiment setup & $0.0_{0.0}$ & $0.0_{0.0}$ & $0.0_{0.0}$ & Environmental Science & $14.1_{22.5}$ & $26.2_{25.0}$ & $42.2_{18.2}$ \\
    Reagent identity & $4.3_{11.2}$ & $8.7_{14.6}$ & $19.7_{16.4}$ & Materials Science & $6.5_{16.8}$ & $13.1_{22.0}$ & $29.6_{24.6}$ \\
    Statistical reporting & $0.0_{0.0}$ & $0.0_{0.0}$ & $0.0_{0.0}$ & Mathematics & $0.0_{0.0}$ & $0.0_{0.0}$ & $0.0_{0.0}$ \\
    & & & & Medicine & $0.0_{0.0}$ & $0.0_{0.0}$ & $0.0_{0.0}$ \\
    & & & & Multidisciplinary & $0.0_{0.0}$ & $0.0_{0.0}$ & $0.0_{0.0}$ \\
    & & & & Physics & $0.0_{0.0}$ & $0.0_{0.0}$ & $0.0_{0.0}$ \\
    \bottomrule
  \end{tabular}
\end{table}

%% file: errata.bbl
\begin{thebibliography}{75}
\providecommand{\natexlab}[1]{#1}
\providecommand{\url}[1]{\texttt{#1}}
\expandafter\ifx\csname urlstyle\endcsname\relax
  \providecommand{\doi}[1]{doi: #1}\else
  \providecommand{\doi}{doi: \begingroup \urlstyle{rm}\Url}\fi

\bibitem[Radford et~al.(2018)Radford, Narasimhan, Salimans, Sutskever, et~al.]{radford2018improving}
Alec Radford, Karthik Narasimhan, Tim Salimans, Ilya Sutskever, et~al.
\newblock Improving language understanding by generative pre-training.
\newblock 2018.

\bibitem[Brown et~al.(2020)Brown, Mann, Ryder, Subbiah, Kaplan, Dhariwal, Neelakantan, Shyam, Sastry, Askell, et~al.]{brown2020language}
Tom Brown, Benjamin Mann, Nick Ryder, Melanie Subbiah, Jared~D Kaplan, Prafulla Dhariwal, Arvind Neelakantan, Pranav Shyam, Girish Sastry, Amanda Askell, et~al.
\newblock Language models are few-shot learners.
\newblock \emph{Advances in neural information processing systems}, 33:\penalty0 1877--1901, 2020.

\bibitem[Guo et~al.(2025{\natexlab{a}})Guo, Xu, Zhang, Song, Peng, Deng, Dong, Nakayama, Geng, Wang, et~al.]{guo2025r}
Meng-Hao Guo, Jiajun Xu, Yi~Zhang, Jiaxi Song, Haoyang Peng, Yi-Xuan Deng, Xinzhi Dong, Kiyohiro Nakayama, Zhengyang Geng, Chen Wang, et~al.
\newblock R-bench: Graduate-level multi-disciplinary benchmarks for llm \& mllm complex reasoning evaluation.
\newblock \emph{arXiv preprint arXiv:2505.02018}, 2025{\natexlab{a}}.

\bibitem[Rein et~al.(2024)Rein, Hou, Stickland, Petty, Pang, Dirani, Michael, and Bowman]{rein2024gpqa}
David Rein, Betty~Li Hou, Asa~Cooper Stickland, Jackson Petty, Richard~Yuanzhe Pang, Julien Dirani, Julian Michael, and Samuel~R Bowman.
\newblock Gpqa: A graduate-level google-proof q\&a benchmark.
\newblock In \emph{First Conference on Language Modeling}, 2024.

\bibitem[Feng et~al.(2025)Feng, Zhao, Liu, Yang, Zhao, Sous, and Cohan]{feng2025physics}
Kaiyue Feng, Yilun Zhao, Yixin Liu, Tianyu Yang, Chen Zhao, John Sous, and Arman Cohan.
\newblock Physics: Benchmarking foundation models on university-level physics problem solving.
\newblock \emph{arXiv preprint arXiv:2503.21821}, 2025.

\bibitem[Si et~al.(2024)Si, Yang, and Hashimoto]{si2024can}
Chenglei Si, Diyi Yang, and Tatsunori Hashimoto.
\newblock Can llms generate novel research ideas? a large-scale human study with 100+ nlp researchers.
\newblock \emph{arXiv preprint arXiv:2409.04109}, 2024.

\bibitem[Park et~al.(2024{\natexlab{a}})Park, Kaplan, Ren, Hsu, Li, Xu, Li, and Li]{park2024can}
Yang~Jeong Park, Daniel Kaplan, Zhichu Ren, Chia-Wei Hsu, Changhao Li, Haowei Xu, Sipei Li, and Ju~Li.
\newblock Can chatgpt be used to generate scientific hypotheses?
\newblock \emph{Journal of Materiomics}, 10\penalty0 (3):\penalty0 578--584, 2024{\natexlab{a}}.

\bibitem[He et~al.(2025)He, Huang, Feng, Lin, Zhang, Li, et~al.]{he2025pasa}
Yichen He, Guanhua Huang, Peiyuan Feng, Yuan Lin, Yuchen Zhang, Hang Li, et~al.
\newblock Pasa: An llm agent for comprehensive academic paper search.
\newblock \emph{arXiv preprint arXiv:2501.10120}, 2025.

\bibitem[Jain and Jain(2024)]{jain2024generative}
Rishab Jain and Aditya Jain.
\newblock Generative ai in writing research papers: a new type of algorithmic bias and uncertainty in scholarly work.
\newblock In \emph{Intelligent Systems Conference}, pages 656--669. Springer, 2024.

\bibitem[Gottweis et~al.(2025)Gottweis, Weng, Daryin, Tu, Palepu, Sirkovic, Myaskovsky, Weissenberger, Rong, Tanno, et~al.]{gottweis2025towards}
Juraj Gottweis, Wei-Hung Weng, Alexander Daryin, Tao Tu, Anil Palepu, Petar Sirkovic, Artiom Myaskovsky, Felix Weissenberger, Keran Rong, Ryutaro Tanno, et~al.
\newblock Towards an ai co-scientist.
\newblock \emph{arXiv preprint arXiv:2502.18864}, 2025.

\bibitem[Lu et~al.(2024)Lu, Lu, Lange, Foerster, Clune, and Ha]{lu2024ai}
Chris Lu, Cong Lu, Robert~Tjarko Lange, Jakob Foerster, Jeff Clune, and David Ha.
\newblock The ai scientist: Towards fully automated open-ended scientific discovery.
\newblock \emph{arXiv preprint arXiv:2408.06292}, 2024.

\bibitem[Penad{\'e}s et~al.(2025)Penad{\'e}s, Gottweis, He, Patkowski, Shurick, Weng, Tu, Palepu, Myaskovsky, Pawlosky, et~al.]{penades2025ai}
Jos{\'e}~R Penad{\'e}s, Juraj Gottweis, Lingchen He, Jonasz~B Patkowski, Alexander Shurick, Wei-Hung Weng, Tao Tu, Anil Palepu, Artiom Myaskovsky, Annalisa Pawlosky, et~al.
\newblock Ai mirrors experimental science to uncover a novel mechanism of gene transfer crucial to bacterial evolution.
\newblock \emph{bioRxiv}, pages 2025--02, 2025.

\bibitem[M.~Bran et~al.(2024)M.~Bran, Cox, Schilter, Baldassari, White, and Schwaller]{m2024augmenting}
Andres M.~Bran, Sam Cox, Oliver Schilter, Carlo Baldassari, Andrew~D White, and Philippe Schwaller.
\newblock Augmenting large language models with chemistry tools.
\newblock \emph{Nature Machine Intelligence}, 6\penalty0 (5):\penalty0 525--535, 2024.

\bibitem[Pan et~al.(2025)Pan, Mudur, Taranto, Tikhanovskaya, Venugopalan, Bahri, Brenner, and Kim]{pan2025quantum}
Haining Pan, Nayantara Mudur, William Taranto, Maria Tikhanovskaya, Subhashini Venugopalan, Yasaman Bahri, Michael~P Brenner, and Eun-Ah Kim.
\newblock Quantum many-body physics calculations with large language models.
\newblock \emph{Communications Physics}, 8\penalty0 (1):\penalty0 49, 2025.

\bibitem[{DeepMind}(2025)]{deepmind2025alphaevolve}
{DeepMind}.
\newblock Alphaevolve: a gemini-powered coding agent for designing advanced algorithms.
\newblock \url{https://deepmind.google/discover/blog/alphaevolve-a-gemini-powered-coding-agent-for-designing-advanced-algorithms/}, 2025.
\newblock Accessed: 2025-05-15.

\bibitem[Zheng et~al.(2023)Zheng, Chiang, Sheng, Zhuang, Wu, Zhuang, Lin, Li, Li, Xing, et~al.]{zheng2023judging}
Lianmin Zheng, Wei-Lin Chiang, Ying Sheng, Siyuan Zhuang, Zhanghao Wu, Yonghao Zhuang, Zi~Lin, Zhuohan Li, Dacheng Li, Eric Xing, et~al.
\newblock Judging llm-as-a-judge with mt-bench and chatbot arena.
\newblock \emph{Advances in Neural Information Processing Systems}, 36:\penalty0 46595--46623, 2023.

\bibitem[Chen et~al.(2019)Chen, Wang, Chen, Zhang, Wang, Li, Zhou, and Wang]{chen2019tabfact}
Wenhu Chen, Hongmin Wang, Jianshu Chen, Yunkai Zhang, Hong Wang, Shiyang Li, Xiyou Zhou, and William~Yang Wang.
\newblock Tabfact: A large-scale dataset for table-based fact verification.
\newblock \emph{arXiv preprint arXiv:1909.02164}, 2019.

\bibitem[Bekoulis et~al.(2021)Bekoulis, Papagiannopoulou, and Deligiannis]{bekoulis2021review}
Giannis Bekoulis, Christina Papagiannopoulou, and Nikos Deligiannis.
\newblock A review on fact extraction and verification.
\newblock \emph{ACM Computing Surveys (CSUR)}, 55\penalty0 (1):\penalty0 1--35, 2021.

\bibitem[Zhang et~al.(2025)Zhang, Anjum, Fan, Zheng, Huang, and Feng]{zhang2025poly}
Hanzhi Zhang, Sumera Anjum, Heng Fan, Weijian Zheng, Yan Huang, and Yunhe Feng.
\newblock Poly-fever: A multilingual fact verification benchmark for hallucination detection in large language models.
\newblock \emph{arXiv preprint arXiv:2503.16541}, 2025.

\bibitem[Ortega and G{\'o}mez-P{\'e}rez(2025)]{ortega2025sciclaims}
Ra{\'u}l Ortega and Jos{\'e}~Manuel G{\'o}mez-P{\'e}rez.
\newblock Sciclaims: An end-to-end generative system for biomedical claim analysis.
\newblock \emph{arXiv preprint arXiv:2503.18526}, 2025.

\bibitem[Kumar et~al.(2025)Kumar, Sharma, Khincha, Shroff, Singh, and Mishra]{kumar2025sciclaimhunt}
Sujit Kumar, Anshul Sharma, Siddharth~Hemant Khincha, Gargi Shroff, Sanasam~Ranbir Singh, and Rahul Mishra.
\newblock Sciclaimhunt: A large dataset for evidence-based scientific claim verification.
\newblock \emph{arXiv preprint arXiv:2502.10003}, 2025.

\bibitem[Siegel et~al.(2024)Siegel, Kapoor, Nagdir, Stroebl, and Narayanan]{siegel2024core}
Zachary~S Siegel, Sayash Kapoor, Nitya Nagdir, Benedikt Stroebl, and Arvind Narayanan.
\newblock Core-bench: Fostering the credibility of published research through a computational reproducibility agent benchmark.
\newblock \emph{arXiv preprint arXiv:2409.11363}, 2024.

\bibitem[Dycke et~al.(2022)Dycke, Kuznetsov, and Gurevych]{dycke2022nlpeer}
Nils Dycke, Ilia Kuznetsov, and Iryna Gurevych.
\newblock Nlpeer: A unified resource for the computational study of peer review.
\newblock \emph{arXiv preprint arXiv:2211.06651}, 2022.

\bibitem[Baumg{\"a}rtner et~al.(2025)Baumg{\"a}rtner, Briscoe, and Gurevych]{baumgartner2025peerqa}
Tim Baumg{\"a}rtner, Ted Briscoe, and Iryna Gurevych.
\newblock Peerqa: A scientific question answering dataset from peer reviews.
\newblock \emph{arXiv preprint arXiv:2502.13668}, 2025.

\bibitem[Bejan et~al.(2023)Bejan, Sokolov, and Filippova]{bejan-etal-2023-make}
Irina Bejan, Artem Sokolov, and Katja Filippova.
\newblock Make every example count: On the stability and utility of self-influence for learning from noisy {NLP} datasets.
\newblock In Houda Bouamor, Juan Pino, and Kalika Bali, editors, \emph{Proceedings of the 2023 Conference on Empirical Methods in Natural Language Processing}, pages 10107--10121, Singapore, December 2023. Association for Computational Linguistics.
\newblock \doi{10.18653/v1/2023.emnlp-main.625}.
\newblock URL \url{https://aclanthology.org/2023.emnlp-main.625/}.

\bibitem[Thorne et~al.(2018)Thorne, Vlachos, Christodoulopoulos, and Mittal]{thorne-etal-2018-fever}
James Thorne, Andreas Vlachos, Christos Christodoulopoulos, and Arpit Mittal.
\newblock {FEVER}: a large-scale dataset for fact extraction and {VER}ification.
\newblock In Marilyn Walker, Heng Ji, and Amanda Stent, editors, \emph{Proceedings of the 2018 Conference of the North {A}merican Chapter of the Association for Computational Linguistics: Human Language Technologies, Volume 1 (Long Papers)}, pages 809--819, New Orleans, Louisiana, June 2018. Association for Computational Linguistics.
\newblock \doi{10.18653/v1/N18-1074}.
\newblock URL \url{https://aclanthology.org/N18-1074/}.

\bibitem[Wadden et~al.(2020)Wadden, Lin, Lo, Wang, van Zuylen, Cohan, and Hajishirzi]{wadden-etal-2020-fact}
David Wadden, Shanchuan Lin, Kyle Lo, Lucy~Lu Wang, Madeleine van Zuylen, Arman Cohan, and Hannaneh Hajishirzi.
\newblock Fact or fiction: Verifying scientific claims.
\newblock In Bonnie Webber, Trevor Cohn, Yulan He, and Yang Liu, editors, \emph{Proceedings of the 2020 Conference on Empirical Methods in Natural Language Processing (EMNLP)}, pages 7534--7550, Online, November 2020. Association for Computational Linguistics.
\newblock \doi{10.18653/v1/2020.emnlp-main.609}.
\newblock URL \url{https://aclanthology.org/2020.emnlp-main.609/}.

\bibitem[Lin et~al.(2023)Lin, Song, Zhou, Chen, and Shi]{lin2023moprd}
Jialiang Lin, Jiaxin Song, Zhangping Zhou, Yidong Chen, and Xiaodong Shi.
\newblock Moprd: A multidisciplinary open peer review dataset.
\newblock \emph{Neural Computing and Applications}, 35\penalty0 (34):\penalty0 24191--24206, 2023.

\bibitem[Shin et~al.(2025)Shin, Tang, Lee, Kim, Lim, Cho, Hong, Lee, and Kim]{shin2025automatically}
Hyungyu Shin, Jingyu Tang, Yoonjoo Lee, Nayoung Kim, Hyunseung Lim, Ji~Yong Cho, Hwajung Hong, Moontae Lee, and Juho Kim.
\newblock Automatically evaluating the paper reviewing capability of large language models.
\newblock \emph{arXiv preprint arXiv:2502.17086}, 2025.

\bibitem[{OpenAI}(2025{\natexlab{a}})]{openai2025o3o4mini}
{OpenAI}.
\newblock Openai o3 and o4-mini system card.
\newblock \url{https://openai.com/index/o3-o4-mini-system-card/}, April 2025{\natexlab{a}}.
\newblock Accessed: 2025-05-12.

\bibitem[{Meta AI}(2025)]{meta2025llama4}
{Meta AI}.
\newblock The llama 4 herd: The beginning of a new era of natively multimodal ai innovation.
\newblock \url{https://ai.meta.com/blog/llama-4-multimodal-intelligence/}, April 2025.
\newblock Accessed: 2025-05-12.

\bibitem[Rao et~al.(2024)Rao, Young, Dietterich, and Callison-Burch]{rao2024WithdrarXiv}
Delip Rao, Jonathan Young, Thomas Dietterich, and Chris Callison-Burch.
\newblock Withdrarxiv: A large-scale dataset for retraction study.
\newblock \emph{arXiv preprint arXiv:2412.03775}, 2024.

\bibitem[Ortega(2022)]{ortega2022classification}
Jos{\'e}~Luis Ortega.
\newblock Classification and analysis of pubpeer comments: How a web journal club is used.
\newblock \emph{Journal of the Association for Information Science and Technology}, 73\penalty0 (5):\penalty0 655--670, 2022.

\bibitem[OpenAI et~al.(2024)OpenAI, :, Hurst, Lerer, Goucher, Perelman, Ramesh, Clark, Ostrow, Welihinda, Hayes, Radford, Mądry, Baker-Whitcomb, Beutel, Borzunov, Carney, Chow, Kirillov, Nichol, Paino, Renzin, Passos, Kirillov, Christakis, Conneau, Kamali, Jabri, Moyer, Tam, Crookes, Tootoochian, Tootoonchian, Kumar, Vallone, Karpathy, Braunstein, Cann, Codispoti, Galu, Kondrich, Tulloch, Mishchenko, Baek, Jiang, Pelisse, Woodford, Gosalia, Dhar, Pantuliano, Nayak, Oliver, Zoph, Ghorbani, Leimberger, Rossen, Sokolowsky, Wang, Zweig, Hoover, Samic, McGrew, Spero, Giertler, Cheng, Lightcap, Walkin, Quinn, Guarraci, Hsu, Kellogg, Eastman, Lugaresi, Wainwright, Bassin, Hudson, Chu, Nelson, Li, Shern, Conger, Barette, Voss, Ding, Lu, Zhang, Beaumont, Hallacy, Koch, Gibson, Kim, Choi, McLeavey, Hesse, Fischer, Winter, Czarnecki, Jarvis, Wei, Koumouzelis, Sherburn, Kappler, Levin, Levy, Carr, Farhi, Mely, Robinson, Sasaki, Jin, Valladares, Tsipras, Li, Nguyen, Findlay, Oiwoh, Wong, Asdar, Proehl, Yang, Antonow,
  Kramer, Peterson, Sigler, Wallace, Brevdo, Mays, Khorasani, Such, Raso, Zhang, von Lohmann, Sulit, Goh, Oden, Salmon, Starace, Brockman, Salman, Bao, Hu, Wong, Wang, Schmidt, Whitney, Jun, Kirchner, de~Oliveira~Pinto, Ren, Chang, Chung, Kivlichan, O'Connell, O'Connell, Osband, Silber, Sohl, Okuyucu, Lan, Kostrikov, Sutskever, Kanitscheider, Gulrajani, Coxon, Menick, Pachocki, Aung, Betker, Crooks, Lennon, Kiros, Leike, Park, Kwon, Phang, Teplitz, Wei, Wolfe, Chen, Harris, Varavva, Lee, Shieh, Lin, Yu, Weng, Tang, Yu, Jang, Candela, Beutler, Landers, Parish, Heidecke, Schulman, Lachman, McKay, Uesato, Ward, Kim, Huizinga, Sitkin, Kraaijeveld, Gross, Kaplan, Snyder, Achiam, Jiao, Lee, Zhuang, Harriman, Fricke, Hayashi, Singhal, Shi, Karthik, Wood, Rimbach, Hsu, Nguyen, Gu-Lemberg, Button, Liu, Howe, Muthukumar, Luther, Ahmad, Kai, Itow, Workman, Pathak, Chen, Jing, Guy, Fedus, Zhou, Mamitsuka, Weng, McCallum, Held, Ouyang, Feuvrier, Zhang, Kondraciuk, Kaiser, Hewitt, Metz, Doshi, Aflak, Simens, Boyd,
  Thompson, Dukhan, Chen, Gray, Hudnall, Zhang, Aljubeh, Litwin, Zeng, Johnson, Shetty, Gupta, Shah, Yatbaz, Yang, Zhong, Glaese, Chen, Janner, Lampe, Petrov, Wu, Wang, Fradin, Pokrass, Castro, de~Castro, Pavlov, Brundage, Wang, Khan, Murati, Bavarian, Lin, Yesildal, Soto, Gimelshein, Cone, Staudacher, Summers, LaFontaine, Chowdhury, Ryder, Stathas, Turley, Tezak, Felix, Kudige, Keskar, Deutsch, Bundick, Puckett, Nachum, Okelola, Boiko, Murk, Jaffe, Watkins, Godement, Campbell-Moore, Chao, McMillan, Belov, Su, Bak, Bakkum, Deng, Dolan, Hoeschele, Welinder, Tillet, Pronin, Tillet, Dhariwal, Yuan, Dias, Lim, Arora, Troll, Lin, Lopes, Puri, Miyara, Leike, Gaubert, Zamani, Wang, Donnelly, Honsby, Smith, Sahai, Ramchandani, Huet, Carmichael, Zellers, Chen, Chen, Nigmatullin, Cheu, Jain, Altman, Schoenholz, Toizer, Miserendino, Agarwal, Culver, Ethersmith, Gray, Grove, Metzger, Hermani, Jain, Zhao, Wu, Jomoto, Wu, Shuaiqi, Xia, Phene, Papay, Narayanan, Coffey, Lee, Hall, Balaji, Broda, Stramer, Xu, Gogineni,
  Christianson, Sanders, Patwardhan, Cunninghman, Degry, Dimson, Raoux, Shadwell, Zheng, Underwood, Markov, Sherbakov, Rubin, Stasi, Kaftan, Heywood, Peterson, Walters, Eloundou, Qi, Moeller, Monaco, Kuo, Fomenko, Chang, Zheng, Zhou, Manassra, Sheu, Zaremba, Patil, Qian, Kim, Cheng, Zhang, He, Zhang, Jin, Dai, and Malkov]{openai2024gpt4ocard}
OpenAI, :, Aaron Hurst, Adam Lerer, Adam~P. Goucher, Adam Perelman, Aditya Ramesh, Aidan Clark, AJ~Ostrow, Akila Welihinda, Alan Hayes, Alec Radford, Aleksander Mądry, Alex Baker-Whitcomb, Alex Beutel, Alex Borzunov, Alex Carney, Alex Chow, Alex Kirillov, Alex Nichol, Alex Paino, Alex Renzin, Alex~Tachard Passos, Alexander Kirillov, Alexi Christakis, Alexis Conneau, Ali Kamali, Allan Jabri, Allison Moyer, Allison Tam, Amadou Crookes, Amin Tootoochian, Amin Tootoonchian, Ananya Kumar, Andrea Vallone, Andrej Karpathy, Andrew Braunstein, Andrew Cann, Andrew Codispoti, Andrew Galu, Andrew Kondrich, Andrew Tulloch, Andrey Mishchenko, Angela Baek, Angela Jiang, Antoine Pelisse, Antonia Woodford, Anuj Gosalia, Arka Dhar, Ashley Pantuliano, Avi Nayak, Avital Oliver, Barret Zoph, Behrooz Ghorbani, Ben Leimberger, Ben Rossen, Ben Sokolowsky, Ben Wang, Benjamin Zweig, Beth Hoover, Blake Samic, Bob McGrew, Bobby Spero, Bogo Giertler, Bowen Cheng, Brad Lightcap, Brandon Walkin, Brendan Quinn, Brian Guarraci, Brian Hsu,
  Bright Kellogg, Brydon Eastman, Camillo Lugaresi, Carroll Wainwright, Cary Bassin, Cary Hudson, Casey Chu, Chad Nelson, Chak Li, Chan~Jun Shern, Channing Conger, Charlotte Barette, Chelsea Voss, Chen Ding, Cheng Lu, Chong Zhang, Chris Beaumont, Chris Hallacy, Chris Koch, Christian Gibson, Christina Kim, Christine Choi, Christine McLeavey, Christopher Hesse, Claudia Fischer, Clemens Winter, Coley Czarnecki, Colin Jarvis, Colin Wei, Constantin Koumouzelis, Dane Sherburn, Daniel Kappler, Daniel Levin, Daniel Levy, David Carr, David Farhi, David Mely, David Robinson, David Sasaki, Denny Jin, Dev Valladares, Dimitris Tsipras, Doug Li, Duc~Phong Nguyen, Duncan Findlay, Edede Oiwoh, Edmund Wong, Ehsan Asdar, Elizabeth Proehl, Elizabeth Yang, Eric Antonow, Eric Kramer, Eric Peterson, Eric Sigler, Eric Wallace, Eugene Brevdo, Evan Mays, Farzad Khorasani, Felipe~Petroski Such, Filippo Raso, Francis Zhang, Fred von Lohmann, Freddie Sulit, Gabriel Goh, Gene Oden, Geoff Salmon, Giulio Starace, Greg Brockman, Hadi
  Salman, Haiming Bao, Haitang Hu, Hannah Wong, Haoyu Wang, Heather Schmidt, Heather Whitney, Heewoo Jun, Hendrik Kirchner, Henrique~Ponde de~Oliveira~Pinto, Hongyu Ren, Huiwen Chang, Hyung~Won Chung, Ian Kivlichan, Ian O'Connell, Ian O'Connell, Ian Osband, Ian Silber, Ian Sohl, Ibrahim Okuyucu, Ikai Lan, Ilya Kostrikov, Ilya Sutskever, Ingmar Kanitscheider, Ishaan Gulrajani, Jacob Coxon, Jacob Menick, Jakub Pachocki, James Aung, James Betker, James Crooks, James Lennon, Jamie Kiros, Jan Leike, Jane Park, Jason Kwon, Jason Phang, Jason Teplitz, Jason Wei, Jason Wolfe, Jay Chen, Jeff Harris, Jenia Varavva, Jessica~Gan Lee, Jessica Shieh, Ji~Lin, Jiahui Yu, Jiayi Weng, Jie Tang, Jieqi Yu, Joanne Jang, Joaquin~Quinonero Candela, Joe Beutler, Joe Landers, Joel Parish, Johannes Heidecke, John Schulman, Jonathan Lachman, Jonathan McKay, Jonathan Uesato, Jonathan Ward, Jong~Wook Kim, Joost Huizinga, Jordan Sitkin, Jos Kraaijeveld, Josh Gross, Josh Kaplan, Josh Snyder, Joshua Achiam, Joy Jiao, Joyce Lee, Juntang
  Zhuang, Justyn Harriman, Kai Fricke, Kai Hayashi, Karan Singhal, Katy Shi, Kavin Karthik, Kayla Wood, Kendra Rimbach, Kenny Hsu, Kenny Nguyen, Keren Gu-Lemberg, Kevin Button, Kevin Liu, Kiel Howe, Krithika Muthukumar, Kyle Luther, Lama Ahmad, Larry Kai, Lauren Itow, Lauren Workman, Leher Pathak, Leo Chen, Li~Jing, Lia Guy, Liam Fedus, Liang Zhou, Lien Mamitsuka, Lilian Weng, Lindsay McCallum, Lindsey Held, Long Ouyang, Louis Feuvrier, Lu~Zhang, Lukas Kondraciuk, Lukasz Kaiser, Luke Hewitt, Luke Metz, Lyric Doshi, Mada Aflak, Maddie Simens, Madelaine Boyd, Madeleine Thompson, Marat Dukhan, Mark Chen, Mark Gray, Mark Hudnall, Marvin Zhang, Marwan Aljubeh, Mateusz Litwin, Matthew Zeng, Max Johnson, Maya Shetty, Mayank Gupta, Meghan Shah, Mehmet Yatbaz, Meng~Jia Yang, Mengchao Zhong, Mia Glaese, Mianna Chen, Michael Janner, Michael Lampe, Michael Petrov, Michael Wu, Michele Wang, Michelle Fradin, Michelle Pokrass, Miguel Castro, Miguel Oom~Temudo de~Castro, Mikhail Pavlov, Miles Brundage, Miles Wang, Minal
  Khan, Mira Murati, Mo~Bavarian, Molly Lin, Murat Yesildal, Nacho Soto, Natalia Gimelshein, Natalie Cone, Natalie Staudacher, Natalie Summers, Natan LaFontaine, Neil Chowdhury, Nick Ryder, Nick Stathas, Nick Turley, Nik Tezak, Niko Felix, Nithanth Kudige, Nitish Keskar, Noah Deutsch, Noel Bundick, Nora Puckett, Ofir Nachum, Ola Okelola, Oleg Boiko, Oleg Murk, Oliver Jaffe, Olivia Watkins, Olivier Godement, Owen Campbell-Moore, Patrick Chao, Paul McMillan, Pavel Belov, Peng Su, Peter Bak, Peter Bakkum, Peter Deng, Peter Dolan, Peter Hoeschele, Peter Welinder, Phil Tillet, Philip Pronin, Philippe Tillet, Prafulla Dhariwal, Qiming Yuan, Rachel Dias, Rachel Lim, Rahul Arora, Rajan Troll, Randall Lin, Rapha~Gontijo Lopes, Raul Puri, Reah Miyara, Reimar Leike, Renaud Gaubert, Reza Zamani, Ricky Wang, Rob Donnelly, Rob Honsby, Rocky Smith, Rohan Sahai, Rohit Ramchandani, Romain Huet, Rory Carmichael, Rowan Zellers, Roy Chen, Ruby Chen, Ruslan Nigmatullin, Ryan Cheu, Saachi Jain, Sam Altman, Sam Schoenholz, Sam
  Toizer, Samuel Miserendino, Sandhini Agarwal, Sara Culver, Scott Ethersmith, Scott Gray, Sean Grove, Sean Metzger, Shamez Hermani, Shantanu Jain, Shengjia Zhao, Sherwin Wu, Shino Jomoto, Shirong Wu, Shuaiqi, Xia, Sonia Phene, Spencer Papay, Srinivas Narayanan, Steve Coffey, Steve Lee, Stewart Hall, Suchir Balaji, Tal Broda, Tal Stramer, Tao Xu, Tarun Gogineni, Taya Christianson, Ted Sanders, Tejal Patwardhan, Thomas Cunninghman, Thomas Degry, Thomas Dimson, Thomas Raoux, Thomas Shadwell, Tianhao Zheng, Todd Underwood, Todor Markov, Toki Sherbakov, Tom Rubin, Tom Stasi, Tomer Kaftan, Tristan Heywood, Troy Peterson, Tyce Walters, Tyna Eloundou, Valerie Qi, Veit Moeller, Vinnie Monaco, Vishal Kuo, Vlad Fomenko, Wayne Chang, Weiyi Zheng, Wenda Zhou, Wesam Manassra, Will Sheu, Wojciech Zaremba, Yash Patil, Yilei Qian, Yongjik Kim, Youlong Cheng, Yu~Zhang, Yuchen He, Yuchen Zhang, Yujia Jin, Yunxing Dai, and Yury Malkov.
\newblock Gpt-4o system card, 2024.
\newblock URL \url{https://arxiv.org/abs/2410.21276}.

\bibitem[{MAA}(2024)]{maa2024aime}
{MAA}.
\newblock {American Invitational Mathematics Examination – AIME}.
\newblock In \emph{American Invitational Mathematics Examination – AIME 2024}, February 2024, February 2024.
\newblock URL \url{https://maa.org/math-competitions/american-invitational-mathematics-examination-aime}.

\bibitem[Petrov et~al.(2025)Petrov, Dekoninck, Baltadzhiev, Drencheva, Minchev, Balunovi{\'c}, Jovanovi{\'c}, and Vechev]{petrov2025proof}
Ivo Petrov, Jasper Dekoninck, Lyuben Baltadzhiev, Maria Drencheva, Kristian Minchev, Mislav Balunovi{\'c}, Nikola Jovanovi{\'c}, and Martin Vechev.
\newblock Proof or bluff? evaluating llms on 2025 usa math olympiad.
\newblock \emph{arXiv preprint arXiv:2503.21934}, 2025.

\bibitem[Starace et~al.(2025)Starace, Jaffe, Sherburn, Aung, Chan, Maksin, Dias, Mays, Kinsella, Thompson, et~al.]{starace2025paperbench}
Giulio Starace, Oliver Jaffe, Dane Sherburn, James Aung, Jun~Shern Chan, Leon Maksin, Rachel Dias, Evan Mays, Benjamin Kinsella, Wyatt Thompson, et~al.
\newblock Paperbench: Evaluating ai's ability to replicate ai research.
\newblock \emph{arXiv preprint arXiv:2504.01848}, 2025.

\bibitem[Seo et~al.(2025)Seo, Baek, Lee, and Hwang]{seo2025paper2code}
Minju Seo, Jinheon Baek, Seongyun Lee, and Sung~Ju Hwang.
\newblock Paper2code: Automating code generation from scientific papers in machine learning.
\newblock \emph{arXiv preprint arXiv:2504.17192}, 2025.

\bibitem[Hurst et~al.(2024)Hurst, Lerer, Goucher, Perelman, Ramesh, Clark, Ostrow, Welihinda, Hayes, Radford, et~al.]{hurst2024gpt}
Aaron Hurst, Adam Lerer, Adam~P Goucher, Adam Perelman, Aditya Ramesh, Aidan Clark, AJ~Ostrow, Akila Welihinda, Alan Hayes, Alec Radford, et~al.
\newblock Gpt-4o system card.
\newblock \emph{arXiv preprint arXiv:2410.21276}, 2024.

\bibitem[{OpenAI}(2025{\natexlab{b}})]{openai2025tiktoken}
{OpenAI}.
\newblock {tiktoken}: A fast bpe tokeniser for use with openai's models.
\newblock \url{https://github.com/openai/tiktoken}, 2025{\natexlab{b}}.
\newblock GitHub repository; version 0.9.0 (Feb. 14, 2025); accessed May 12, 2025.

\bibitem[Ni et~al.(2024)Ni, Xue, Yue, Deng, Shah, Jain, Neubig, and You]{ni2024mixeval}
Jinjie Ni, Fuzhao Xue, Xiang Yue, Yuntian Deng, Mahir Shah, Kabir Jain, Graham Neubig, and Yang You.
\newblock Mixeval: Deriving wisdom of the crowd from llm benchmark mixtures.
\newblock \emph{arXiv preprint arXiv:2406.06565}, 2024.

\bibitem[Kulal et~al.(2019)Kulal, Pasupat, Chandra, Lee, Padon, Aiken, and Liang]{kulal2019spoc}
Sumith Kulal, Panupong Pasupat, Kartik Chandra, Mina Lee, Oded Padon, Alex Aiken, and Percy~S Liang.
\newblock Spoc: Search-based pseudocode to code.
\newblock \emph{Advances in Neural Information Processing Systems}, 32, 2019.

\bibitem[Chen et~al.(2021)Chen, Tworek, Jun, Yuan, Pinto, Kaplan, Edwards, Burda, Joseph, Brockman, et~al.]{chen2021evaluating}
Mark Chen, Jerry Tworek, Heewoo Jun, Qiming Yuan, Henrique Ponde De~Oliveira Pinto, Jared Kaplan, Harri Edwards, Yuri Burda, Nicholas Joseph, Greg Brockman, et~al.
\newblock Evaluating large language models trained on code.
\newblock \emph{arXiv preprint arXiv:2107.03374}, 2021.

\bibitem[{OpenAI}(2025{\natexlab{c}})]{openai_gpt41}
{OpenAI}.
\newblock {GPT-4.1}.
\newblock \url{https://openai.com/index/gpt-4-1/}, 2025{\natexlab{c}}.
\newblock Accessed: May 15, 2025.

\bibitem[{Google Cloud}(2025{\natexlab{a}})]{google2025gemini2_5_pro}
{Google Cloud}.
\newblock Gemini 2.5 pro.
\newblock \url{https://cloud.google.com/vertex-ai/generative-ai/docs/models/gemini/2-5-pro}, 2025{\natexlab{a}}.
\newblock Accessed: 2025-05-12.

\bibitem[{Google Cloud}(2025{\natexlab{b}})]{google_cloud_gemini_flash_lite}
{Google Cloud}.
\newblock {Gemini 2.0 Flash Lite}.
\newblock \url{https://cloud.google.com/vertex-ai/generative-ai/docs/models/gemini/2-0-flash-lite}, 2025{\natexlab{b}}.
\newblock Accessed: May 15, 2025.

\bibitem[{Anthropic}(2025)]{anthropic_claude3_7_sonnet}
{Anthropic}.
\newblock {Claude 3.7 Sonnet System Card}.
\newblock \url{https://assets.anthropic.com/m/785e231869ea8b3b/original/claude-3-7-sonnet-system-card.pdf}, 2025.
\newblock Accessed: May 15, 2025.

\bibitem[Bai et~al.(2025)Bai, Chen, Liu, Wang, Ge, Song, Dang, Wang, Wang, Tang, et~al.]{bai2025qwen2}
Shuai Bai, Keqin Chen, Xuejing Liu, Jialin Wang, Wenbin Ge, Sibo Song, Kai Dang, Peng Wang, Shijie Wang, Jun Tang, et~al.
\newblock Qwen2. 5-vl technical report.
\newblock \emph{arXiv preprint arXiv:2502.13923}, 2025.

\bibitem[Yue et~al.(2024)Yue, Ni, Zhang, Zheng, Liu, Zhang, Stevens, Jiang, Ren, Sun, et~al.]{yue2024mmmu}
Xiang Yue, Yuansheng Ni, Kai Zhang, Tianyu Zheng, Ruoqi Liu, Ge~Zhang, Samuel Stevens, Dongfu Jiang, Weiming Ren, Yuxuan Sun, et~al.
\newblock Mmmu: A massive multi-discipline multimodal understanding and reasoning benchmark for expert agi.
\newblock In \emph{Proceedings of the IEEE/CVF Conference on Computer Vision and Pattern Recognition}, pages 9556--9567, 2024.

\bibitem[Lu et~al.(2023)Lu, Bansal, Xia, Liu, Li, Hajishirzi, Cheng, Chang, Galley, and Gao]{lu2023mathvista}
Pan Lu, Hritik Bansal, Tony Xia, Jiacheng Liu, Chunyuan Li, Hannaneh Hajishirzi, Hao Cheng, Kai-Wei Chang, Michel Galley, and Jianfeng Gao.
\newblock Mathvista: Evaluating mathematical reasoning of foundation models in visual contexts.
\newblock \emph{arXiv preprint arXiv:2310.02255}, 2023.

\bibitem[Wang et~al.(2024)Wang, Ma, Zhang, Ni, Chandra, Guo, Ren, Arulraj, He, Jiang, et~al.]{wang2024mmlu}
Yubo Wang, Xueguang Ma, Ge~Zhang, Yuansheng Ni, Abhranil Chandra, Shiguang Guo, Weiming Ren, Aaran Arulraj, Xuan He, Ziyan Jiang, et~al.
\newblock Mmlu-pro: A more robust and challenging multi-task language understanding benchmark.
\newblock In \emph{The Thirty-eight Conference on Neural Information Processing Systems Datasets and Benchmarks Track}, 2024.

\bibitem[Phan et~al.(2025)Phan, Gatti, Han, Li, Hu, Zhang, Zhang, Shaaban, Ling, Shi, et~al.]{phan2025humanity}
Long Phan, Alice Gatti, Ziwen Han, Nathaniel Li, Josephina Hu, Hugh Zhang, Chen Bo~Calvin Zhang, Mohamed Shaaban, John Ling, Sean Shi, et~al.
\newblock Humanity's last exam.
\newblock \emph{arXiv preprint arXiv:2501.14249}, 2025.

\bibitem[Guo et~al.(2017)Guo, Pleiss, Sun, and Weinberger]{guo2017calibration}
Chuan Guo, Geoff Pleiss, Yu~Sun, and Kilian~Q Weinberger.
\newblock On calibration of modern neural networks.
\newblock In \emph{International conference on machine learning}, pages 1321--1330. PMLR, 2017.

\bibitem[Ovadia et~al.(2019)Ovadia, Fertig, Ren, Nado, Sculley, Nowozin, Dillon, Lakshminarayanan, and Snoek]{ovadia2019can}
Yaniv Ovadia, Emily Fertig, Jie Ren, Zachary Nado, David Sculley, Sebastian Nowozin, Joshua Dillon, Balaji Lakshminarayanan, and Jasper Snoek.
\newblock Can you trust your model's uncertainty? evaluating predictive uncertainty under dataset shift.
\newblock \emph{Advances in neural information processing systems}, 32, 2019.

\bibitem[Guo et~al.(2025{\natexlab{b}})Guo, Yang, Zhang, Song, Zhang, Xu, Zhu, Ma, Wang, Bi, et~al.]{guo2025deepseek}
Daya Guo, Dejian Yang, Haowei Zhang, Junxiao Song, Ruoyu Zhang, Runxin Xu, Qihao Zhu, Shirong Ma, Peiyi Wang, Xiao Bi, et~al.
\newblock Deepseek-r1: Incentivizing reasoning capability in llms via reinforcement learning.
\newblock \emph{arXiv preprint arXiv:2501.12948}, 2025{\natexlab{b}}.

\bibitem[Liu et~al.(2024)Liu, Feng, Xue, Wang, Wu, Lu, Zhao, Deng, Zhang, Ruan, et~al.]{liu2024deepseek}
Aixin Liu, Bei Feng, Bing Xue, Bingxuan Wang, Bochao Wu, Chengda Lu, Chenggang Zhao, Chengqi Deng, Chenyu Zhang, Chong Ruan, et~al.
\newblock Deepseek-v3 technical report.
\newblock \emph{arXiv preprint arXiv:2412.19437}, 2024.

\bibitem[Yang et~al.(2025)Yang, Li, Yang, Zhang, Hui, Zheng, Yu, Gao, Huang, Lv, Zheng, Liu, Zhou, Huang, Hu, Ge, Wei, Lin, Tang, Yang, Tu, Zhang, Yang, Yang, Zhou, Zhou, Lin, Dang, Bao, Yang, Yu, Deng, Li, Xue, Li, Zhang, Wang, Zhu, Men, Gao, Liu, Luo, Li, Tang, Yin, Ren, Wang, Zhang, Ren, Fan, Su, Zhang, Zhang, Wan, Liu, Wang, Cui, Zhang, Zhou, and Qiu]{yang2025qwen3technicalreport}
An~Yang, Anfeng Li, Baosong Yang, Beichen Zhang, Binyuan Hui, Bo~Zheng, Bowen Yu, Chang Gao, Chengen Huang, Chenxu Lv, Chujie Zheng, Dayiheng Liu, Fan Zhou, Fei Huang, Feng Hu, Hao Ge, Haoran Wei, Huan Lin, Jialong Tang, Jian Yang, Jianhong Tu, Jianwei Zhang, Jianxin Yang, Jiaxi Yang, Jing Zhou, Jingren Zhou, Junyang Lin, Kai Dang, Keqin Bao, Kexin Yang, Le~Yu, Lianghao Deng, Mei Li, Mingfeng Xue, Mingze Li, Pei Zhang, Peng Wang, Qin Zhu, Rui Men, Ruize Gao, Shixuan Liu, Shuang Luo, Tianhao Li, Tianyi Tang, Wenbiao Yin, Xingzhang Ren, Xinyu Wang, Xinyu Zhang, Xuancheng Ren, Yang Fan, Yang Su, Yichang Zhang, Yinger Zhang, Yu~Wan, Yuqiong Liu, Zekun Wang, Zeyu Cui, Zhenru Zhang, Zhipeng Zhou, and Zihan Qiu.
\newblock Qwen3 technical report, 2025.
\newblock URL \url{https://arxiv.org/abs/2505.09388}.

\bibitem[Petersen and Tommasi(2024)]{petersen2024}
Dan Petersen and Orsola Tommasi.
\newblock Multiplicative chow-k\"unneth decomposition and homology splitting of configuration spaces, 2024.
\newblock URL \url{https://arxiv.org/abs/2401.06455}.

\bibitem[Ye et~al.(2025)Ye, Liu, Mu, Tao, Yang, Gao, Yang, and Jiang]{ye2025superacid}
Xingyao Ye, Ruoyang Liu, Xinyu Mu, Shanshan Tao, Hao Yang, Xuejiao~J Gao, Shuo-Wang Yang, and Donglin Jiang.
\newblock Superacid in situ protected synthesis of covalent organic frameworks.
\newblock \emph{Journal of the American Chemical Society}, 2025.

\bibitem[Son et~al.(2024)Son, Ko, Lee, Kim, and Hong]{son2024llm}
Guijin Son, Hyunwoo Ko, Hoyoung Lee, Yewon Kim, and Seunghyeok Hong.
\newblock Llm-as-a-judge \& reward model: What they can and cannot do.
\newblock \emph{arXiv preprint arXiv:2409.11239}, 2024.

\bibitem[Jernite et~al.(2017)Jernite, Bowman, and Sontag]{jernite2017discourse}
Yacine Jernite, Samuel~R Bowman, and David Sontag.
\newblock Discourse-based objectives for fast unsupervised sentence representation learning.
\newblock \emph{arXiv preprint arXiv:1705.00557}, 2017.

\bibitem[Wang et~al.(2022)Wang, Kordi, Mishra, Liu, Smith, Khashabi, and Hajishirzi]{wang2022self}
Yizhong Wang, Yeganeh Kordi, Swaroop Mishra, Alisa Liu, Noah~A Smith, Daniel Khashabi, and Hannaneh Hajishirzi.
\newblock Self-instruct: Aligning language models with self-generated instructions.
\newblock \emph{arXiv preprint arXiv:2212.10560}, 2022.

\bibitem[Gao et~al.(2024)Gao, Brantley, and Joachims]{gao2024reviewer2}
Zhaolin Gao, Kiant{\'e} Brantley, and Thorsten Joachims.
\newblock Reviewer2: Optimizing review generation through prompt generation.
\newblock \emph{arXiv preprint arXiv:2402.10886}, 2024.

\bibitem[Zeng et~al.(2024)Zeng, Sidhu, Blume, Chan, Wang, and Ji]{zeng2024scientific}
Qi~Zeng, Mankeerat Sidhu, Ansel Blume, Hou~Pong Chan, Lu~Wang, and Heng Ji.
\newblock Scientific opinion summarization: Paper meta-review generation dataset, methods, and evaluation.
\newblock In \emph{Artificial Intelligence for Research and Democracy: First International Workshop, AI4Research 2024, and 4th International Workshop, DemocrAI 2024, Held in Conjunction with IJCAI 2024, Jeju, South Korea, August 5, 2024, Proceedings}, page~20. Springer Nature, 2024.

\bibitem[Cortes and Lawrence(2021)]{cortes2021inconsistency}
Corinna Cortes and Neil~D Lawrence.
\newblock Inconsistency in conference peer review: Revisiting the 2014 neurips experiment.
\newblock \emph{arXiv preprint arXiv:2109.09774}, 2021.

\bibitem[Bonavia and Marin-Garcia(2023)]{bonavia2023noise}
Tomas Bonavia and Juan~A Marin-Garcia.
\newblock A noise audit of the peer review of a scientific article: a wpom journal case study.
\newblock \emph{WPOM-Working Papers on Operations Management}, 14\penalty0 (2):\penalty0 137--166, 2023.

\bibitem[Vodrahalli et~al.(2024)Vodrahalli, Ontanon, Tripuraneni, Xu, Jain, Shivanna, Hui, Dikkala, Kazemi, Fatemi, et~al.]{vodrahalli2024michelangelo}
Kiran Vodrahalli, Santiago Ontanon, Nilesh Tripuraneni, Kelvin Xu, Sanil Jain, Rakesh Shivanna, Jeffrey Hui, Nishanth Dikkala, Mehran Kazemi, Bahare Fatemi, et~al.
\newblock Michelangelo: Long context evaluations beyond haystacks via latent structure queries.
\newblock \emph{arXiv preprint arXiv:2409.12640}, 2024.

\bibitem[Jones(2021)]{jones2021scaling}
Andy~L Jones.
\newblock Scaling scaling laws with board games.
\newblock \emph{arXiv preprint arXiv:2104.03113}, 2021.

\bibitem[Son et~al.(2025)Son, Hong, Ko, and Thorne]{son2025linguistic}
Guijin Son, Jiwoo Hong, Hyunwoo Ko, and James Thorne.
\newblock Linguistic generalizability of test-time scaling in mathematical reasoning.
\newblock \emph{arXiv preprint arXiv:2502.17407}, 2025.

\bibitem[Snell et~al.(2024)Snell, Lee, Xu, and Kumar]{snell2024scaling}
Charlie Snell, Jaehoon Lee, Kelvin Xu, and Aviral Kumar.
\newblock Scaling llm test-time compute optimally can be more effective than scaling model parameters.
\newblock \emph{arXiv preprint arXiv:2408.03314}, 2024.

\bibitem[Muennighoff et~al.(2025)Muennighoff, Yang, Shi, Li, Fei-Fei, Hajishirzi, Zettlemoyer, Liang, Cand{\`e}s, and Hashimoto]{muennighoff2025s1}
Niklas Muennighoff, Zitong Yang, Weijia Shi, Xiang~Lisa Li, Li~Fei-Fei, Hannaneh Hajishirzi, Luke Zettlemoyer, Percy Liang, Emmanuel Cand{\`e}s, and Tatsunori Hashimoto.
\newblock s1: Simple test-time scaling.
\newblock \emph{arXiv preprint arXiv:2501.19393}, 2025.

\bibitem[Ahuja et~al.(2025)Ahuja, Sclar, and Tsvetkov]{ahuja2025finding}
Kabir Ahuja, Melanie Sclar, and Yulia Tsvetkov.
\newblock Finding flawed fictions: Evaluating complex reasoning in language models via plot hole detection.
\newblock \emph{arXiv preprint arXiv:2504.11900}, 2025.

\bibitem[Park et~al.(2024{\natexlab{b}})Park, Park, and Yoo]{park2024}
Jeehoon Park, Junyeong Park, and Philsang Yoo.
\newblock Algebraic description of complex conjugation on cohomology of a smooth projective hypersurface, 2024{\natexlab{b}}.
\newblock URL \url{https://arxiv.org/abs/2402.14546}.

\bibitem[Altuijri et~al.(2024)Altuijri, Atta, Abdeltwab, and Abdelhamied]{altuijri2024impacts}
Reem Altuijri, A~Atta, E~Abdeltwab, and MM~Abdelhamied.
\newblock Impacts of low energy argon beam on enhancing the surface wettability and electrical performance of ca/pani films.
\newblock \emph{ECS Journal of Solid State Science and Technology}, 13\penalty0 (4):\penalty0 043017, 2024.

\bibitem[Simpson(2024)]{simpson2024scientific}
Michael Simpson.
\newblock The scientific case against net zero: Falsifying the greenhouse gas hypothesis.
\newblock \emph{Journal of Sustainable Development}, 17\penalty0 (6):\penalty0 137--157, 2024.
\newblock \doi{10.5539/jsd.v17n6p137}.
\newblock URL \url{https://ideas.repec.org/a/ibn/jsd123/v17y2024i6p137.html}.

\end{thebibliography}
